\documentclass{article} %
\usepackage{main}

\usepackage{booktabs}
\usepackage{graphicx}
\usepackage{enumitem}
\usepackage{wrapfig}
\usepackage{wrapfig}
\usepackage{float}
\usepackage{microtype}
\usepackage{amsmath}
\usepackage{amssymb}
\usepackage{colortbl}
\usepackage[utf8]{inputenc}
\definecolor{lightgray}{rgb}{0.9,0.9,0.9}
\usepackage{caption}
\usepackage{subcaption}
\usepackage{xcolor}
\usepackage{setspace}
\usepackage{url}
\usepackage{multirow}
\usepackage{colortbl}
\usepackage{tabularx}
\usepackage{blindtext}
\usepackage{pgfplots}
\pgfplotsset{compat=1.18} 
\usepackage{tikz}
\usetikzlibrary{er,positioning,bayesnet}
\usepackage{makecell}
\usepackage{siunitx}
\usepackage{nicefrac}
\usepackage{tocloft}
\usepackage{listings}
\usepackage[raster,skins,breakable]{tcolorbox} 
\usepackage{xltabular}
\usepackage{adjustbox}
\usepackage{xurl}

\makeatother

\usepackage{xspace}
\newcommand{\dataset}{\textsc{SetwiseEvalKit}\xspace}

\newcommand{\method}{\textsc{Rubric4Setwise}\xspace}

\newcommand{\nb}[1]{\noindent\textbf{#1}}

\usepackage{xcolor}
\usepackage{dashrule}

\definecolor{systembg}{HTML}{daeced}
\definecolor{userbg}{HTML}{EEEEEE}
\definecolor{darkblue}{HTML}{00008B}
\colorlet{insert}{darkblue}

\newtcolorbox{promptbox}[1]{%
    colback=#1, colframe=userbg,
    boxrule=0pt,
    arc=0pt,
    left=2pt,right=2pt,top=2pt,bottom=2pt,
    breakable,
    width=\textwidth,
    before skip=0pt,
    after skip=0pt
}
\newcommand{\systemprompt}[1]{\noindent\begin{promptbox}{systembg}
        \textbf{System:}~#1%
    \end{promptbox}}

\newcommand{\prompttag}[1]{\hspace{0pt}\texttt{#1}}
\newcommand{\prompt}[2]{\subsection{#1}
    \vspace{-2.5ex}
    {\noindent\\
    \rule[0pt]{\textwidth}{0.7pt}\scriptsize #2
    \noindent\rule[0pt]{\textwidth}{0.6pt}\normalsize}\relax}

\usepackage{xcolor}  

\newcommand{\ie}{\textit{i.e., }}
\newcommand{\eg}{\textit{e.g., }}

\newcommand{\bfit}[1]{\textbf{\textit{#1}}}

\definecolor{CustomImageBlue}{HTML}{4169E1}

\newtcolorbox{redbox}[1]{%
  enhanced,
  colback=red!2, colframe=red!55!black, boxrule=0.7pt, arc=2pt,
  left=6pt, right=6pt, top=4pt, bottom=4pt,
  title=\textbf{#1}, coltitle=white, colbacktitle=red!55!black,
  fonttitle=\bfseries\small
}

\newtcolorbox{greenbox}[1]{%
  enhanced,
  colback=green!2, colframe=green!45!black, boxrule=0.7pt, arc=2pt,
  left=6pt, right=6pt, top=4pt, bottom=4pt,
  title=\textbf{#1}, coltitle=white, colbacktitle=green!55!black,
  fonttitle=\bfseries\small
}

\makeatletter
  \newcommand\figcaption{\def\@captype{figure}\caption}
  \newcommand\tabcaption{\def\@captype{table}\caption}
\makeatother

\newcommand{\dalgshifted}{\raisebox{0.5\depth}{$\downarrow$}}
\newcommand{\daugshifted}{\raisebox{0.5\depth}{$\uparrow$}}

\definecolor{backred}{RGB}{255, 190, 190}
\definecolor{backblue}{RGB}{210, 230, 250}
\definecolor{verylightgray}{gray}{0.95} 
\definecolor{backorange}{RGB}{255, 230, 204}
\definecolor{backyellow_soft}{RGB}{255, 252, 51}

\definecolor{myred}{RGB}{192,0,0}
\definecolor{mygreen}{RGB}{88,142,49}

\definecolor{verylightgray}{gray}{0.95} 
\definecolor{tong}{RGB}{152, 1, 0}

\definecolor{mygray}{gray}{.9}
\definecolor{mylinkcolor}{RGB}{115,194,251}

\usepackage{ulem}  

\definecolor{iccvblue}{rgb}{0.21,0.49,0.74}
\definecolor{myblue}{HTML}{486ED7}
\definecolor{yuangreen}{RGB}{0,204,112}

\usepackage{marvosym}

\definecolor{deepred}{RGB}{152, 1, 0}
\definecolor{grey1}{RGB}{96, 101, 102}
\hypersetup{
    colorlinks=true,
    linkcolor=yuangreen,
    filecolor=magenta,
    urlcolor=yuangreen,
    citecolor=yuangreen,
}

\tcbset{
    challenges/.style={
    colback=gray!5!white, 
        colframe=white, 
        boxrule=0.5mm, 
        left=1mm, 
        right=1mm, 
        top=1mm, 
        bottom=1mm, 
        arc=3mm, 
        boxsep=3mm,
        drop shadow={black!50!white}, 
        enhanced,
        overlay={
            \node[fill=cyan!70!black, text=white, rounded corners=1mm, font=\bfseries\scriptsize, inner sep=1mm] 
            at (frame.north west) 
            [xshift=8mm, yshift=-0mm] {Challenges};
            }
    }
}

\tcbset{
    insights/.style={
    colback=gray!5!white, 
        colframe=white, 
        boxrule=0.5mm, 
        left=1mm, 
        right=1mm, 
        top=1mm, 
        bottom=1mm, 
        arc=3mm, 
        boxsep=3mm,
        drop shadow={black!50!white}, 
        enhanced,
        overlay={
            \node[fill=orange!70!black, text=white, rounded corners=1mm, font=\bfseries\scriptsize, inner sep=1mm] 
            at (frame.north west) 
            [xshift=8mm, yshift=-0mm] {Insights};
            }
    }
}

\usepackage{graphicx}
\usepackage{caption}
\usepackage{subcaption}

\usepackage{color,xcolor}
\usepackage{epsfig}
\usepackage{inconsolata}
\usepackage{graphicx}
\usepackage{microtype}
\usepackage{bbm}
\usepackage{bm}
\usepackage{graphicx}
\usepackage{subcaption}
\usepackage{adjustbox}
\usepackage{multirow}

\usepackage{tikz}
\usetikzlibrary{arrows}

\usepackage{listings}
\usepackage{esdiff}
\usepackage{lastpage}
\usepackage{pgf}

\usepackage{pifont}
\usepackage{tabularx}
\usepackage{adjustbox}
\usepackage{array}
\newcolumntype{H}{>{\setbox0=\hbox\bgroup}c<{\egroup}@{}}
\usepackage{booktabs}
\usepackage{colortbl}
\usepackage{float}
\usepackage{wrapfig}
\usepackage{hhline}
\usepackage{multirow}
\usepackage{numprint}
\usepackage{makecell}

\usepackage{amsmath,amsfonts,amsthm,amssymb}
\usepackage{mathtools}
\usepackage{bm}
\usepackage{nicefrac}
\usepackage{dsfont}

\usepackage{amsmath} 
\usepackage{amssymb} 

\usepackage{enumitem}
\setlist{nosep}

\usepackage{tocloft}

\usepackage{bbding}   

\usepackage{titlesec}

\usepackage[page]{appendix}
\usepackage[hang,flushmargin]{footmisc}

\newtcbox{\hlredtabA}{on line, box align=base, colback=red!10, colframe=white, size=fbox, arc=3pt, 
  before upper=\strut, top=-2pt, bottom=-4pt, left=-2pt, right=-2pt, boxrule=0pt}

\newtcbox{\hlredtabB}{on line, box align=base, colback=green!10, colframe=white, size=fbox, arc=3pt, 
  before upper=\strut, top=-2pt, bottom=-4pt, left=-2pt, right=-2pt, boxrule=0pt}

\newtcbox{\hlredtabC}{on line, box align=base, colback=yellow!50, colframe=white, size=fbox, arc=3pt, 
  before upper=\strut, top=-2pt, bottom=-4pt, left=-2pt, right=-2pt, boxrule=0pt}

\newtcbox{\hlredtabDA}{on line, box align=base, colback=red!10, colframe=white, size=fbox, arc=3pt, 
  before upper=\strut, top=-2pt, bottom=-4pt, left=-2pt, right=-2pt, boxrule=0pt} 
\newtcbox{\hlredtabDB}{on line, box align=base, colback=red!20, colframe=white, size=fbox, arc=3pt, 
  before upper=\strut, top=-2pt, bottom=-4pt, left=-2pt, right=-2pt, boxrule=0pt} 
\newtcbox{\hlredtabDC}{on line, box align=base, colback=red!30, colframe=white, size=fbox, arc=3pt, 
  before upper=\strut, top=-2pt, bottom=-4pt, left=-2pt, right=-2pt, boxrule=0pt} 
\newtcbox{\hlredtabDD}{on line, box align=base, colback=red!40, colframe=white, size=fbox, arc=3pt, 
  before upper=\strut, top=-2pt, bottom=-4pt, left=-2pt, right=-2pt, boxrule=0pt} 
\newtcbox{\hlredtabDE}{on line, box align=base, colback=red!50, colframe=white, size=fbox, arc=3pt, 
  before upper=\strut, top=-2pt, bottom=-4pt, left=-2pt, right=-2pt, boxrule=0pt} 
\newtcbox{\hlredtabDF}{on line, box align=base, colback=red!60, colframe=white, size=fbox, arc=3pt, 
  before upper=\strut, top=-2pt, bottom=-4pt, left=-2pt, right=-2pt, boxrule=0pt} 

\newcommand{\dabA}[1]{\hlredtabDA{\normalsize \dalgshifted{#1}}} 
\newcommand{\dabB}[1]{\hlredtabDB{\normalsize \dalgshifted{#1}}} 
\newcommand{\dabC}[1]{\hlredtabDC{\normalsize \dalgshifted{#1}}} 
\newcommand{\dabD}[1]{\hlredtabDD{\normalsize \dalgshifted{#1}}} 
\newcommand{\dabE}[1]{\hlredtabDE{\normalsize \dalgshifted{#1}}} 
\newcommand{\dabF}[1]{\hlredtabDF{\normalsize \dalgshifted{#1}}} 

\newcommand{\ColorValue}[1]{
    \pgfmathparse{#1}
    \ifdim\pgfmathresult pt>50pt
        \dabF{#1\%}
    \else
        \ifdim\pgfmathresult pt>40pt
            \dabE{#1\%}
        \else
            \ifdim\pgfmathresult pt>30pt
                \dabD{#1\%}
            \else
                \ifdim\pgfmathresult pt>20pt
                    \dabC{#1\%}
                \else
                    \ifdim\pgfmathresult pt>10pt
                        \dabB{#1\%}
                    \else
                        \ifdim\pgfmathresult pt>0pt
                            \dabA{#1\%}
                        \fi
                    \fi
                \fi
            \fi
        \fi
    \fi
}

\newcommand{\checkicon}{\raisebox{-.25em}{\includegraphics[width=1em]{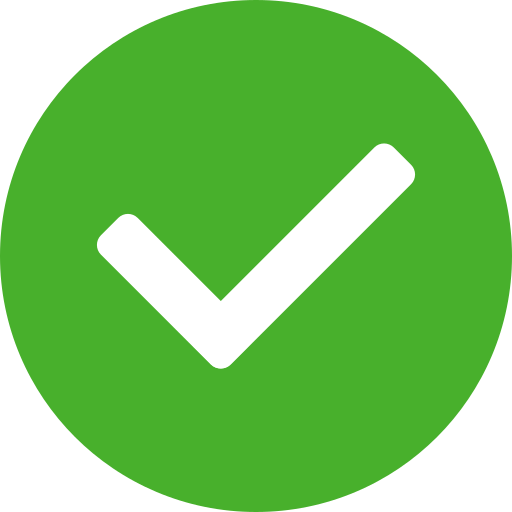}}}
\newcommand{\crossicon}{\raisebox{-.25em}{\includegraphics[width=1em]{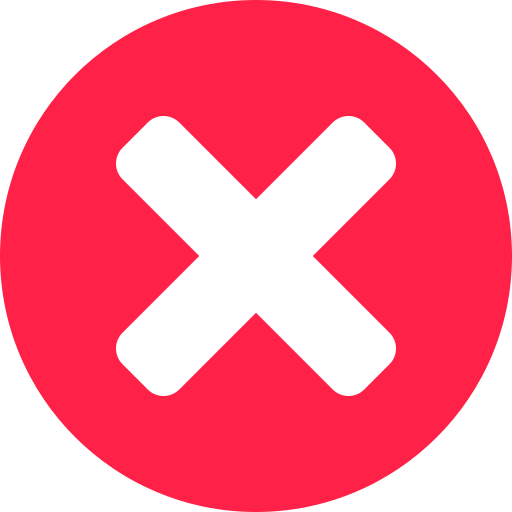}}}

\usepackage{graphicx}

\usepackage{amsmath, amssymb}
\usepackage{booktabs}
\usepackage{multirow}

\usepackage{algorithm}
\usepackage{algpseudocode}
\usepackage{pifont} 

\usepackage{bbm}
\usepackage{tikz}
\newlength{\barwidth}
\usepackage{lipsum}

\newcommand\blfootnote[1]{%
  \begingroup
  \renewcommand\thefootnote{}\footnote{#1}%
  \addtocounter{footnote}{-1}%
  \endgroup
}

\usepackage{hyperref}
\usepackage{CJKutf8}

\title{Beyond Relevance-Centric Retrieval: Rubric-Oriented Document Set Selection and Ranking}

\author{
Kailin Jiang\textsuperscript{1,{\color{yuangreen}*}} \quad Lei Liu\textsuperscript{2,{\color{yuangreen}\dag}} \quad Jian Xi\textsuperscript{1} \quad Hui Xu\textsuperscript{1} \quad Junlin Liu\textsuperscript{2}\\[6pt]
Baochen Fu\textsuperscript{2} \quad Bin Li\textsuperscript{2} \quad Vichwang\textsuperscript{1} \quad Yu Lu\textsuperscript{1,{\color{yuangreen}\dag}} \quad Haibo Shi\textsuperscript{1}\\[6pt]
\textsuperscript{1}Yuanbao Team, Tencent \quad \textsuperscript{2}Independent Researcher
}

\begin{document}

\maketitle
\blfootnote{\textsuperscript{{\color{yuangreen}*}}Work done during internship at Tencent Yuanbao. \textsuperscript{{\color{yuangreen}\dag}}Corresponding author: liulei13@ustc.edu.cn.}

\begin{abstract}
As large language models and AI agents become the primary consumers of search results, document set quality determines the upper bound of downstream generation. Yet existing evaluation systems remain confined to scoring documents independently and aggregating via nDCG, ignoring inter-document interactions (\eg redundancy, conflict, complementarity) and unable to answer what makes one document set better than another. To address these issues, we propose a complete evaluate-diagnose-optimize framework. We design \textbf{\dataset}, a three-level, nine-dimension document set evaluation benchmark covering both short-form and long-form scenarios, comprising approximately 28K high-quality evaluation rubrics. We systematically evaluate 12 rerankers: even the best method achieves no more than 45\% coverage, cross-document coordination dimensions are universally weak, and no single method maintains top performance across both settings. Building on this, we propose \textbf{\method}, a training-free method that converts rubric-based evaluation criteria into document set selection signals, achieving the best downstream generation performance with fewer documents and search rounds. It is the only method that maintains state-of-the-art results across both scenarios, validating the effectiveness of closing the loop from evaluation to optimization. 

\vspace{0.5em}
\noindent\raisebox{-0.3em}{\includegraphics[height=1em]{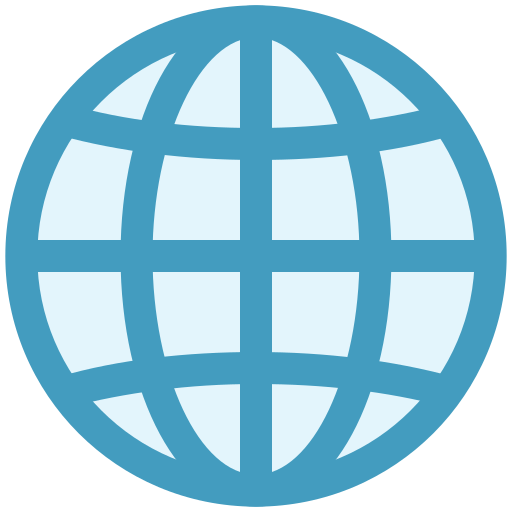}}~~\textbf{Project Page:} \href{https://rubric4setwise.github.io}{\texttt{https://rubric4setwise.github.io}}\\[0.3em]
\noindent\raisebox{-0.3em}{\includegraphics[height=1em]{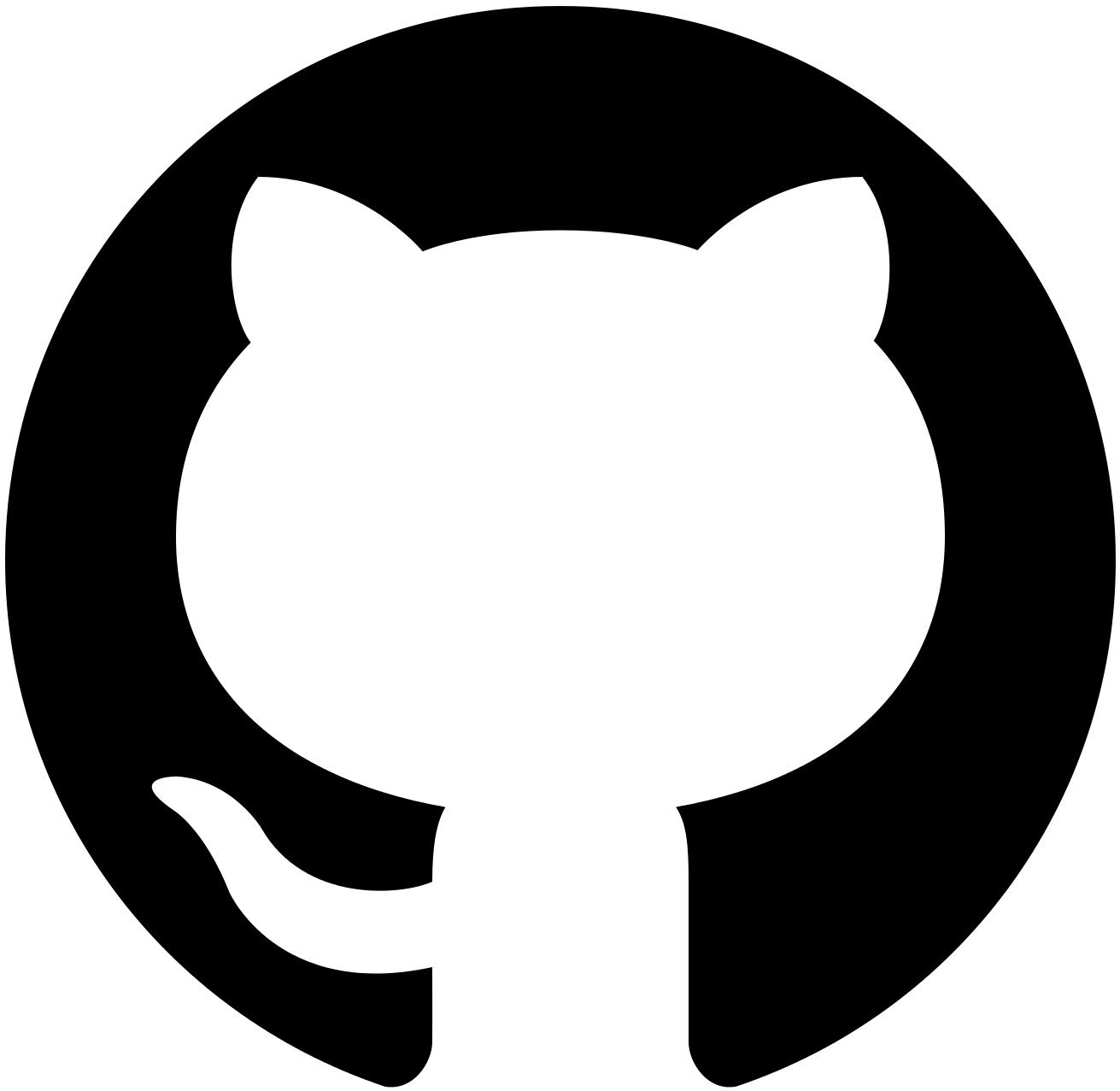}}~~\textbf{GitHub Repo:} \href{https://github.com/Rubric4Setwise/Rubric4Setwise}{\texttt{https://github.com/Rubric4Setwise/Rubric4Setwise}}\\[0.3em]
\noindent\raisebox{-0.3em}{\includegraphics[height=1em]{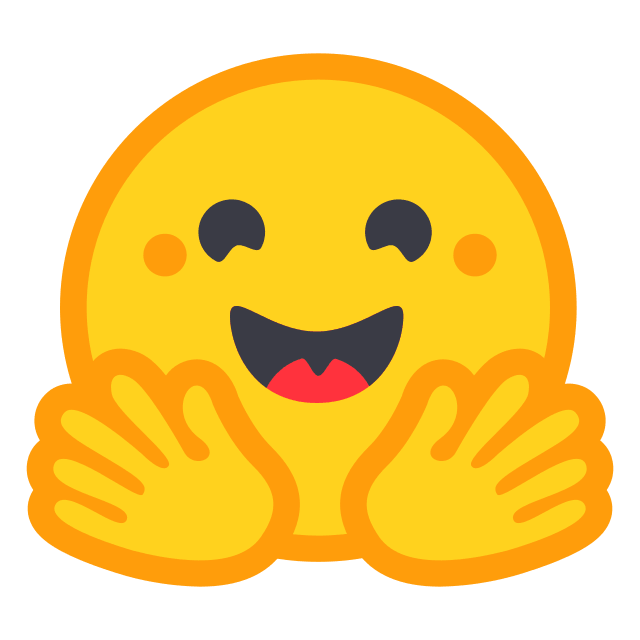}}~~\textbf{Dataset:} \href{https://huggingface.co/collections/placeholder}{\texttt{https://huggingface.co/datasets/kailinjiang/SetwiseEvalKit}}
\end{abstract}

\section{Introduction}

The role of information retrieval systems is undergoing a fundamental shift. In the traditional paradigm, search results are browsed by human users who possess the cognitive ability to filter redundancy, resolve conflicts, and identify gaps. In the Retrieval-Augmented Generation (RAG) paradigm~\citep{Lewis2020RetrievalAugmentedGF,Guu2020REALMRL,Huang2024ASO,jiang2025kore}, however, retrieved document sets are directly injected into the context window of large language models (LLMs) as the basis for reasoning and generation~\citep{RankGPT,Ma2023ZeroShotLD,Tang2023FoundIT,liu2026amo,liu2026general365,jiang2026mined}. This paradigm shift means that document set quality directly determines the upper bound of generation quality; the evaluation standard must accordingly evolve from ``useful to humans'' to ``effective for LLM generation''~\citep{Shi2023LargeLM,Liu2023LostIT,Cuconasu2024ThePO,jiang2026large,peng2025can,jia2026benchmarking,jiang2025mmke}.

\begin{wrapfigure}{r}{0.5\textwidth}
    \centering
    \vspace{-1em}
    \includegraphics[width=0.48\textwidth]{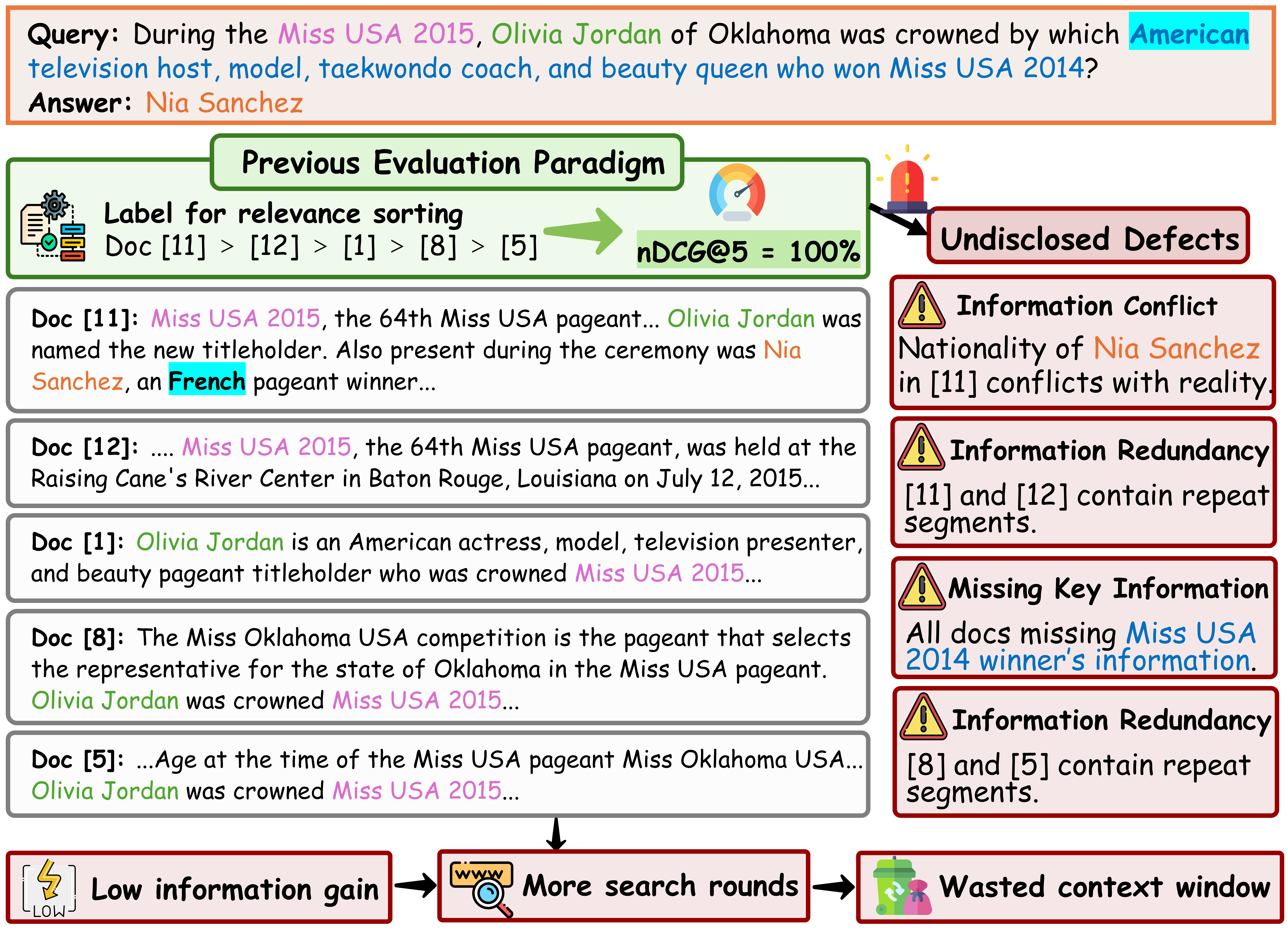} 
    \caption{\textbf{Limitations of previous evaluation.} Although all retrieved documents are relevant (nDCG@5 = 100\%), the set exhibits critical defects undetectable by traditional metrics: information conflict (Doc [11] contains factual errors), redundancy (Docs [11]\&[12], [8]\&[5] contain repeated content), and missing key information.}\label{fig:motivation}
\end{wrapfigure}

Under the constraint of limited context windows, relevance-only document selection exhibits structural deficiencies. Even when all retrieved documents are relevant to the query (nDCG@5 = 100\%), the selected set may still suffer from critical defects: highly overlapping documents waste precious context capacity with redundant information; factual conflicts between documents mislead the generator into producing incorrect answers; and missing key evidence forces agents to initiate additional search rounds (Figure~\ref{fig:motivation}). In contrast, a set-oriented perspective optimizes document combinations holistically, maximizing information coverage through complementarity, compressing the required document count through redundancy detection, and purifying input quality through conflict identification, ultimately achieving better generation with fewer documents and fewer search rounds.

However, from classic retrieval benchmarks such as TREC-DL, BEIR, and MTEB~\citep{TREC-DL,BEIR,MTEB}, to LLM-Judge-based evaluation methods such as RAGAS and ARES~\citep{RAGAS,ARES}, to the most recent RA-nWG and BRIGHT~\citep{RA-nWG,BRIGHT}, existing evaluation systems consistently exhibit two structural blind spots (Table~\ref{tab:metric_taxonomy}). \ding{182} \textit{The evaluation granularity is singular}: all methods score documents independently and aggregate via nDCG, implicitly assuming that set quality equals the sum of individual qualities while ignoring inter-document interactions. A set of five highly relevant but heavily overlapping documents achieves a perfect nDCG score, yet its actual value for downstream generation is far lower than a complementary combination. \ding{183} \textit{The evaluation dimension is singular}: all benchmarks cover only relevance, unable to answer \textit{why} one document set is better than another, whether due to stronger complementarity, lower redundancy, or more complete reasoning chain coverage. This gap means we can neither accurately diagnose the true bottlenecks of existing methods nor convert evaluation signals into optimization guidance.

To address these issues, we propose a complete evaluate-diagnose-optimize framework. We design \textbf{\dataset}, a three-level, nine-dimension document set evaluation benchmark covering both  Short-form and Long-form scenarios, with a query-specific rubric generation pipeline that produces approximately 28K high-quality hybrid rubrics. We systematically evaluate 12 rerankers: even the best method achieves no more than 45\% overall coverage, and rubric coverage scores are strongly predictive of downstream generation quality, confirming the benchmark's credibility. Building on this, we propose \textbf{\method}, a training-free method that converts rubric-based evaluation criteria into document set selection signals, achieving the best downstream generation performance with fewer documents and search rounds.

\textbf{Our contributions can be concluded as follows:}
\begin{itemize}[leftmargin=1.5em, itemsep=3pt, topsep=3pt]
\item  We contribute the first multi-dimensional evaluation framework for document set quality in the LLM era. \dataset defines a three-level, nine-dimension evaluation system covering both single-turn retrieval and multi-turn agent scenarios, with scores validated to be highly predictive of downstream generation quality, enabling reliable guidance for document set optimization.

\item We provide the systematic multi-dimensional diagnosis of 12 rerankers, exposing quality deficiencies invisible to traditional metrics, particularly a universal bottleneck on cross-document coordination dimensions, and identifying clear generalization ceilings of current methods.

\item We demonstrate that evaluation signals can directly serve as selection signals: \method achieves state-of-the-art downstream generation in both scenarios with fewer documents and search rounds, and is the only method maintaining top performance across both settings.

\end{itemize}

\section{Related Work}

\nb{Retrieval Evaluation Benchmarks.} Evaluating retrieval quality has long relied on document-level relevance scoring followed by metric aggregation. TREC-DL~\citep{TREC-DL}, BEIR~\citep{BEIR}, MTEB~\citep{MTEB}, and BRIGHT~\citep{BRIGHT} annotate per-document relevance and aggregate via nDCG, treating set quality as the sum of individual scores. RAGAS~\citep{RAGAS} and ARES~\citep{ARES} introduce LLM-based judges but still reduce to binary per-document relevance. RA-nWG~\citep{RA-nWG} proposes an order-free set metric, yet scores each document independently without assessing inter-document interactions. All existing benchmarks share two structural blind spots (Table~\ref{tab:metric_taxonomy}): \ding{182} evaluation at the document level only, ignoring inter-document relationships; \ding{183} coverage of relevance alone, leaving set-level properties like complementarity, redundancy, and conflict unexamined. Our work fills this gap with the first retrieval set evaluation framework spanning three granularity levels and nine dimensions.

\nb{LLM-based Document Reranking.} LLM-based rerankers have evolved along a trajectory from scoring documents independently to optimizing set composition~\citep{qi2025context,Fu2026CanML,Fu2026MMKUBenchAM,li2026knowhal}. Pointwise methods~\citep{BGE-Reranker,MonoT5} score each document in isolation. Listwise and Setwise methods~\citep{RankGPT,Setwise,Liu2024SlidingWA,Liu2024DemoRankSE,Liu2025CoRankingCR,Feng2026SumRankAS} compare multiple documents simultaneously. Reasoning-enhanced methods~\citep{Rank-R1,ReasonRank} incorporate chain-of-thought reasoning into ranking decisions. Most recently, set-oriented methods~\citep{SetR,Rank4Gen,jiang2026adatutorank} directly optimize document subsets for downstream generation. Despite this methodological progression from individual to set-level selection, the evaluation paradigm has not kept pace.

\nb{Rubric-based Evaluation.} In educational assessment and writing evaluation, rubrics have long served as a standard tool for transparent, reproducible, and multi-dimensional quality judgment~\citep{liu2026rules}. The rise of LLM-as-Judge has brought rubrics into NLP: MT-Bench~\citep{MT-Bench} uses GPT-4 to assess dialogue quality across multiple dimensions; RubricEval~\citep{RubricEval} shows that structured rubrics with cross-arbitration among frontier LLMs improve consistency and reduce position bias; ResearchQA~\citep{ResearchQA} combines rubric coverage with direct judgment to score academic QA per knowledge point. These works validate rubric-guided LLM judges for generation evaluation, yet none applies rubrics to retrieval set quality.

We transfer rubrics from answer evaluation to set evaluation, designing structured criteria for each of the nine dimensions. Beyond evaluation, \method demonstrates that the same rubrics can serve as a training-free set selection signal, bridging the gap from passive assessment to active guidance.

\begin{table*}[t!]
\caption{\textbf{Overall comparison with existing retrieval evaluation benchmarks.} \ding{182} Target: Document-level (Doc) or Set-level (Set). \ding{183} Dimension abbreviations: Relevance (Rel.), Authenticity (Aut.), Quality (Qua.), Complementarity (Cmp.), Redundancy (Red.), Conflict (Con.), Completeness (Cov.), Density (Den.), and Reachability (Rea.). }
\label{tab:metric_taxonomy}
\vspace{-5pt}
  \centering
  \renewcommand{\arraystretch}{1.2}  
  \resizebox{\textwidth}{!}{%
    \begin{tabular}{l|cc|cc|ccc|ccc|ccc|c}
        \toprule
        \multirow{2}{*}[-3pt]{\textbf{Benchmark}}
          & \multicolumn{2}{c|}{\textbf{Target}}
          & \multicolumn{2}{c|}{\textbf{Round}}
          & \multicolumn{9}{c|}{\textbf{Dimension}}
          & \multirow{2}{*}[-3pt]{\textbf{Metric}} \\
        \cmidrule(l{2pt}r{2pt}){2-3} \cmidrule(l{2pt}r{2pt}){4-5} \cmidrule(l{2pt}r{2pt}){6-14}
          & \textbf{Doc} & \textbf{Set}
          & \textbf{Single} & \textbf{Multi}
          & \textbf{Rel.} & \textbf{Aut.} & \textbf{Qua.} 
          & \textbf{Cmp.} & \textbf{Red.} & \textbf{Con.}
          & \textbf{Cov.} & \textbf{Den.} & \textbf{Rea.}
          & \\
        \midrule

        \textbf{TREC-DL}
        ~\citep{TREC-DL}
         & \checkicon & \crossicon & \checkicon & \crossicon & \checkicon & \crossicon & \crossicon & \crossicon & \crossicon & \crossicon & \crossicon & \crossicon & \crossicon
         & NDCG\\

        \textbf{BEIR}
        ~\citep{BEIR}
         & \checkicon & \crossicon & \checkicon & \crossicon & \checkicon & \crossicon & \crossicon & \crossicon & \crossicon & \crossicon & \crossicon & \crossicon & \crossicon
         & NDCG\\

        \textbf{MTEB}
        ~\citep{MTEB}
         & \checkicon & \crossicon & \checkicon & \crossicon & \checkicon & \crossicon & \crossicon & \crossicon & \crossicon & \crossicon & \crossicon & \crossicon & \crossicon
         & NDCG\\

                  \textbf{RAGAS}
        ~\citep{RAGAS}
         & \checkicon & \crossicon & \checkicon & \crossicon & \checkicon & \crossicon & \crossicon & \crossicon & \crossicon & \crossicon & \checkicon & \crossicon & \crossicon
         & LLM-Judge \\

         \textbf{ARES}
        ~\citep{ARES}
         & \checkicon & \crossicon & \checkicon & \crossicon & \checkicon & \crossicon & \crossicon & \crossicon & \crossicon & \crossicon & \crossicon & \crossicon & \crossicon
         & LLM-Judge \\

\textbf{BRIGHT}
        ~\citep{BRIGHT}
         & \checkicon & \crossicon & \checkicon & \crossicon & \checkicon & \crossicon & \crossicon & \crossicon & \crossicon & \crossicon & \crossicon & \crossicon & \crossicon
         & NDCG\\

\textbf{RA-nWG}
        ~\citep{RA-nWG}
         & \checkicon & \crossicon & \checkicon & \crossicon & \checkicon & \crossicon & \crossicon & \crossicon & \checkicon & \crossicon & \checkicon & \crossicon & \crossicon
         & NDCG \\

        \midrule

        \textbf{\dataset} {\small(Ours)} & \checkicon & \checkicon & \checkicon & \checkicon & \checkicon & \checkicon & \checkicon & \checkicon & \checkicon & \checkicon & \checkicon & \checkicon & \checkicon
         & Rubric-Judge\\

        \bottomrule
    \end{tabular}
    }

\end{table*}

\section{Rubrics-Oriented Evaluation and Set Selection}

\subsection{Task Formulation}

Given a user query $q$ and the corresponding reference answer $a$, we first construct a set of query-specific evaluation rubrics $\mathcal{R} = \{r_1, r_2, \ldots, r_K\}$ based on the pair $(q, a)$. Each rubric defines a concrete evaluation criterion, collectively covering nine quality dimensions across the doc, set, and global-level.

Each reranker method selects a document subset $S_m \subseteq \mathcal{C}$ from the candidate pool $\mathcal{C}$, subject to $|S_m| \leq k$. Traditional evaluation computes a relevance score for each document in $S_m$ independently and then aggregates them via weighted summation, implicitly assuming that set quality equals the sum of individual document qualities. This assumption fails to capture inter-document interactions.

We instead evaluate $S_m$ as a whole using the rubrics $\mathcal{R}$ generated from $(q, a)$, producing a multi-dimensional score $\text{Score}(S_m) = \mathcal{E}(q, S_m, \mathcal{R}) \in \mathbb{R}^9$, where $\mathcal{E}$ denotes the LLM-based evaluator that scores the document set along each dimension after comprehensive reading. Since the rubrics are jointly determined by the query and the reference answer, the evaluation criteria precisely reflect the quality of document selection under the given information need, capturing characteristics such as redundancy and conflict.

\subsection{\dataset}

\nb{Data Collection.} \textbf{Short-form Scenario:} Multi-hop question answering inherently requires synthesizing information from multiple documents, making it well suited for evaluating document set quality. We construct this scenario from four multi-hop QA datasets: HotpotQA~\citep{HotpotQA}, 2WikiMultihopQA~\citep{2WikiMultihopQA}, MuSiQue~\citep{MuSiQue}, and Bamboogle~\citep{Bamboogle}, sampling 2,000 instances from each of the first three and retaining all 125 from Bamboogle. For each query, BM25~\citep{BM25} retrieves the top-20 documents from a Wikipedia corpus as the candidate pool $\mathcal{C}$, from which multiple rerankers each select a subset $S_m$. After filtering queries where BM25 fails to recall sufficiently relevant documents, we retain 2,061 high-quality samples.

\begin{wrapfigure}{r}{0.5\textwidth}
    \centering
    \vspace{-1em}
    \includegraphics[width=0.48\textwidth]{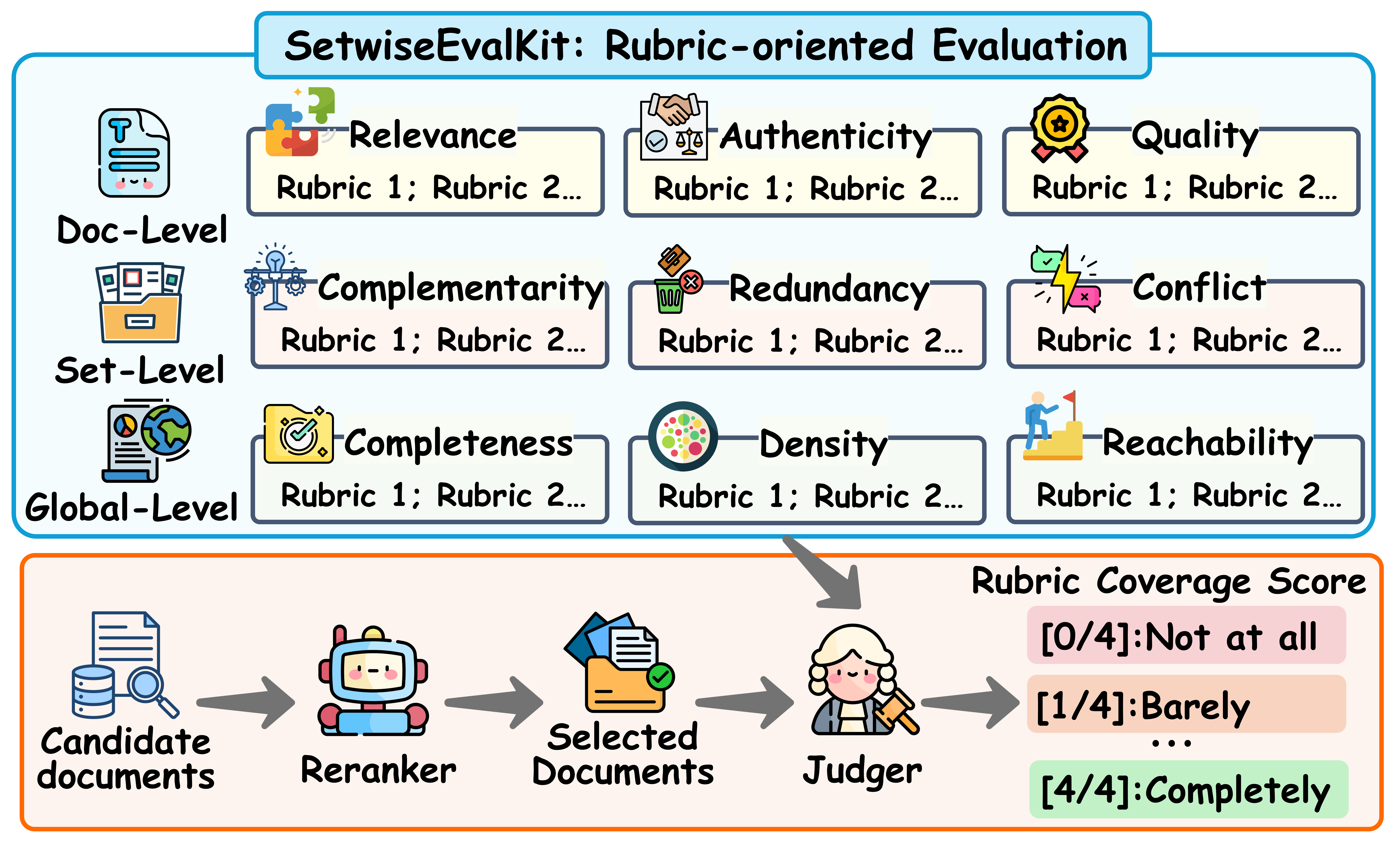} 
    \caption{Overview of \dataset's evaluation. \ding{182} Three-level, nine-dimension rubric taxonomy for document set evaluation. \ding{183} Evaluation pipeline where a reranker selects documents from the candidate pool and a judge scores the selected set against query-specific rubrics on a 0–4 scale.}\label{fig:overview}
\end{wrapfigure}

\textbf{Long-form Scenario:} We randomly select 200 instances from ResearchQA~\citep{ResearchQA} and use the open-source search agent DR.Tulu-8B~\citep{DR-Tulu} to extract multi-turn search trajectories. At each turn, the agent calls the Google Search API to retrieve candidate documents. \textbf{We wrap all rerankers as tools via the MCP protocol} and embed them into the pipeline, applying reranking after each retrieval turn.

\nb{Multi-dimensional Rubric Design.} Traditional evaluation scores each document in $S_m$ independently for relevance, failing to capture inter-document relationships or holistic reasoning support. We instead design a rubric system spanning three levels and nine dimensions, characterizing set quality from progressively broader perspectives: individual document attributes, inter-document relationships, and holistic set utility.

\noindent \textbf{Level 1: Document-level.} Evaluates the independent attributes of each document individually.

\begin{itemize}[leftmargin=*]
  \item \textbf{Relevance {\small(Rel.)}} Whether the document semantically addresses the core information need beyond surface keyword matching.
  \item \textbf{Authenticity {\small(Aut.)}} Whether the stated facts are consistent with the correct reference answer.
  \item \textbf{Quality {\small(Qua.)}}Whether the document is well-structured and topically focused, allowing direct extraction of target information.
\end{itemize}

\noindent \textbf{Level 2: Set-level.} Evaluates synergy and conflict effects within the set from the perspective of inter-document relationships.

\begin{itemize}[leftmargin=*]
  \item \textbf{Complementarity {\small(Cmp.)}} Whether documents collectively cover all key information elements required by the answer, constructing a more complete response than any single document.
  \item \textbf{Redundancy {\small(Red.)}} Whether multiple documents convey essentially the same information without incremental contribution. 
  \item \textbf{Conflict {\small(Con.)}} Whether different documents provide contradictory statements about the same fact.
\end{itemize}

\noindent \textbf{Level 3: Global-level.} Treats the set as a whole and evaluates its overall utility when serving as the input context for an LLM.

\begin{itemize}[leftmargin=*]
  \item \textbf{Completeness {\small(Cov.)}} Whether the set fully answers the query with every key information element present.
  \item \textbf{Density {\small(Den.)}} Whether content useful for answering constitutes the majority of each document, rather than being dominated by irrelevant filler. 
  \item \textbf{Reachability {\small(Rea.)}} Whether a model with no external knowledge can complete the full reasoning chain to produce the correct answer solely from this set.
\end{itemize}

\nb{Hybrid Rubrics Generation.} To obtain query-specific evaluation rubrics, we adopt a multi-model generation and aggregation strategy. For each pair $(q, a)$, two frontier LLMs (\ie GPT 5.1 and Gemini 3.1-Pro-Preview) independently generate candidate rubrics. Each rubric takes the form of an evaluation question targeting document set quality and must explicitly reference the specific entities, facts, or numerical values in the query and answer; vague phrasing such as ``relevant content'' is prohibited. DeepSeek-V4 Pro then aggregates and filters candidates from both sources, removing duplicates and low-quality items. This yields approximately 24K rubrics for short-form and 4K for long-form.

\begin{figure*}[t!]
  \centering
\includegraphics[width=1\linewidth]{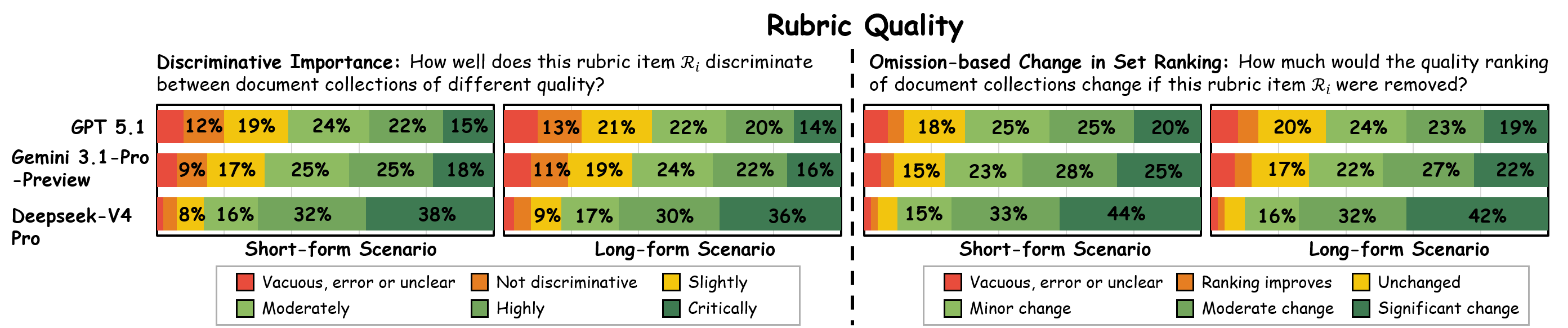}
  \caption{\dataset rubric quality ratings by Ph.D. level experts.}
  \label{fig:human_study}
\end{figure*}

\nb{Rubric Quality.} To verify rubric quality, we randomly select 50 samples per scenario and conduct a human study (Figure~\ref{fig:human_study}) evaluating discriminative power and removal impact. After DeepSeek-V4 Pro aggregation, approximately 70\% of rubrics are judged Highly or Critically discriminative, outperforming GPT 5.1 (37\%) and Gemini 3.1-Pro-Preview (43\%), with a waste rate of only about 8\%. Furthermore, approximately 77\% of aggregated rubrics cause Moderate or Significant ranking changes when removed, compared to 45\% for GPT 5.1 and 53\% for Gemini 3.1-Pro-Preview. Consistent trends across both scenarios confirm that multi-model generation with aggregation effectively eliminates low-quality candidates and produces highly discriminative rubrics.

\subsection{\method}

The rubric system above provides a systematic evaluation framework for document set quality. A natural corollary is that \textbf{\textit{if rubrics can quantify set quality, they can equally serve as a guiding signal for selection}}. Based on this insight, we propose \method, a training-free method that directly converts evaluation criteria into selection guidance.

Formally, given a query $q$, query-specific rubrics $\mathcal{R}$, and candidate pool $\mathcal{C}$, \method selects an optimal subset $S^* = \arg\max_{S \subseteq \mathcal{C}} f(S;\, q, \mathcal{R})$, where $f(\cdot)$ is a rubric-based set utility function and $|S^*|$ is adaptively determined by rubric satisfaction rather than fixed in advance. We use Qwen3-8B~\citep{Qwen3} as the reasoning backbone, implicitly modeling $f$ through chain-of-thought prompting.

\section{Experiments}

\subsection{\label{sec:Setup} Experimental Setup}

\noindent \textbf{Reranker.} We evaluate 12 mainstream rerankers on \dataset. \textbf{(1) Adhoc reranking}, including the cross-encoder BGE-Reranker-Large~\citep{BGE-Reranker}, the sequence-to-sequence models MonoT5~\citep{MonoT5} and RankT5~\citep{RankT5}, and the LLM-based listwise methods RankLlama~\citep{RankLlama}, RankVicuna~\citep{RankVicuna}, RankZephyr~\citep{RankZephyr}, and Setwise~\citep{Setwise}. \textbf{(2) Reasoning-enhanced reranking}, including Rank1~\citep{Rank-R1}, Rearank~\citep{Rearank}, and ReasonRank~\citep{ReasonRank}. \textbf{(3) Setwise reranking}, including SetR~\citep{SetR} and Rank4Gen~\citep{Rank4Gen}. We additionally report raw BM25 retrieval (\ie Only Retrieval) and Google Search (\ie Only Search) results without any reranking as lower-bound baselines.

\noindent \textbf{Evaluation Protocol.} We adopt the LLM-as-Judge paradigm for automatic evaluation. For each query-document set pair $(q, S_m)$, we feed the query-specific rubrics $\mathcal{R}$ along with the document set $S_m$ into DeepSeek-V4 Pro, which produces scores across all dimensions in a single pass. For the long-form scenario, documents selected by the reranker across all search rounds are aggregated and deduplicated to form the input set for scoring.

The implementation details are elaborated below.  \ding{182} Grading Criteria: Each rubric is scored on a 0--4 scale, with higher scores indicating better performance. Doc-level dimensions are scored per document individually, while set-level and global-level dimensions are scored on the document set as a whole. \ding{183} Coverage \%: We average the scores of all rubrics within the same dimension, divide by 4, and multiply by 100\% to obtain the coverage rate for that dimension: $\text{Cov}_d = \frac{1}{|\mathcal{R}_d|} \sum_{r \in \mathcal{R}_d} \frac{\text{score}(r)}{4} \times 100\%.$ \ding{184} Relevance Gating Mechanism: If every document in the set receives a relevance score of 0, only the Relevance score is reported and the remaining eight dimensions are skipped and excluded from aggregation. This prevents noise from evaluating set-level and global-level properties on entirely irrelevant documents. \ding{185} Stability of Judge: We run two independent scoring passes per sample and report the mean and standard deviation of coverage rates in Table~\ref{tab:eval_short} and~\ref{tab:eval_long_union}.

\subsection{Results on Rubric-Oriented Evaluation}

\nb{Benchmark Credibility.} Before analyzing the results of \dataset, a prerequisite must be verified: \textit{whether the rubric coverage scores genuinely reflect document set quality} (\ie whether document sets with higher evaluation scores indeed lead to better downstream generation performance). If the two are inconsistent, all subsequent analyses based on evaluation scores would lack practical significance. 

\begin{wrapfigure}{r}{0.5\textwidth}
    \centering
    
    \includegraphics[width=0.48\textwidth]{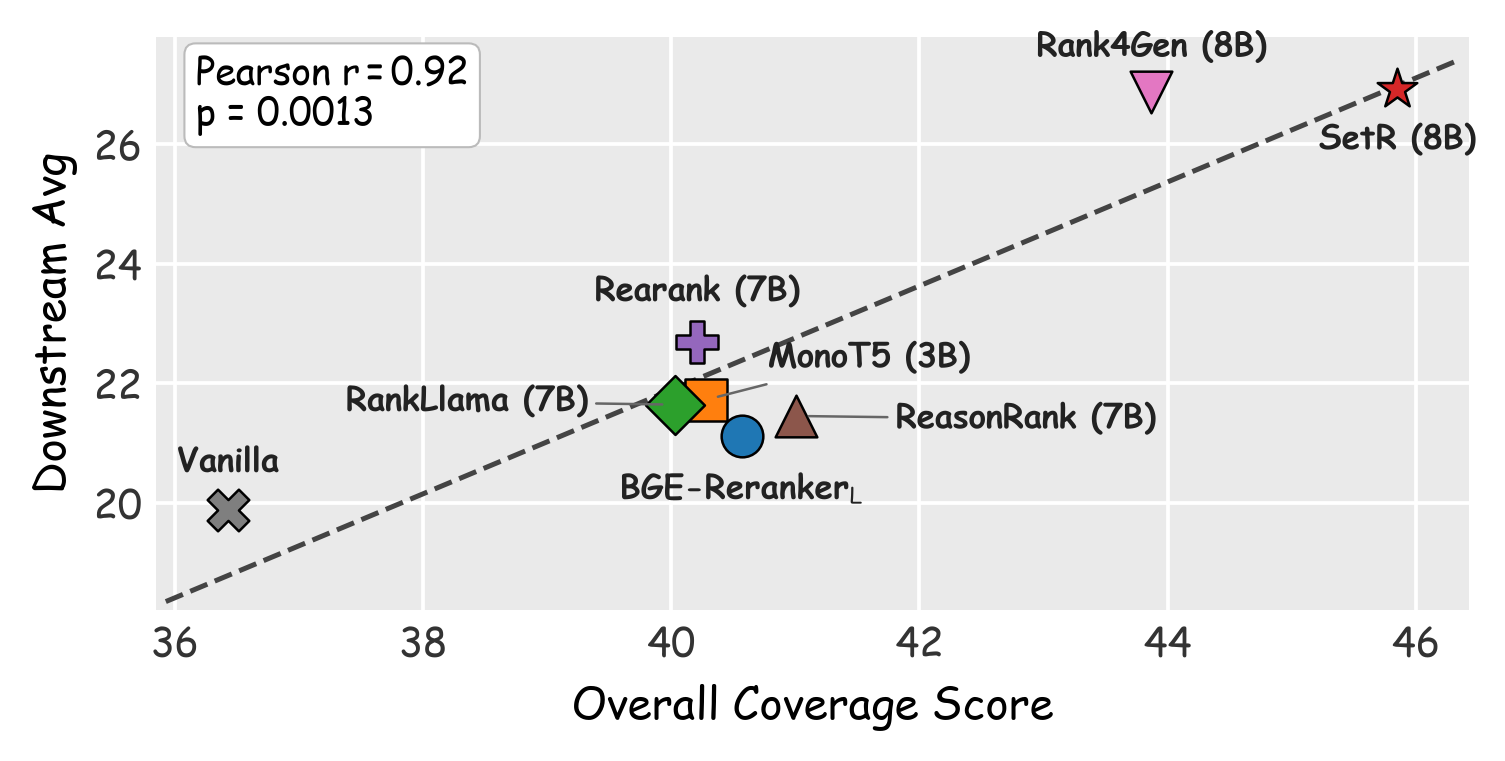}
    \vspace{-5pt}
    \caption{\textbf{Correlation between rubric coverage and downstream generation performance.} Each point represents a reranker method. The strong linear correlation (Pearson $r = 0.92$) demonstrates that \dataset's evaluation scores are highly predictive of actual generation quality.}
    \label{fig:coverage_vs_answer_consistency}
\end{wrapfigure}

To validate this prerequisite, we conduct a consistency analysis between the overall coverage scores evaluated on \dataset and the corresponding downstream generation performance across different rerankers. In Figure \ref{fig:coverage_vs_answer_consistency}, the two exhibit a strong positive correlation with Pearson $r = 0.92$ and $p = 0.0013$, confirming that methods achieving higher rubric coverage consistently yield higher-quality generated answers.

\begin{figure*}[t!]
  \centering
\includegraphics[width=1\linewidth]{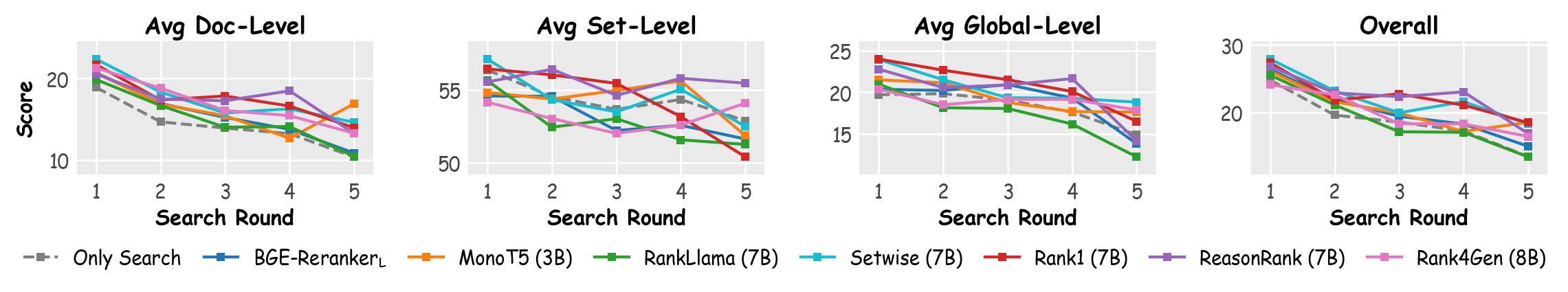}
  \caption{\textbf{Per-round Rubric Coverage Scores in the Long-form Scenario.} We independently evaluate the document set selected by each reranker at each search round.}
  \label{fig:long_form_round}
\end{figure*}

With this credibility established, we summarize key observations on method-level behavior (Table~\ref{tab:eval_short}, \ref{tab:eval_long_union}; Figure~\ref{fig:long_form_round}):

\nb{Behavior Analysis in Short-form Scenario.} \ding{182} \textit{Setwise reranking methods lead by a clear margin on Overall.} SetR and Rank4Gen reach $45.85$ and $43.87$, ahead of the strongest Adhoc and Reasoning-Enhanced methods, indicating that explicitly optimizing set composition within a fixed candidate pool is the most effective strategy. \ding{183} \textit{Different paradigms exhibit distinct advantage distributions across the three levels.} Setwise methods lead at the Doc-Level, where SetR reaches $33.44$ against $20.30$ for the strongest Adhoc method, and at the Set-Level, where Rank4Gen reaches $67.65$ driven by a Redundancy score of $80.41$, demonstrating that set-granularity selection ensures individual document quality while simultaneously optimizing inter-document relationships. Reasoning-Enhanced methods narrow the gap at the Global-Level, where ReasonRank attains $38.82$, above Rank4Gen's $38.12$ and within $1.26$ of SetR, suggesting that reasoning capabilities help construct sets that support complete reasoning chains. Adhoc methods remain the weakest at every level, their best Doc-, Set-, and Global-Level averages reaching only $20.30$, $63.17$, and $38.34$, exposing the inherent limitations of the conventional ranking paradigm.  \ding{184} \textit{Conflict scores remain uniformly high with minimal variance across methods.} All methods fall within $92.61$ to $94.54$ on the Conflict dimension, indicating that factual contradiction is not a primary bottleneck in the short-form multi-hop QA scenario.

\begin{table*}[t!]
  \centering
    \caption{\textbf{Overall Performance Comparison ($\%$) on \dataset{} (Short-form Scenario).}
The top two performing results are highlighted in
\colorbox{backred!50}{red} (1\textsuperscript{st}) and
\colorbox{backyellow_soft!40}{yellow} (2\textsuperscript{nd}) backgrounds, respectively. All metrics are reported such that higher is better (\daugshifted).
Only Retrieval (BM25), Adhoc Reranking, and Reasoning-Enhanced Reranking use a \textbf{fixed top-5 documents}; Setwise Reranking \textbf{adaptively selects the number of documents}.}
  \label{tab:eval_short}

  \vspace{-5pt}
  \renewcommand{\arraystretch}{1.1}
  \newcommand{\std}[1]{$_{(\pm#1)}$}
  \resizebox{\textwidth}{!}{%
    \begin{tabular}{l|cccc|cccc|cccc|c}
      \toprule
      \multirow{2.5}{*}{\textbf{Ranker}}
        & \multicolumn{4}{c|}{\textbf{Doc-Level}}
        & \multicolumn{4}{c|}{\textbf{Set-Level}}
        & \multicolumn{4}{c|}{\textbf{Global-Level}}
        & \cellcolor{gray!12} \\
      \noalign{\global\aboverulesep=0pt \global\belowrulesep=0pt}
      \cmidrule(l{2pt}r{2pt}){2-5} \cmidrule(l{2pt}r{2pt}){6-9} \cmidrule(l{2pt}r{2pt}){10-13}
      \noalign{\global\aboverulesep=.65ex \global\belowrulesep=.65ex}
        &
        \cellcolor{gray!12}\textbf{Avg} & \textbf{Rel.} & \textbf{Aut.} & \textbf{Qua.} 
        & \cellcolor{gray!12}\textbf{Avg} & \textbf{Cmp.} & \textbf{Red.} & \textbf{Con.} 
        & \cellcolor{gray!12}\textbf{Avg} & \textbf{Cov.} & \textbf{Den.} & \textbf{Rea.} 
        & \cellcolor{gray!12}\raisebox{4.5pt}[0pt][0pt]{\textbf{Overall}} \\
      \midrule


      Only Retrieval& \cellcolor{gray!12}16.30 & 14.67 & 17.18 & 17.05 & \cellcolor{gray!12}60.91 & 18.29 & 71.57 & 92.88 & \cellcolor{gray!12}32.08 & 46.59 & 20.05 & 29.60 & \cellcolor{gray!12}36.43\std{0.84} \\

      \midrule

      \rowcolor{gray!10}
      \multicolumn{14}{c}{\fontsize{10}{12}\selectfont \textit{\textbf{Adhoc Reranking}}} \\

      BGE-Reranker$_{L.}$& \cellcolor{gray!12}20.30 & 18.89 & 21.32 & 20.70 & \cellcolor{gray!12}63.05 & \colorbox{backred!50}{28.04} & 67.86 & 93.26 & \cellcolor{gray!12}38.34 & 54.35 & 21.03 & 39.64 & \cellcolor{gray!12}40.57\std{0.96} \\

      MonoT5 {\small(3B)}& \cellcolor{gray!12}19.90 & 19.00 & 20.35 & 20.36 & \cellcolor{gray!12}62.76 & 26.92 & 68.74 & 92.61 & \cellcolor{gray!12}38.17 & 53.97 & 21.29 & 39.26 & \cellcolor{gray!12}40.28\std{0.95} \\

      RankT5 {\small(3B)}& \cellcolor{gray!12}20.12 & 19.47 & 20.43 & 20.47 & \cellcolor{gray!12}62.86 & 26.86 & 68.61 & 93.12 & \cellcolor{gray!12}37.97 & 54.08 & 21.17 & 38.67 & \cellcolor{gray!12}40.32\std{0.95} \\

      RankLlama {\small(7B)}& \cellcolor{gray!12}19.63 & 19.13 & 19.94 & 19.81 & \cellcolor{gray!12}63.08 & 26.07 & 69.51 & 93.66 & \cellcolor{gray!12}37.38 & 53.46 & 20.75 & 37.94 & \cellcolor{gray!12}40.03\std{0.94} \\

      RankVicuna {\small(7B)}& \cellcolor{gray!12}18.59 & 17.44 & 19.35 & 18.97 & \cellcolor{gray!12}62.42 & 23.45 & 69.72 & 94.11 & \cellcolor{gray!12}35.66 & 50.88 & 20.88 & 35.20 & \cellcolor{gray!12}38.89\std{0.91} \\

      RankZephyr {\small(7B)}& \cellcolor{gray!12}18.93 & 17.96 & 19.78 & 19.06 & \cellcolor{gray!12}61.45 & 22.35 & 68.35 & 93.65 & \cellcolor{gray!12}35.35 & 50.27 & 20.75 & 35.03 & \cellcolor{gray!12}38.58\std{0.90} \\

      {Setwise {\small(7B)}}& \cellcolor{gray!12}19.52 & 18.81 & 19.94 & 19.81 & \cellcolor{gray!12}63.17 & 26.27 & 69.78 & 93.45 & \cellcolor{gray!12}37.51 & 53.20 & 20.89 & 38.46 & \cellcolor{gray!12}40.07\std{0.95} \\

      \midrule
      \rowcolor{gray!10}
      \multicolumn{14}{c}{\fontsize{10}{12}\selectfont \textit{\textbf{Reasoning-Enhanced Reranking}}} \\

      {Rank1 {\small(7B)}}& \cellcolor{gray!12}19.86 & 19.27 & 20.21 & 20.09 & \cellcolor{gray!12}62.93 & 25.90 & 70.03 & 92.88 & \cellcolor{gray!12}37.75 & 53.51 & 20.88 & 38.86 & \cellcolor{gray!12}40.18\std{0.95} \\

      {Rearank {\small(7B)}}& \cellcolor{gray!12}19.28 & 18.47 & 19.93 & 19.44 & \cellcolor{gray!12}63.34 & 26.24 & 70.23 & 93.55 & \cellcolor{gray!12}38.00 & 53.83 & 20.99 & 39.17 & \cellcolor{gray!12}40.21\std{0.95} \\

      {ReasonRank {\small(7B)}}& \cellcolor{gray!12}19.23 & 18.59 & 19.88 & 19.23 & \cellcolor{gray!12}\colorbox{backyellow_soft!40}{64.99} & \colorbox{backyellow_soft!40}{27.90} & \colorbox{backyellow_soft!40}{72.52} & \colorbox{backred!50}{94.54} & \cellcolor{gray!12}\colorbox{backyellow_soft!40}{38.82} & \colorbox{backyellow_soft!40}{54.62} & 21.39 & \colorbox{backred!50}{40.45} & \cellcolor{gray!12}41.01\std{0.96} \\

      \midrule
      \rowcolor{gray!10}
      \multicolumn{14}{c}{\fontsize{10}{12}\selectfont \textit{\textbf{Setwise Reranking}}} \\

      {SetR {\small(8B)}}& \cellcolor{gray!12}\colorbox{backred!50}{33.44} & \colorbox{backred!50}{29.98} & \colorbox{backred!50}{35.57} & \colorbox{backred!50}{34.77} & \cellcolor{gray!12}64.04 & 27.82 & 71.41 & 92.88 & \cellcolor{gray!12}\colorbox{backred!50}{40.08} & \colorbox{backred!50}{55.35} & \colorbox{backred!50}{24.62} & \colorbox{backyellow_soft!40}{40.28} & \cellcolor{gray!12}\colorbox{backred!50}{45.85}\std{1.12} \\

      {Rank4Gen {\small(8B)}}& \cellcolor{gray!12}\colorbox{backyellow_soft!40}{25.85} & \colorbox{backyellow_soft!40}{25.07} & \colorbox{backyellow_soft!40}{26.08} & \colorbox{backyellow_soft!40}{26.39} & \cellcolor{gray!12}\colorbox{backred!50}{67.65} & \colorbox{backred!50}{28.04} & \colorbox{backred!50}{80.41} & \colorbox{backyellow_soft!40}{94.49} & \cellcolor{gray!12}38.12 & 54.01 & \colorbox{backyellow_soft!40}{21.99} & 38.37 & \cellcolor{gray!12}\colorbox{backyellow_soft!40}{43.87}\std{1.05} \\

      \bottomrule
    \end{tabular}%

  }
\end{table*}

\nb{Behavior Analysis in Long-form Scenario.} \ding{185} \textit{Reasoning-enhanced methods surpass Setwise reranking to become the leading paradigm.} All three reasoning-enhanced rerankers occupy the top three Overall positions ($38.19$, $37.18$, $37.15$), while the Setwise methods that dominated in Short-form fall to $36.52$ and $36.18$, the latter only $0.10$ above Only Search. This reversal suggests that multi-turn cumulative scenarios depend more on reasoning capabilities to assess the marginal value of newly retrieved documents. \ding{186} \textit{Performance gaps across methods shrink substantially.} The Overall range spans $2.91$ points ($35.28$ to $38.19$), against roughly $9$ in Short-form, and two rerankers even fall below the $36.08$ obtained without any reranking. This compression suggests that ranker and agent currently operate in isolation, with no information flow between accumulated context and document selection, limiting the performance ceiling and \textbf{highlighting the need for co-optimizing the ranker and search agent throughout the multi-turn process}. \ding{187} \textit{Score degradation across successive search turns.} Rubric scores of all rerankers decrease monotonically across search turns (Figure~\ref{fig:long_form_round}), with particularly pronounced drops at doc and global levels. This reflects not reranker degradation but a shift from broad information acquisition in early turns to targeted gap-filling in later turns, yielding narrower candidate pools that systematically lower scores. We therefore adopt union scoring after cross-turn deduplication (Table~\ref{tab:eval_long_union}), measuring the quality of the cumulative document set delivered to the downstream generator.

\begin{table*}[t!]
  \centering
  \caption{\textbf{Overall Performance Comparison ($\%$) on \dataset{} (Long-form Scenario).}
The \colorbox{backred!50}{\textbf{best}} (1\textsuperscript{st}) and \colorbox{backyellow_soft!40}{\underline{second-best}} (2\textsuperscript{nd}) results are highlighted.
The per-level \textbf{Avg} is the arithmetic mean of the three dimensions within that level, and \textbf{Overall} is the mean of all nine dimensions.
All metrics are reported such that higher is better (\daugshifted).
Only Search (Google Search), Adhoc Reranking, and Reasoning-Enhanced Reranking use a \textbf{fixed top-5 documents}; Setwise Reranking \textbf{adaptively selects the number of documents}.
Subscripts denote the standard deviation of the \textbf{Overall} score across two independent judge passes.}
  \label{tab:eval_long_union}

  \vspace{-5pt}
  \renewcommand{\arraystretch}{1.1}
  \newcommand{\std}[1]{$_{(\pm#1)}$}
  \resizebox{\textwidth}{!}{%
    \begin{tabular}{l|cccc|cccc|cccc|c}
      \toprule
      \multirow{2.5}{*}{\textbf{Ranker}}
        & \multicolumn{4}{c|}{\textbf{Doc-Level}}
        & \multicolumn{4}{c|}{\textbf{Set-Level}}
        & \multicolumn{4}{c|}{\textbf{Global-Level}}
        & \cellcolor{gray!12} \\
      \noalign{\global\aboverulesep=0pt \global\belowrulesep=0pt}
      \cmidrule(l{2pt}r{2pt}){2-5} \cmidrule(l{2pt}r{2pt}){6-9} \cmidrule(l{2pt}r{2pt}){10-13}
      \noalign{\global\aboverulesep=.65ex \global\belowrulesep=.65ex}
        &
        \cellcolor{gray!12}\textbf{Avg} & \textbf{Rel.} & \textbf{Aut.} & \textbf{Qua.} 
        & \cellcolor{gray!12}\textbf{Avg} & \textbf{Cmp.} & \textbf{Red.} & \textbf{Con.} 
        & \cellcolor{gray!12}\textbf{Avg} & \textbf{Cov.} & \textbf{Den.} & \textbf{Rea.} 
        & \cellcolor{gray!12}\raisebox{4.5pt}[0pt][0pt]{\textbf{Overall}} \\
      \midrule

      Only Search& \cellcolor{gray!12}20.20 & 29.28 & 11.38 & 19.93 & \cellcolor{gray!12}\colorbox{backyellow_soft!40}{\underline{64.52}} & 31.90 & \colorbox{backyellow_soft!40}{\underline{71.06}} & 90.61 & \cellcolor{gray!12}23.51 & 22.26 & 31.93 & 16.35 & \cellcolor{gray!12}36.08\std{0.31} \\

      \midrule

      \rowcolor{gray!10}
      \multicolumn{14}{c}{\fontsize{10}{12}\selectfont \textit{\textbf{Adhoc Reranking}}} \\

      BGE-Reranker$_{L.}$& \cellcolor{gray!12}21.57 & 31.66 & 12.76 & 20.30 & \cellcolor{gray!12}62.10 & 33.33 & 65.09 & 87.87 & \cellcolor{gray!12}25.27 & 23.48 & 33.96 & 18.37 & \cellcolor{gray!12}36.31\std{0.40} \\

      MonoT5 {\small(3B)}& \cellcolor{gray!12}21.76 & 32.23 & 12.57 & 20.47 & \cellcolor{gray!12}63.85 & \colorbox{backyellow_soft!40}{\underline{34.70}} & 69.61 & 87.25 & \cellcolor{gray!12}25.11 & \colorbox{backyellow_soft!40}{\underline{24.89}} & 32.80 & 17.64 & \cellcolor{gray!12}36.91\std{0.50} \\

      RankT5 {\small(3B)}& \cellcolor{gray!12}20.01 & 29.85 & 11.34 & 18.84 & \cellcolor{gray!12}62.31 & 30.83 & 69.62 & 86.49 & \cellcolor{gray!12}23.52 & 22.06 & 32.54 & 15.97 & \cellcolor{gray!12}35.28\std{0.30} \\

      RankLlama {\small(7B)}& \cellcolor{gray!12}20.96 & 29.95 & 13.32 & 19.62 & \cellcolor{gray!12}62.69 & 32.53 & 64.76 & 90.77 & \cellcolor{gray!12}23.93 & 22.58 & 31.38 & 17.82 & \cellcolor{gray!12}35.86\std{0.34} \\

      {Setwise {\small(7B)}}& \cellcolor{gray!12}21.71 & 31.14 & 13.44 & 20.55 & \cellcolor{gray!12}63.06 & 30.03 & 68.19 & \colorbox{backyellow_soft!40}{\underline{90.96}} & \cellcolor{gray!12}23.71 & 22.43 & 33.19 & 15.52 & \cellcolor{gray!12}36.16\std{0.85} \\

      \midrule
      \rowcolor{gray!10}
      \multicolumn{14}{c}{\fontsize{10}{12}\selectfont \textit{\textbf{Reasoning-Enhanced Reranking}}} \\

      {Rank1 {\small(7B)}}& \cellcolor{gray!12}22.93 & 32.56 & \colorbox{backred!50}{\textbf{13.84}} & \colorbox{backred!50}{\textbf{22.39}} & \cellcolor{gray!12}62.49 & 33.51 & 65.95 & 88.00 & \cellcolor{gray!12}\colorbox{backyellow_soft!40}{\underline{26.04}} & 24.13 & 35.26 & \colorbox{backyellow_soft!40}{\underline{18.72}} & \cellcolor{gray!12}37.15\std{0.61} \\

      {Rearank {\small(7B)}}& \cellcolor{gray!12}22.50 & 32.63 & 13.24 & 21.64 & \cellcolor{gray!12}63.64 & 34.37 & 67.53 & 89.02 & \cellcolor{gray!12}25.41 & 24.41 & 33.64 & 18.18 & \cellcolor{gray!12}\colorbox{backyellow_soft!40}{\underline{37.18}}\std{0.40} \\

      {ReasonRank {\small(7B)}}& \cellcolor{gray!12}\colorbox{backyellow_soft!40}{\underline{22.95}} & \colorbox{backyellow_soft!40}{\underline{33.94}} & 13.39 & 21.52 & \cellcolor{gray!12}\colorbox{backred!50}{\textbf{64.82}} & \colorbox{backred!50}{\textbf{36.94}} & 67.98 & 89.55 & \cellcolor{gray!12}\colorbox{backred!50}{\textbf{26.80}} & \colorbox{backred!50}{\textbf{25.92}} & \colorbox{backyellow_soft!40}{\underline{35.34}} & \colorbox{backred!50}{\textbf{19.14}} & \cellcolor{gray!12}\colorbox{backred!50}{\textbf{38.19}}\std{0.03} \\

      \midrule
      \rowcolor{gray!10}
      \multicolumn{14}{c}{\fontsize{10}{12}\selectfont \textit{\textbf{Setwise Reranking}}} \\

      {SetR {\small(8B)}}& \cellcolor{gray!12}21.13 & 30.46 & 12.45 & 20.48 & \cellcolor{gray!12}63.31 & 32.53 & 65.73 & \colorbox{backred!50}{\textbf{91.68}} & \cellcolor{gray!12}24.10 & 24.38 & 29.86 & 18.06 & \cellcolor{gray!12}36.18\std{0.52} \\

      {Rank4Gen {\small(8B)}}& \cellcolor{gray!12}\colorbox{backred!50}{\textbf{23.57}} & \colorbox{backred!50}{\textbf{35.09}} & \colorbox{backyellow_soft!40}{\underline{13.45}} & \colorbox{backyellow_soft!40}{\underline{22.18}} & \cellcolor{gray!12}62.54 & 28.00 & \colorbox{backred!50}{\textbf{74.64}} & 84.98 & \cellcolor{gray!12}23.45 & 20.34 & \colorbox{backred!50}{\textbf{36.48}} & 13.53 & \cellcolor{gray!12}36.52\std{0.21} \\

      \bottomrule
    \end{tabular}%
  }
\end{table*}

\nb{Behavior Analysis Across Scenarios.} \ding{188} \textit{The optimal strategy varies by scenario, and no method is universally superior.} SetR tops the Short-form scenario at $45.85$ but ranks seventh of eleven in Long-form at $36.18$, while ReasonRank rises from third in Short-form to first in Long-form at $38.19$. No existing reranker is best in both settings. This finding reveals a generalization gap in current methods and motivates the development of more versatile rerankers that can adapt to varying query complexities and document interaction patterns. \ding{189} \textit{Weaknesses concentrate on the dimensions that require composing evidence rather than filtering it.} Redundancy and Conflict are the only dimensions on which every reranker scores highly, and both are satisfied by avoiding bad combinations rather than by assembling good ones. Complementarity, Density, and Reachability, which all demand positive construction, remain low in both scenarios. Existing rerankers can therefore filter but not compose, further validating the necessity of the multi-dimensional rubric evaluation proposed here.

\subsection{Results on Rubric-Oriented Set Selection}

The above analysis shows that existing rerankers score poorly on the dimensions that require composing evidence. This naturally raises a question: \textit{if rubrics can quantify set quality, can they also guide document selection?} We probe this question with \method, which supplies the rubrics directly as selection criteria and requires no additional training.

\nb{Metric \& Analysis.} Table~\ref{tab:em_f1} reports the downstream generation performance of document sets selected by different methods. The Short-form scenario is evaluated using EM and F1, while the Long-form scenario is evaluated using LLM-judge scoring. \ding{182} In the short-form scenario, \method attains the best result on every metric, ahead of the strongest baseline SetR by $0.97\uparrow$ on EM and $0.62\uparrow$ on F1, while using only $2.66$ documents on average against the fixed top-5 of conventional rerankers. Rubric guidance therefore identifies compact yet sufficient subsets by discarding redundant or distracting documents. \ding{183} In the long-form scenario the gain lies in efficiency rather than in score. \method reaches $70.57$, on par with SetR at $70.54$, but attains that level with $20.52$ documents instead of $29.23$ and $4.52$ search turns instead of $4.73$. Rubric guidance thus meets the same quality requirements at roughly $30\%$ lower retrieval cost. Since the rubrics derive in part from the reference answer, these results indicate what the evaluation criteria make possible rather than describe a deployable reranker.

\nb{Outlook.} Prompting uses a rubric only at inference, where the criteria are read once and never internalized. In future work, the same rubrics may enter training at three points. They may select silver labels for supervised fine-tuning by retaining the subsets that satisfy them; they may define the scoring standard of a reward model, turning set quality into a learnable objective for reinforcement learning; and they may act as privileged information for a teacher in on-policy distillation, supplied during training but withheld at inference, so that the deployed reranker needs no rubric at all. Viewed this way, \dataset{} can serve not only as a benchmark for evaluation but also, potentially, as a source of supervision throughout training, thereby offering a direction for subsequent work.

\begin{table*}[t!]
  \centering
  \caption{\textbf{Downstream Answer Performance of Different Rerankers.} \textit{\# of Psg} in Long-form denotes the number of unique documents used across all search rounds.}
  \label{tab:em_f1}

  \vspace{-5pt}
  \renewcommand{\arraystretch}{1.1}
  \newcommand{\gain}[1]{$_{(+#1)}$}
  \newcommand{\gainb}[1]{$_{(\mathbf{+#1})}$}
  \resizebox{\textwidth}{!}{%
    \begin{tabular}{l|l|llllllll}
      \toprule
      \multirow{2}{*}{\textbf{Ranker}} & \textbf{Rubric4Setwise} & \textbf{SetR} & \textbf{Rank4Gen} & \textbf{ReasonRank} & \textbf{Rearank} & \textbf{RankLlama} & \textbf{MonoT5} & \textbf{BGE-Reranker$_{L.}$} & \textbf{Vanilla} \\
       & 8B (Ours) & 8B & 8B & 7B & 7B & 7B & 3B &550M &No Reranking \\
      \midrule

      \rowcolor{gray!10}
      \multicolumn{10}{c}{\fontsize{10}{12}\selectfont \textit{\textbf{Short-form Scenario}}} \\

      \# of Psg & 2.66 & 2.75 & 3.05 & 5 & 5 & 5 & 5 & 5 & 5 \\
\midrule
      EM \daugshifted  & \textbf{26.10}\gainb{7.47} & \underline{25.13}\gain{6.50} & 25.08\gain{6.45} & 20.14\gain{1.51} & 21.11\gain{2.48} & 20.28\gain{1.65} & 20.23\gain{1.60} & 19.94\gain{1.31} & 18.63 \\

      F1-Score \daugshifted  & \textbf{29.32}\gainb{8.22} & \underline{28.70}\gain{7.60} & 28.65\gain{7.55} & 22.75\gain{1.65} & 24.25\gain{3.15} & 23.00\gain{1.90} & 23.20\gain{2.10} & 22.27\gain{1.17} & 21.10 \\

      Avg \daugshifted  & \textbf{27.71}\gainb{7.84} & \underline{26.92}\gain{7.05} & 26.87\gain{7.00} & 21.45\gain{1.58} & 22.68\gain{2.81} & 21.64\gain{1.77} & 21.72\gain{1.85} & 21.11\gain{1.24} & 19.87 \\

      \midrule
      \rowcolor{gray!10}
      \multicolumn{10}{c}{\fontsize{10}{12}\selectfont \textit{\textbf{Long-form Scenario}}} \\

      \# of Psg & 20.52 & 29.23 & 12.13 & 20.30 & 21.39 & 22.24 & 20.42 & 20.84 & 20.94 \\
      \# Avg Rounds & 4.52 & 4.73 & 5.01 & 4.30 & 4.63 & 4.78 & 4.31 & 4.50 & 4.49 \\
\midrule
      LLM-judge \daugshifted &\textbf{70.57\gainb{3.49}} & \underline{70.54}\gain{3.46}& 68.32\gain{1.24}&68.36\gain{1.28} & 68.43\gain{1.35}& 69.10\gain{2.02}& 70.33\gain{3.25}& 69.48\gain{2.40}& 67.08 \\

      \bottomrule
    \end{tabular}%
  }
\end{table*}

\section{Case Study and Qualitative Analysis}

For the query in Figure~\ref{fig:case}, only Doc [11] directly describes the crowning event; the remaining documents are retrieved through surface keyword overlap with ``Miss USA'' and ``crowned.'' Redundancy (2/4) and Conflict (2/4) reveal repeated yet contradictory claims about the crowning identity; Reachability (1/4) further confirms the documents cannot jointly reconstruct the full reasoning chain. This case demonstrates that traditional nDCG cannot capture set-level coordination deficiencies, whereas our rubric design pinpoints weaknesses and guides document set optimization.

\begin{figure*}[h!]
  \centering
  \includegraphics[width=1\linewidth]{figures/case.png}
  \caption{\textbf{Case Study.} Each rubric dimension pinpoints specific document set deficiencies, guiding targeted optimization.}
  \label{fig:case}
\end{figure*}

\section{Conclusion and Limitation}

We present an evaluate-diagnose-optimize framework for document set quality. We design \dataset, the first set-level evaluation benchmark spanning three granularity levels and nine dimensions, revealing that existing rerankers are universally weak on cross-document coordination, with no single method maintaining top performance across both settings. Building on this, \method converts rubric signals into selection guidance, achieving the best downstream generation with fewer documents and search rounds. We note that \method operates under an oracle setting (rubrics generated with reference answers), establishing an empirical upper bound that validates rubric-as-signal effectiveness. A natural next step is distilling rubric preferences into trainable rewards, enabling rerankers to internalize multi-dimensional quality awareness without reference answers. We hope this work provides an evaluation foundation and optimization pathway for advancing retrieval systems from retrieving relevant documents to assembling optimal document sets.

\bibliography{references}
\bibliographystyle{main}

\clearpage
\appendix

\section{Implementation Details of \dataset}

\subsection{Data Source}

\textbf{Short-form Datasets.}
We adopt four established multi-hop question answering benchmarks.
\ding{182} \textbf{HotpotQA}~\citep{HotpotQA} is a large-scale multi-hop QA dataset containing over 112K crowd-sourced question-answer pairs based on Wikipedia, requiring models to reason across multiple supporting documents to derive answers.
\ding{183}  \textbf{2WikiMultihopQA}~\citep{2WikiMultihopQA} is constructed from Wikipedia and Wikidata with approximately 192K samples, providing explicit evidence paths in the form of triples and covering four question types: comparison, inference, compositional, and bridge-comparison.
\ding{184}  \textbf{MuSiQue}~\citep{MuSiQue} employs a bottom-up construction method that composes 2-to-4-hop questions from carefully selected single-hop questions, featuring six composition structures that make it more challenging and less susceptible to reasoning shortcuts than prior benchmarks.
\ding{185}  \textbf{Bamboogle}~\citep{Bamboogle} is a curated test set of 125 two-hop questions specifically designed so that no single search query can directly retrieve the answer, requiring models to identify and utilize intermediate bridging entities.

\textbf{Long-form Dataset.} \textbf{ResearchQA}~\citep{ResearchQA} is a large-scale scholarly QA evaluation resource distilled from survey articles across 75 research fields, comprising 21K queries paired with 160K rubric items. Each rubric lists query-specific evaluation criteria including citing papers, making explanations, and describing limitations. Validation by 31 Ph.D. annotators confirms that 96\% of queries reflect doctoral-level information needs.

\subsection{Generator and Search Agent}

\textbf{Short-form Generator.}
For the Short-form scenario, we use \textit{Llama-3.1-8B-Instruct} as the downstream generator. Given the top-$k$ documents selected by each reranker, the generator produces a concise answer conditioned on the retrieved passages. We adopt this moderate-scale instruction-tuned model to ensure that downstream performance differences primarily reflect document set quality rather than generator capacity.

\textbf{Long-form Search Agent.}
For the Long-form scenario, we use \textit{DR.Tulu-8B}~\citep{DR-Tulu}, the first open-source deep research agent built on Qwen3-8B and trained end-to-end via Reinforcement Learning with Evolving Rubrics. DR.Tulu operates an autonomous multi-turn search loop, alternating among internal planning (\texttt{think}), tool invocation (\texttt{call\_tool}), and final answer generation (\texttt{answer}). Its tool suite includes web search, web browsing, and paper retrieval. On scientific, medical, and general long-form benchmarks, DR.Tulu-8B matches or surpasses commercial deep research systems at approximately three orders of magnitude lower cost per query. In our setup, we integrate different rerankers into the agent's retrieval pipeline to evaluate how document set selection affects the quality of the final synthesized report.

\subsection{Reranker}

The baselines we used for comparison are as fol lows, each selected for its unique strengths and capabilities in specific retrieval or reranking tasks:

\textbf{BGE-Reranker-Large}~\citep{BGE-Reranker} is a cross-encoder reranker developed as part of the C-Pack embedding suite by BAAI. It computes relevance scores by concatenating the query and document as a single input sequence, allowing full bidirectional attention between them. We use the \texttt{bge-reranker-large} variant with approximately 560M parameters based on the XLM-RoBERTa architecture.

\textbf{MonoT5}~\citep{MonoT5} is a pointwise reranker that casts document relevance estimation as a sequence-to-sequence task using the T5 architecture. Given a query-document pair, it generates the target token ``true'' or ``false'' to indicate relevance, and the softmax probability of ``true'' serves as the relevance score. We use the \texttt{monot5-3b-msmarco-10k} checkpoint.

\textbf{RankT5}~\citep{RankT5} extends the T5 encoder-decoder framework for ranking by directly producing a scalar relevance score from the decoder's output logits, rather than relying on token generation probabilities. It supports both pointwise and pairwise/listwise ranking losses during training. We use the \texttt{rankt5-base} checkpoint fine-tuned on MS MARCO.

\textbf{RankLLaMA}~\citep{RankLlama} is a pointwise reranker built by fine-tuning LLaMA-2-7B on the MS MARCO passage ranking dataset. It extracts the hidden state of the last token as the sequence representation and applies a linear projection to produce a scalar relevance score. Training uses pairwise ranking loss with hard negatives from BM25 and dense retrievers.

\textbf{RankVicuna}~\citep{RankVicuna} is one of the first fully open-source LLMs for zero-shot listwise passage reranking. Built on the Vicuna-7B model, it is trained via permutation distillation from GPT-3.5-based RankGPT, learning to produce a reranked permutation of candidate passages given a query and a passage window.

\textbf{RankZephyr}~\citep{RankZephyr} is a 7B-parameter listwise reranker built on the Zephyr language model. It is trained through a combination of permutation distillation from GPT-3.5/4 outputs and direct preference optimization (DPO), achieving effectiveness that bridges or surpasses proprietary GPT-4 on standard benchmarks while remaining fully open-source.

\textbf{Setwise}~\citep{Setwise} is a prompting-based reranking approach that presents a set of candidate documents simultaneously and asks the LLM to select the most relevant one. By combining the comparative advantages of pairwise methods with the efficiency of pointwise approaches, it achieves a favorable trade-off between ranking effectiveness and computational cost through techniques such as Heapsort-based selection.

\textbf{Rank1}~\citep{Rank-R1} is the first pointwise reranker that leverages extended test-time computation via chain-of-thought reasoning. It is trained by distilling reasoning traces from DeepSeek-R1 on over 600K query-document pairs from MS MARCO, enabling the model to produce step-by-step reasoning before outputting a binary relevance judgment.

\begin{table*}[t!]
  \centering
  \caption{\textbf{Rubric Count Statistics on \dataset.} We report the total number of rubrics generated per dimension (\textit{top}) and the average number of rubrics per query (\textit{bottom}), grouped by \textbf{Doc-Level}, \textbf{Set-Level}, and \textbf{Global-Level} dimensions.}
  \label{tab:rubric_count}
  \vspace{-5pt}
  \renewcommand{\arraystretch}{1.1}
  \resizebox{\textwidth}{!}{%
    \begin{tabular}{l|cccc|cccc|cccc|c}
      \toprule
      & \multicolumn{4}{c|}{\textbf{Doc-Level}}
      & \multicolumn{4}{c|}{\textbf{Set-Level}}
      & \multicolumn{4}{c|}{\textbf{Global-Level}}
      & \multirow{2}{*}{\textbf{Overall}} \\
      \noalign{\global\aboverulesep=0pt \global\belowrulesep=0pt}
      \cmidrule(l{2pt}r{2pt}){2-5} \cmidrule(l{2pt}r{2pt}){6-9} \cmidrule(l{2pt}r{2pt}){10-13}
      \noalign{\global\aboverulesep=.65ex \global\belowrulesep=.65ex}
      & \textbf{Rel.} & \textbf{Aut.} & \textbf{Qua.} & \textbf{Overall}
      & \textbf{Cmp.} & \textbf{Red.} & \textbf{Con.} & \textbf{Overall}
      & \textbf{Cov.} & \textbf{Den.} & \textbf{Rea.} & \textbf{Overall}
      & \\
      \midrule

      \rowcolor{gray!10}
      \multicolumn{14}{c}{\textbf{\textit{Short-form Scenario}}} \\

      \multicolumn{14}{l}{\textbf{\textit{Total Rubric Count}}} \\
      \midrule

      GPT 5.1        & 4,065 & 3,768 & 3,286 & 11,119 & 4,176 & 3,057 & 3,269 & 10,502 & 4,196 & 3,765 & 3,940 & 11,901 & 33,522 \\
      Gemini 3.1-Pro-Preview & 2,243 & 2,701 & 2,231 & 7,175  & 2,241 & 2,235 & 2,501 & 6,977  & 2,231 & 2,231 & 2,231 & 6,693  & 20,845 \\
      Deepseek-V4 Pro & 2,581 & 3,321 & 2,799 & 8,701 & 2,633 & 2,361 & 3,217 & 8,211 & 2,594 & 2,231 & 2,315 & 7,140  & 24,052 \\

      \midrule
      \multicolumn{14}{l}{\textbf{\textit{\# Rubrics / Query}}} \\
      \midrule

      GPT 5.1        & 1.82 & 1.69 & 1.69 & 4.98 & 1.87 & 1.37 & 1.47 & 4.71 & 1.88 & 1.69 & 1.77 & 5.33 & 15.03 \\
      Gemini 3.1-Pro-Preview & 1.01 & 1.21 & 1.00 & 3.22 & 1.00 & 1.00 & 1.12 & 3.13 & 1.00 & 1.00 & 1.00 & 3.00 & 9.34  \\
      Deepseek-V4 Pro & 1.16 & 1.49 & 1.25 & 3.90 & 1.18 & 1.06 & 1.44 & 3.68 & 1.16 & 1.00 & 1.04 & 3.20 & 10.78 \\

\midrule  
      \rowcolor{gray!10}
      \multicolumn{14}{c}{\textbf{\textit{Long-form Scenario}}} \\
      
      \multicolumn{14}{l}{\textbf{\textit{Total Rubric Count}}} \\
      
      \midrule

      GPT 5.1        &   486 &   539 &   523 & 1,548 &   532 &   428 &   535 & 1,495 &   413 &   342 &   421 & 1,176 & 4,219 \\
      Gemini 3.1-Pro-Preview &   212 &   292 &   201 &   705 &   251 &   209 &   276 &   736 &   200 &   200 &   200 &   600 & 2,041 \\
      Deepseek-V4 Pro &   398 &   637 &   497 & 1,532 &   539 &   368 &   605 & 1,512 &   426 &   247 &   319 &   992 & 4,036 \\

      \midrule
      \multicolumn{14}{l}{\textbf{\textit{\# Rubrics / Query}}} \\
      \midrule

      GPT 5.1        & 2.43 & 2.69 & 2.62 & 7.74 & 2.66 & 2.14 & 2.67 & 7.47 & 2.06 & 1.71 & 2.10 & 5.88 & 21.09 \\
      Gemini 3.1-Pro-Preview & 1.06 & 1.46 & 1.00 & 3.52 & 1.25 & 1.04 & 1.38 & 3.68 & 1.00 & 1.00 & 1.00 & 3.00 & 10.21 \\
      Deepseek-V4 Pro & 1.99 & 3.19 & 2.48 & 7.66 & 2.69 & 1.84 & 3.02 & 7.56 & 2.13 & 1.24 & 1.59 & 4.96 & 20.18 \\

      \bottomrule
    \end{tabular}%
  }
\end{table*}

\subsection{Rubric Statistics}

Table~\ref{tab:rubric_count} reports the rubric count statistics of \dataset. We employ three frontier LLMs (GPT 5.1, Gemini 3.1-Pro-Preview, and DeepSeek-V4 Pro) to independently generate rubrics for each query. In the Short-form scenario, the three models produce 33,522, 20,845, and 24,052 total rubrics respectively, averaging approximately 15, 9, and 11 rubrics per query across all nine dimensions. In the Long-form scenario, multi-turn search yields richer information needs, resulting in 4,219, 2,041, and 4,036 total rubrics with averages of roughly 21, 10, and 20 per query. Across both scenarios, rubric counts are relatively balanced among the three granularity levels (Doc, Set, Global), indicating that our dimension design elicits comprehensive coverage rather than concentrating on any single level. After multi-model aggregation, \dataset comprises approximately 28K high-quality rubrics in total.

\subsection{Human Study Interface}

\begin{figure*}[t!]
  \centering
\includegraphics[width=1\linewidth]{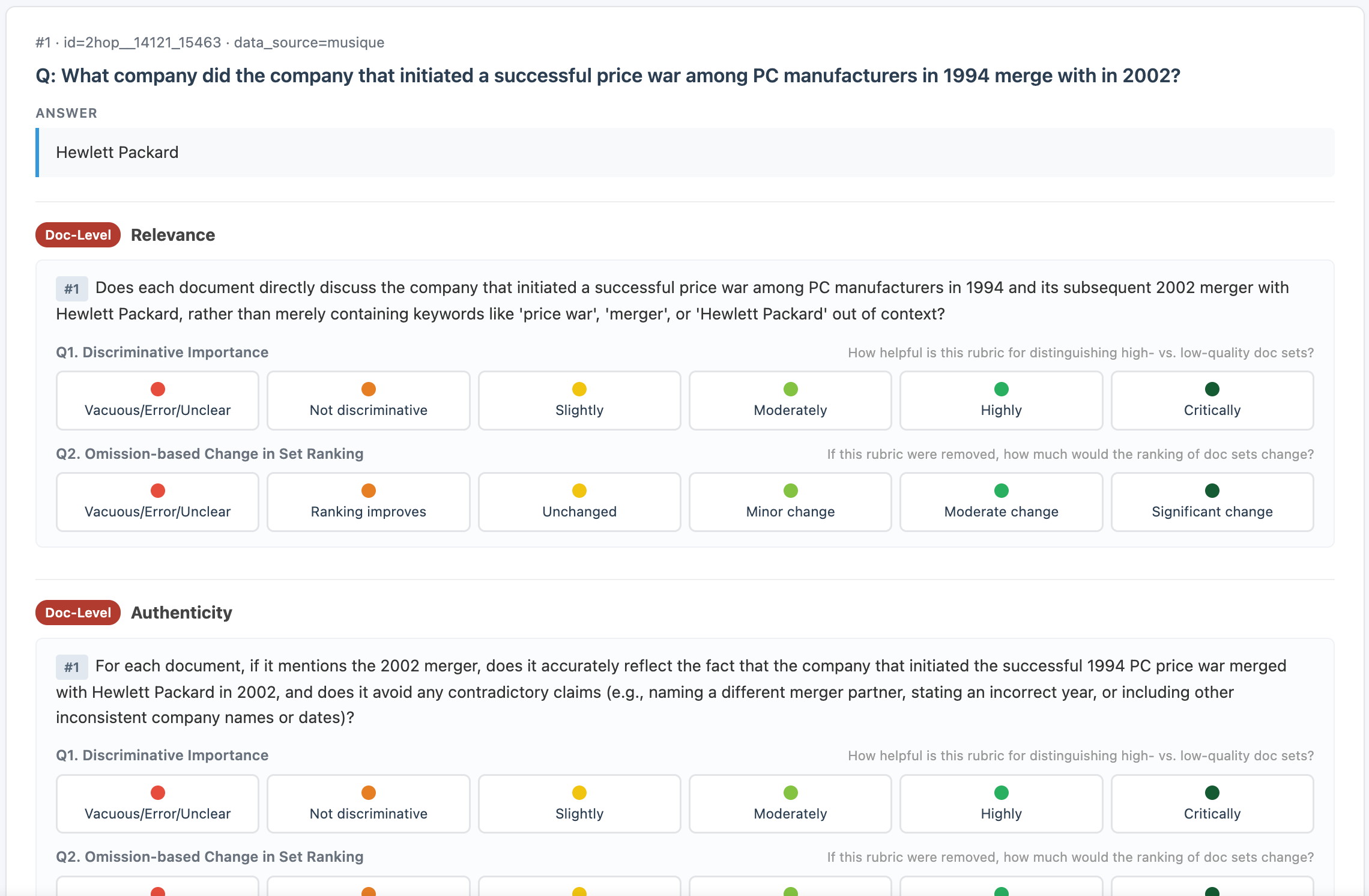}
  \caption{\textbf{Annotation Interface for Human Study of Rubric Quality in the Short-form Scenario. } For each rubric item, annotators answer two questions: (Q1) \textit{Discriminative Importance}, rating how helpful the rubric is for distinguishing high- vs.\ low-quality document sets on a 6-point scale from Vacuous/Error to Critically discriminative; (Q2) \textit{Omission-based Change in Set Ranking}, rating how much the ranking of document sets would change if this rubric were removed, from Unchanged to Significant change.}

  \label{fig:human_study_short}
\end{figure*}

\begin{figure*}[t!]
  \centering
\includegraphics[width=1\linewidth]{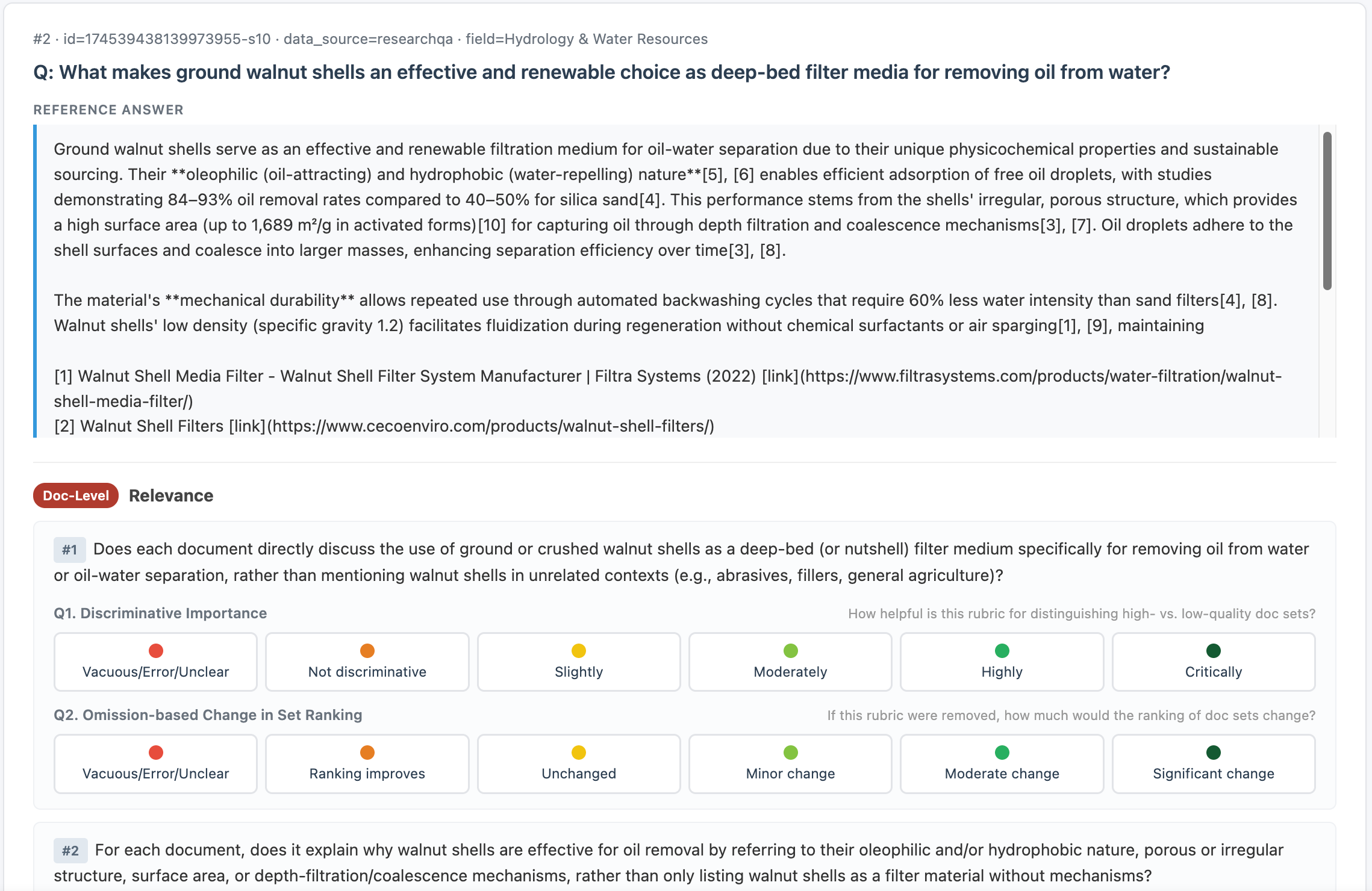}
  \caption{\textbf{Annotation Interface for Human Study of Rubric Quality in the Long-form Scenario. } For each rubric item, annotators answer two questions: (Q1) \textit{Discriminative Importance}, rating how helpful the rubric is for distinguishing high- vs.\ low-quality document sets on a 6-point scale from Vacuous/Error to Critically discriminative; (Q2) \textit{Omission-based Change in Set Ranking}, rating how much the ranking of document sets would change if this rubric were removed, from Unchanged to Significant change.}
  \label{fig:human_study_long}
\end{figure*}

To validate rubric quality, we design a web-based annotation interface and recruit annotators to evaluate each rubric item along two dimensions. Figures~\ref{fig:human_study_short} and~\ref{fig:human_study_long} show the annotation interfaces for the Short-form and Long-form scenarios, respectively.

For each sample, annotators are presented with the query, the gold answer (Short-form) or reference answer (Long-form), and the generated rubric items grouped by their corresponding dimensions (e.g., Doc-Level Relevance, Doc-Level Authenticity). For each rubric item, annotators answer two questions on 6-point Likert scales:

\begin{itemize}
\item \textbf{Q1: Discriminative Importance.} How helpful is this rubric for distinguishing high- vs.\ low-quality document sets? Options range from \textit{Vacuous/Error/Unclear} (the rubric is nonsensical or irrelevant) to \textit{Critically} discriminative (the rubric alone can decisively separate good sets from bad ones).
\item \textbf{Q2: Omission-based Change in Set Ranking.} If this rubric were removed from the evaluation, how much would the ranking of document sets change? Options range from \textit{Vacuous/Error/Unclear} to \textit{Significant change}, with an additional \textit{Ranking improves} option indicating the rubric introduces noise.
\end{itemize}

The Short-form interface displays the query with its short answer, while the Long-form interface additionally provides the full reference answer with citations, reflecting the richer information structure of multi-turn research scenarios. All rubric items are presented in their original dimension groupings to provide annotators with sufficient context for judgment, facilitating efficient and consistent scoring.

\begin{table}[t]
  \centering
  \caption{\textbf{Inter-annotator Agreement.} Krippendorff's $\alpha$ (ordinal) and ICC(3,1) among three Ph.D. experts on 6-point ordinal Likert scales for each evaluation dimension across both scenarios.}
  \label{tab:iaa}
  \renewcommand{\arraystretch}{1.1}
  \setlength{\tabcolsep}{4pt}
  \fontsize{9}{11}\selectfont
    \begin{tabular}{l|l|cc}
      \toprule
      \textbf{Dimension} & \textbf{Scenario} & \textbf{Kripp.\ $\alpha$} & \textbf{ICC(3,1)} \\
      \midrule
      \multirow{2}{*}{Q1}
        & Short-form & 0.79 & 0.82 \\
        & Long-form  & 0.76 & 0.79 \\
      \midrule
      \multirow{2}{*}{Q2}
        & Short-form & 0.85 & 0.88 \\
        & Long-form  & 0.83 & 0.86 \\
      \midrule
      \multicolumn{2}{c|}{\textbf{Overall}} & \textbf{0.81} & \textbf{0.84} \\
      \bottomrule
    \end{tabular}
  
\end{table}

\subsection{Inter-Annotator Agreement Numbers of Human Study}

Since three annotators rate on a 6-point ordinal Likert scale, we adopt the following metrics to measure inter-annotator agreement:

\textbf{Krippendorff's $\alpha$ (ordinal)}, which accounts for the distance between rating levels and is suitable for multi-rater ordinal data:
\begin{equation}
\alpha = 1 - \frac{D_o}{D_e}
\end{equation}
where $D_o$ denotes the observed disagreement among annotators and $D_e$ denotes the expected disagreement under random assignment. $\alpha > 0.667$ indicates acceptable agreement, and $\alpha > 0.800$ indicates high agreement.

\textbf{Intraclass Correlation Coefficient ICC(3,1)}, which treats ratings as continuous values and measures the absolute consistency of annotator scores:
\begin{equation}
\text{ICC} = \frac{MS_R - MS_E}{MS_R + (k-1) \cdot MS_E}
\end{equation}
where $MS_R$ is the between-item mean square, $MS_E$ is the residual mean square, and $k=3$ is the number of annotators. ICC $> 0.75$ indicates excellent reliability.

As shown in Table~\ref{tab:iaa}, the overall results yield Krippendorff's $\alpha = 0.81$ and ICC $= 0.84$, both reaching the almost perfect agreement level, confirming that the annotation protocol supports reliable human judgments.

\subsection{Experiment Resources}

All experiments are conducted on NVIDIA H20 GPUs with 96GB HBM3 memory. Each reranking model is deployed on a single GPU for inference.

\section{More Experimental Results}

\subsection{Reproducibility of Rubric Scoring Across Two Independent Passes}

\begin{figure*}[h!]
  \centering
    \includegraphics[width=1\linewidth]{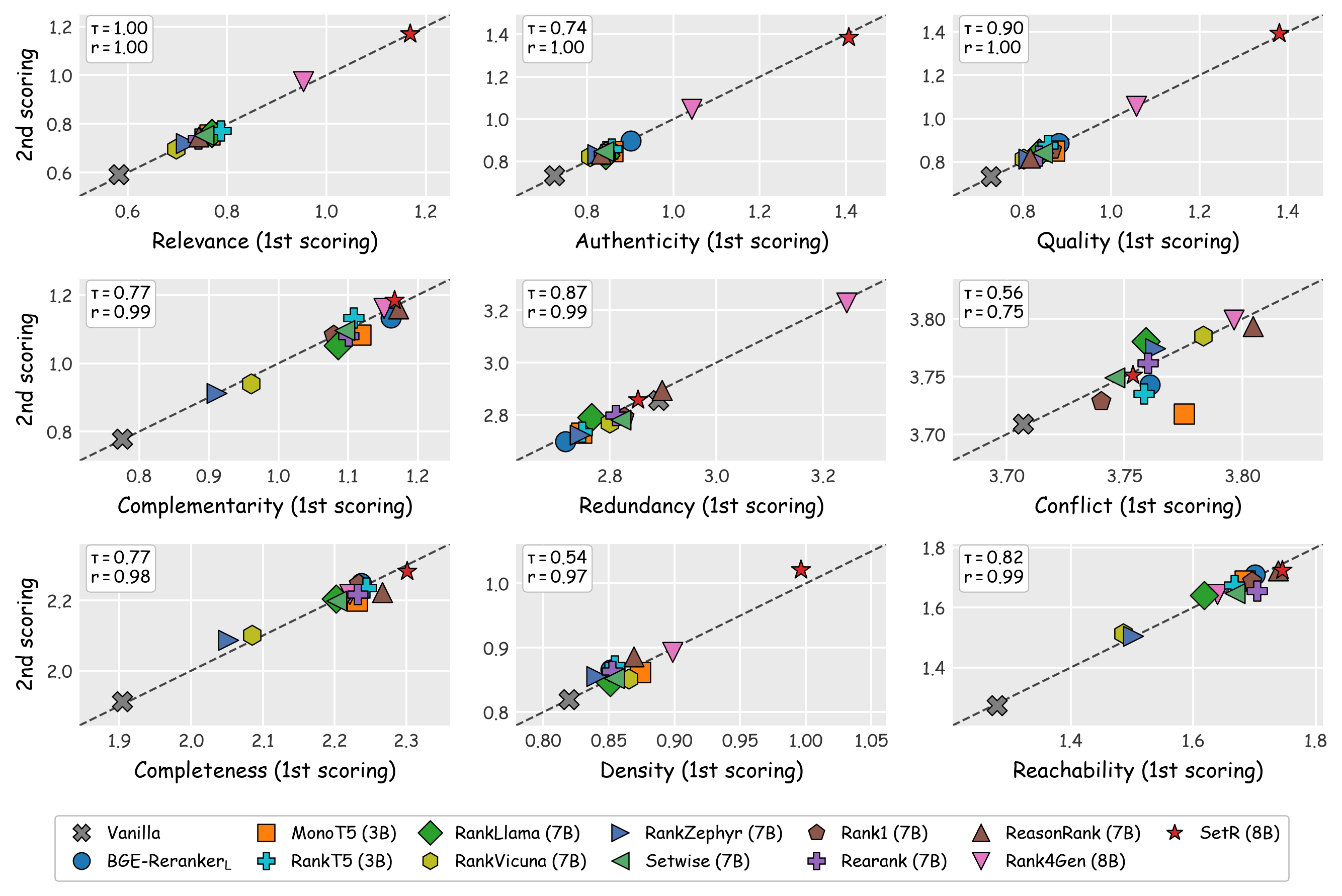}
  \caption{\textbf{Reproducibility of rubric scoring across two independent passes (Short-form Scenario).}}
\label{fig:scoring_stability_short}

\end{figure*}

\begin{figure*}[h!]
  \centering
    \includegraphics[width=1\linewidth]{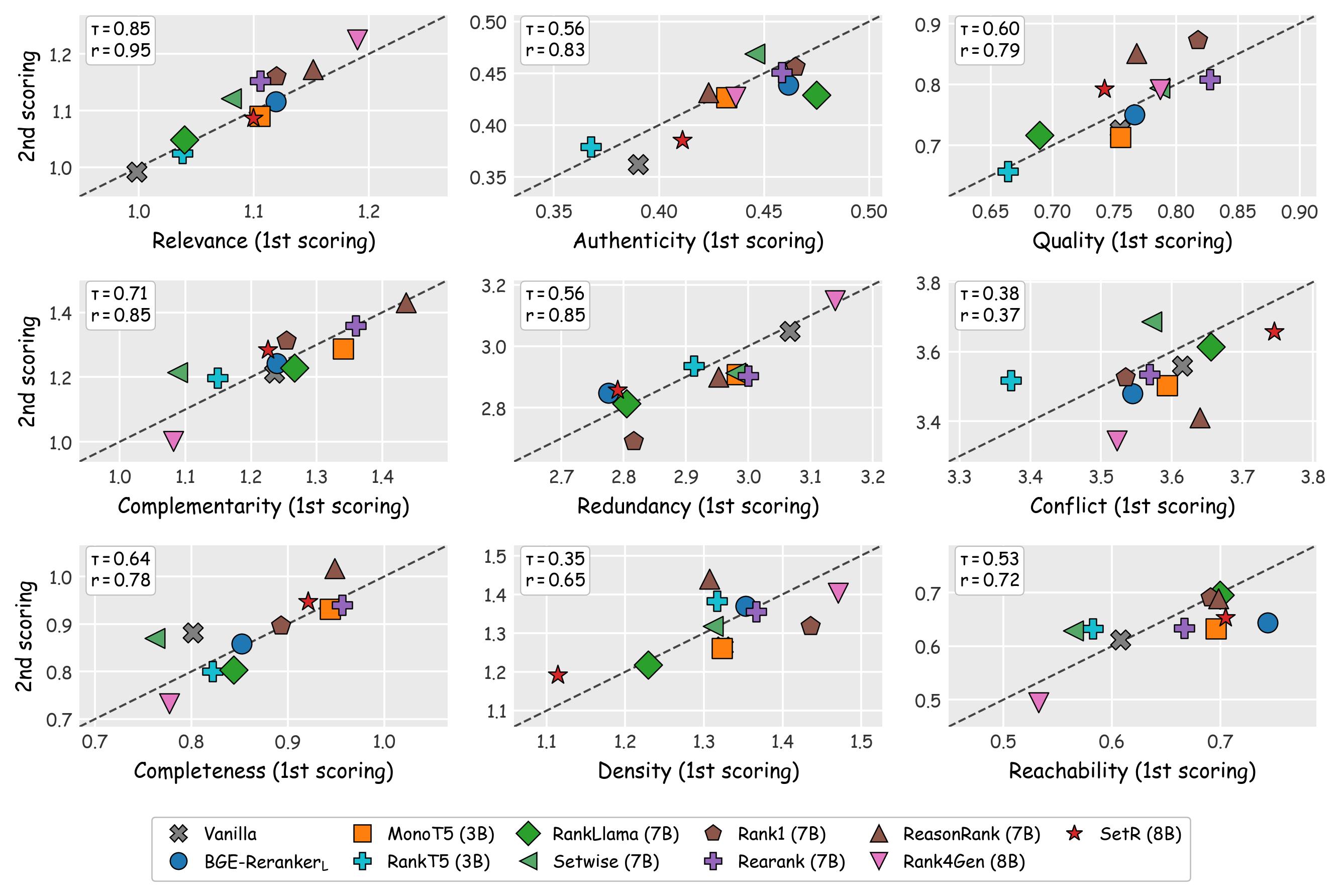}
  \caption{\textbf{Reproducibility of rubric scoring across two independent passes (Long-form Scenario).}}
\label{fig:scoring_stability_long}
\end{figure*}

Figures~\ref{fig:scoring_stability_short} and~\ref{fig:scoring_stability_long} report the between-pass agreement of our LLM-judge rubric scoring on the Short-form and Long-form Scenarios. Each ranker is re-scored end-to-end with an independent second pass using the same prompt and judge; each panel plots one dimension, one marker per ranker, and reports Kendall's $\tau$ on the ranker-mean vector (ordering stability) and Pearson's $r$ on its absolute values (magnitude stability).

Concretely, for a given dimension let $\bar{s}_i^{(1)}$ and $\bar{s}_i^{(2)}$ denote the mean rubric score of the $i$-th reranker (over all its scored (query, sub-rubric, doc) items) under the first and second scoring pass, computed over $N$ rerankers. Pearson's $r$ measures the linear agreement of the two magnitude vectors,
\begin{equation}
r \;=\; \frac{\sum_{i=1}^{N}\bigl(\bar{s}_i^{(1)}-\bar{\mu}^{(1)}\bigr)\bigl(\bar{s}_i^{(2)}-\bar{\mu}^{(2)}\bigr)}
              {\sqrt{\sum_{i=1}^{N}\bigl(\bar{s}_i^{(1)}-\bar{\mu}^{(1)}\bigr)^{2}}\,\sqrt{\sum_{i=1}^{N}\bigl(\bar{s}_i^{(2)}-\bar{\mu}^{(2)}\bigr)^{2}}},
\end{equation}
where $\bar{\mu}^{(k)}=\tfrac{1}{N}\sum_i \bar{s}_i^{(k)}$. Kendall's $\tau$ measures the ordering agreement of the two vectors and is the (concordant $-$ discordant) fraction of reranker pairs,
\begin{equation}
\tau \;=\; \frac{2}{N(N-1)} \sum_{i<j} \operatorname{sgn}\!\bigl(\bar{s}_i^{(1)}-\bar{s}_j^{(1)}\bigr)\,\operatorname{sgn}\!\bigl(\bar{s}_i^{(2)}-\bar{s}_j^{(2)}\bigr).
\end{equation}
Both range in $[-1,1]$; a value near $1$ indicates that the two passes reproduce each other, in magnitude ($r$) or in ordering ($\tau$).

\vspace{5pt}

\textbf{(1) Doc-Level dimensions are the most reproducible in both scenarios.}
In the Short-form Scenario, Relevance, Authenticity and Quality all reach $\tau\!\geq\!0.74$ and $r\!=\!1.00$, corroborated by weighted Cohen's $\kappa \in [0.92,\,0.94]$ at the item level. In the Long-form Scenario, the same three dimensions remain the most stable of the nine ($\tau \in [0.56,\,0.85]$, $r \in [0.79,\,0.95]$), even though the doc pool moves from 5-document short-answer sets to 10--40-document research-QA sets. Relevance in particular retains $\tau\!=\!0.85$ and its top-3 rerankers coincide exactly, confirming that document-quality judgements form the most stable substrate of the rubric under any doc-set size.

\textbf{(2) Global-Level dimensions preserve magnitude but show narrower-range fragility.}
Reachability and Completeness maintain very high magnitude agreement in both scenarios (Short-form $r\!=\!0.99/0.98$; Long-form $r\!=\!0.72/0.78$). Density is the extreme case: its item-level $\%\!\pm\!1$ agreement is $99.7\%$ in the Short-form and $90.9\%$ in the Long-form Scenario, yet $\tau$ drops to $0.54$ and $0.35$ respectively. Inspection of Figures~\ref{fig:scoring_stability_short} and~\ref{fig:scoring_stability_long} shows that all rerankers cluster inside a very narrow Density interval, so residual scoring noise reshuffles neighbouring points even though absolute scores are essentially reproduced. This is a discriminative-range property of the dimension on the current reranker pool, not a defect of the scoring pipeline.

\textbf{(3) Set-Level dimensions are robust in magnitude but rely on the reranker separation.}
Complementarity and Redundancy stay at $r\!\geq\!0.85$ across both scenarios with $\tau$ from $0.56$ to $0.87$. Conflict is the weakest ordering dimension in both settings ($\tau\!=\!0.56$ Short-form, $\tau\!=\!0.38$ Long-form) precisely because $\geq 91\%$ of paired items are scored identically at the item level yet the reranker means fall into a $\leq 0.35$-wide interval, so any noise crosses adjacent rankers. This mirrors the correlation analysis in Appendix~\ref{appendix:correlation_analysis}: current rerankers are essentially undifferentiated on set-level objectives, and the leaderboard ordering on those dimensions is consequently the noisiest.

\textbf{(4) The Long-form Scenario is uniformly less stable than the Short-form Scenario, but the leaderboard is still robust.}
Every dimension drops by 6--20 percentage points on $r$ and $\tau$ when moving from the Short-form to the Long-form Scenario. The gap is expected: Long-form queries are open-ended research questions, each doc set is 2--8 times larger than in the Short-form Scenario, and the union scoring aggregates more sub-rubrics whose edge decisions accumulate noise. Nevertheless, the top-3 rerankers on Relevance agree exactly across the two Long-form passes, and 6 of 9 dimensions retain $r\!\geq\!0.72$ on the ranker-mean vector, so the main-text conclusions on relative reranker ordering are reproducible under independent LLM-judge scoring.

\textbf{Summary.}
Across both scenarios, rubric scoring is highly reproducible where it needs to be: Doc-Level dimensions attain almost-perfect item-level agreement ($\kappa \in [0.77,\,0.94]$) and preserve the reranker ordering. Ordering fragility appears only in dimensions whose reranker separation is intrinsically narrow (Density and Conflict), while their absolute scores remain reproducible ($\%\!\pm\!1 \geq 90\%$). The Long-form Scenario is noisier, as expected from its larger doc sets and open-ended answers, but the leaderboard positions reported in the main text are stable under an independent scoring pass.

\vspace{8pt}

\begin{table*}[h!]
  \centering
    \caption{\textbf{Per-Round Rubric Coverage Scores (\%) on \dataset{} (Long-form Scenario).}
The top two performing results are highlighted in
\colorbox{backred!50}{red} (1\textsuperscript{st}) and
\colorbox{backyellow_soft!40}{yellow} (2\textsuperscript{nd}) backgrounds, respectively.
The per-level \textbf{Avg} is the arithmetic mean of the three dimensions within that level, and \textbf{Overall} is the mean of all nine dimensions.
All metrics are reported such that higher is better (\daugshifted).
We independently evaluate the document set selected by each reranker at each search round. Results are reported for the first round, the last round, and the average across all rounds; the best and second-best results are marked within each block.}
  \label{tab:longform_per_round}

  \vspace{-5pt}
  \renewcommand{\arraystretch}{1.1}
  \newcommand{\std}[1]{$_{(\pm#1)}$}
  \resizebox{\textwidth}{!}{%
    \begin{tabular}{l|cccc|cccc|cccc|c}
      \toprule
      \multirow{2.5}{*}{\textbf{Ranker}}
        & \multicolumn{4}{c|}{\textbf{Doc-Level}}
        & \multicolumn{4}{c|}{\textbf{Set-Level}}
        & \multicolumn{4}{c|}{\textbf{Global-Level}}
        & \cellcolor{gray!12} \\
      \noalign{\global\aboverulesep=0pt \global\belowrulesep=0pt}
      \cmidrule(l{2pt}r{2pt}){2-5} \cmidrule(l{2pt}r{2pt}){6-9} \cmidrule(l{2pt}r{2pt}){10-13}
      \noalign{\global\aboverulesep=.65ex \global\belowrulesep=.65ex}
        &
        \cellcolor{gray!12}\textbf{Avg} & \textbf{Rel.} & \textbf{Aut.} & \textbf{Qua.}
        & \cellcolor{gray!12}\textbf{Avg} & \textbf{Cmp.} & \textbf{Red.} & \textbf{Con.}
        & \cellcolor{gray!12}\textbf{Avg} & \textbf{Cov.} & \textbf{Den.} & \textbf{Rea.}
        & \cellcolor{gray!12}\raisebox{4.5pt}[0pt][0pt]{\textbf{Overall}} \\
      \midrule

      \rowcolor{gray!10}
      \multicolumn{14}{c}{\fontsize{10}{12}\selectfont \textit{\textbf{First Round of Reranking}}} \\

      Only Search& \cellcolor{gray!12}21.37 & 29.96 & 14.09 & 20.07 & \cellcolor{gray!12}64.60 & 16.34 & \colorbox{backyellow_soft!40}{84.23} & 93.22 & \cellcolor{gray!12}19.72 & 14.84 & 37.80 & 6.51 & \cellcolor{gray!12}35.23 \\

      BGE-Reranker$_{L.}$& \cellcolor{gray!12}22.68 & 31.47 & 15.65 & 20.93 & \cellcolor{gray!12}63.10 & 18.12 & 78.27 & 92.92 & \cellcolor{gray!12}20.37 & 15.26 & 38.24 & 7.61 & \cellcolor{gray!12}35.39 \\

      MonoT5 {\small(3B)}& \cellcolor{gray!12}22.94 & 33.65 & 15.09 & 20.09 & \cellcolor{gray!12}62.45 & 17.59 & 81.16 & 88.60 & \cellcolor{gray!12}21.50 & 14.59 & 41.28 & 8.64 & \cellcolor{gray!12}35.63 \\

      RankT5 {\small(3B)}& \cellcolor{gray!12}22.01 & 33.64 & 13.76 & 18.62 & \cellcolor{gray!12}64.21 & 18.16 & 83.03 & 91.45 & \cellcolor{gray!12}19.54 & 13.43 & 37.39 & 7.79 & \cellcolor{gray!12}35.25 \\

      Setwise {\small(7B)}& \cellcolor{gray!12}\colorbox{backred!50}{24.99} & 33.97 & \colorbox{backred!50}{17.81} & \colorbox{backred!50}{23.18} & \cellcolor{gray!12}\colorbox{backyellow_soft!40}{65.22} & 19.91 & 78.36 & \colorbox{backred!50}{97.38} & \cellcolor{gray!12}\colorbox{backyellow_soft!40}{23.95} & 15.63 & \colorbox{backred!50}{46.08} & \colorbox{backyellow_soft!40}{10.13} & \cellcolor{gray!12}\colorbox{backred!50}{38.05} \\

      RankLlama {\small(7B)}& \cellcolor{gray!12}22.35 & 30.99 & 15.71 & 20.35 & \cellcolor{gray!12}62.72 & 18.26 & 80.44 & 89.46 & \cellcolor{gray!12}20.97 & 15.44 & 37.48 & 10.00 & \cellcolor{gray!12}35.35 \\

      Rank1 {\small(7B)}& \cellcolor{gray!12}24.18 & \colorbox{backyellow_soft!40}{34.44} & 16.22 & 21.88 & \cellcolor{gray!12}63.93 & \colorbox{backred!50}{22.52} & 75.50 & 93.77 & \cellcolor{gray!12}\colorbox{backred!50}{24.00} & 15.24 & 45.17 & \colorbox{backred!50}{11.60} & \cellcolor{gray!12}\colorbox{backyellow_soft!40}{37.37} \\

      {Rearank {\small(7B)}}& \cellcolor{gray!12}24.53 & \colorbox{backred!50}{34.49} & 16.40 & 22.71 & \cellcolor{gray!12}61.74 & 15.87 & 77.25 & 92.09 & \cellcolor{gray!12}22.87 & \colorbox{backyellow_soft!40}{15.64} & 44.12 & 8.84 & \cellcolor{gray!12}36.38 \\

      {ReasonRank {\small(7B)}}& \cellcolor{gray!12}22.91 & 32.20 & 14.48 & 22.06 & \cellcolor{gray!12}63.90 & 19.34 & 77.77 & 94.58 & \cellcolor{gray!12}22.78 & \colorbox{backred!50}{15.83} & 43.39 & 9.13 & \cellcolor{gray!12}36.53 \\

      {SetR {\small(8B)}}& \cellcolor{gray!12}21.63 & 31.64 & 13.95 & 19.30 & \cellcolor{gray!12}62.08 & \colorbox{backyellow_soft!40}{20.02} & 74.30 & 91.91 & \cellcolor{gray!12}19.24 & 14.77 & 35.47 & 7.47 & \cellcolor{gray!12}34.31 \\

      {Rank4Gen {\small(8B)}}& \cellcolor{gray!12}\colorbox{backyellow_soft!40}{24.73} & 33.81 & \colorbox{backyellow_soft!40}{17.33} & \colorbox{backyellow_soft!40}{23.05} & \cellcolor{gray!12}\colorbox{backred!50}{65.46} & 11.99 & \colorbox{backred!50}{87.32} & \colorbox{backyellow_soft!40}{97.06} & \cellcolor{gray!12}20.38 & 10.07 & \colorbox{backyellow_soft!40}{45.82} & 5.24 & \cellcolor{gray!12}36.85 \\

      \midrule

      \rowcolor{gray!10}
      \multicolumn{14}{c}{\fontsize{10}{12}\selectfont \textit{\textbf{Last Round of Reranking}}} \\

      Only Search& \cellcolor{gray!12}18.34 & 21.78 & 13.56 & 19.68 & \cellcolor{gray!12}\colorbox{backyellow_soft!40}{65.25} & 17.15 & \colorbox{backyellow_soft!40}{85.94} & 92.65 & \cellcolor{gray!12}18.75 & \colorbox{backred!50}{15.40} & 33.71 & 7.13 & \cellcolor{gray!12}34.11 \\

      BGE-Reranker$_{L.}$& \cellcolor{gray!12}17.22 & 21.33 & 13.57 & 16.76 & \cellcolor{gray!12}63.34 & 13.60 & 78.79 & \colorbox{backyellow_soft!40}{97.62} & \cellcolor{gray!12}19.38 & 14.34 & 36.80 & 7.01 & \cellcolor{gray!12}33.31 \\

      MonoT5 {\small(3B)}& \cellcolor{gray!12}\colorbox{backyellow_soft!40}{21.46} & \colorbox{backyellow_soft!40}{26.88} & \colorbox{backred!50}{16.17} & \colorbox{backyellow_soft!40}{21.32} & \cellcolor{gray!12}64.33 & 17.77 & 79.87 & 95.35 & \cellcolor{gray!12}\colorbox{backred!50}{20.85} & 14.24 & \colorbox{backred!50}{40.15} & 8.15 & \cellcolor{gray!12}\colorbox{backyellow_soft!40}{35.54} \\

      RankT5 {\small(3B)}& \cellcolor{gray!12}18.84 & 24.63 & 12.16 & 19.72 & \cellcolor{gray!12}64.82 & 18.11 & 82.27 & 94.08 & \cellcolor{gray!12}19.33 & 13.62 & 36.14 & 8.24 & \cellcolor{gray!12}34.33 \\

      Setwise {\small(7B)}& \cellcolor{gray!12}19.59 & 23.03 & \colorbox{backyellow_soft!40}{16.16} & 19.59 & \cellcolor{gray!12}64.15 & 17.21 & 80.11 & 95.12 & \cellcolor{gray!12}\colorbox{backred!50}{20.85} & 14.41 & \colorbox{backyellow_soft!40}{39.21} & \colorbox{backred!50}{8.94} & \cellcolor{gray!12}34.86 \\

      RankLlama {\small(7B)}& \cellcolor{gray!12}16.61 & 20.66 & 13.05 & 16.12 & \cellcolor{gray!12}59.62 & 12.05 & 78.21 & 88.60 & \cellcolor{gray!12}15.69 & 10.52 & 31.14 & 5.40 & \cellcolor{gray!12}30.64 \\

      Rank1 {\small(7B)}& \cellcolor{gray!12}20.48 & 26.54 & 13.83 & 21.08 & \cellcolor{gray!12}63.58 & \colorbox{backred!50}{18.94} & 79.55 & 92.26 & \cellcolor{gray!12}19.80 & \colorbox{backyellow_soft!40}{14.74} & 35.83 & \colorbox{backyellow_soft!40}{8.84} & \cellcolor{gray!12}34.62 \\

      {Rearank {\small(7B)}}& \cellcolor{gray!12}18.81 & 23.64 & 14.39 & 18.41 & \cellcolor{gray!12}65.16 & 16.77 & 81.92 & 96.80 & \cellcolor{gray!12}18.74 & 12.72 & 36.23 & 7.28 & \cellcolor{gray!12}34.24 \\

      {ReasonRank {\small(7B)}}& \cellcolor{gray!12}\colorbox{backred!50}{21.93} & \colorbox{backred!50}{28.94} & 14.34 & \colorbox{backred!50}{22.52} & \cellcolor{gray!12}\colorbox{backred!50}{65.32} & \colorbox{backyellow_soft!40}{18.25} & 80.08 & \colorbox{backyellow_soft!40}{97.62} & \cellcolor{gray!12}\colorbox{backyellow_soft!40}{20.36} & 14.62 & 38.12 & 8.33 & \cellcolor{gray!12}\colorbox{backred!50}{35.87} \\

      {SetR {\small(8B)}}& \cellcolor{gray!12}19.61 & 24.62 & 14.89 & 19.32 & \cellcolor{gray!12}64.84 & 18.08 & 77.11 & \colorbox{backred!50}{99.32} & \cellcolor{gray!12}18.61 & 13.62 & 36.10 & 6.10 & \cellcolor{gray!12}34.35 \\

      {Rank4Gen {\small(8B)}}& \cellcolor{gray!12}20.06 & 24.54 & 15.47 & 20.16 & \cellcolor{gray!12}62.96 & 9.03 & \colorbox{backred!50}{88.56} & 91.30 & \cellcolor{gray!12}18.03 & 10.72 & 38.59 & 4.79 & \cellcolor{gray!12}33.68 \\

      \midrule

      \rowcolor{gray!10}
      \multicolumn{14}{c}{\fontsize{10}{12}\selectfont \textit{\textbf{Average All Reranking Rounds}}} \\

      Only Search& \cellcolor{gray!12}17.76 & 24.74 & 11.65 & 16.89 & \cellcolor{gray!12}63.79 & 13.93 & \colorbox{backyellow_soft!40}{84.93} & 92.51 & \cellcolor{gray!12}17.41 & 13.04 & 32.88 & 6.31 & \cellcolor{gray!12}32.99 \\

      BGE-Reranker$_{L.}$& \cellcolor{gray!12}19.20 & 26.30 & 13.41 & 17.90 & \cellcolor{gray!12}62.55 & 15.51 & 80.28 & 91.85 & \cellcolor{gray!12}18.49 & 13.06 & 35.70 & 6.72 & \cellcolor{gray!12}33.41 \\

      MonoT5 {\small(3B)}& \cellcolor{gray!12}20.40 & 27.99 & 14.09 & 19.13 & \cellcolor{gray!12}\colorbox{backyellow_soft!40}{64.26} & 16.82 & 81.79 & 94.16 & \cellcolor{gray!12}19.43 & 13.64 & 36.66 & 7.99 & \cellcolor{gray!12}34.70 \\

      RankT5 {\small(3B)}& \cellcolor{gray!12}19.95 & 27.74 & 13.13 & 18.98 & \cellcolor{gray!12}62.78 & 16.98 & 80.84 & 90.51 & \cellcolor{gray!12}19.01 & 13.53 & 35.74 & 7.76 & \cellcolor{gray!12}33.91 \\

      Setwise {\small(7B)}& \cellcolor{gray!12}20.91 & 27.86 & \colorbox{backred!50}{15.09} & 19.77 & \cellcolor{gray!12}64.00 & 16.17 & 80.03 & \colorbox{backred!50}{95.79} & \cellcolor{gray!12}\colorbox{backyellow_soft!40}{20.28} & 13.70 & \colorbox{backred!50}{38.86} & \colorbox{backyellow_soft!40}{8.29} & \cellcolor{gray!12}\colorbox{backred!50}{35.06} \\

      RankLlama {\small(7B)}& \cellcolor{gray!12}19.19 & 24.96 & 14.31 & 18.30 & \cellcolor{gray!12}62.41 & 13.63 & 80.36 & 93.23 & \cellcolor{gray!12}17.94 & 12.88 & 34.17 & 6.76 & \cellcolor{gray!12}33.18 \\

      Rank1 {\small(7B)}& \cellcolor{gray!12}\colorbox{backred!50}{21.14} & \colorbox{backyellow_soft!40}{28.59} & 14.17 & \colorbox{backred!50}{20.67} & \cellcolor{gray!12}63.29 & \colorbox{backred!50}{19.26} & 79.22 & 91.39 & \cellcolor{gray!12}\colorbox{backred!50}{20.56} & 13.63 & \colorbox{backyellow_soft!40}{38.81} & \colorbox{backred!50}{9.23} & \cellcolor{gray!12}\colorbox{backyellow_soft!40}{35.00} \\

      {Rearank {\small(7B)}}& \cellcolor{gray!12}\colorbox{backyellow_soft!40}{20.94} & 28.32 & \colorbox{backyellow_soft!40}{14.82} & 19.69 & \cellcolor{gray!12}61.66 & 15.37 & 79.18 & 90.43 & \cellcolor{gray!12}19.48 & 13.52 & 37.87 & 7.06 & \cellcolor{gray!12}34.03 \\

      {ReasonRank {\small(7B)}}& \cellcolor{gray!12}20.69 & \colorbox{backred!50}{28.90} & 13.25 & \colorbox{backyellow_soft!40}{19.92} & \cellcolor{gray!12}\colorbox{backred!50}{64.30} & 17.30 & 81.21 & \colorbox{backyellow_soft!40}{94.40} & \cellcolor{gray!12}20.19 & \colorbox{backred!50}{13.87} & 38.48 & 8.21 & \cellcolor{gray!12}\colorbox{backred!50}{35.06} \\

      {SetR {\small(8B)}}& \cellcolor{gray!12}19.59 & 26.83 & 13.42 & 18.51 & \cellcolor{gray!12}62.99 & \colorbox{backyellow_soft!40}{17.74} & 77.22 & 94.02 & \cellcolor{gray!12}18.31 & \colorbox{backyellow_soft!40}{13.71} & 33.70 & 7.53 & \cellcolor{gray!12}33.63 \\

      {Rank4Gen {\small(8B)}}& \cellcolor{gray!12}20.59 & 28.09 & \colorbox{backyellow_soft!40}{14.82} & 18.85 & \cellcolor{gray!12}64.20 & 9.69 & \colorbox{backred!50}{88.78} & 94.13 & \cellcolor{gray!12}17.38 & 9.31 & 38.45 & 4.38 & \cellcolor{gray!12}34.06 \\

      \bottomrule
    \end{tabular}%
  }
\end{table*}

\subsection{Detailed Per-Round Analysis in Long-form Scenario}

Table~\ref{tab:longform_per_round} presents a fine-grained breakdown of reranker performance across search rounds in the Long-form scenario, reporting results at the first round, the last round, and averaged over all rounds. Several key observations emerge:

\textbf{(1) Systematic performance degradation across successive search rounds.} The most striking finding is that \textit{Overall scores fall from the first to the last round for almost every reranker}, the only exception being SetR, which stays essentially flat (34.31$\to$34.35). The best Overall score drops from 38.05 (Setwise, first round) to 35.87 (ReasonRank, last round), and Doc-Level averages decline without exception (e.g., Setwise: 24.99$\to$19.59; Rank1: 24.18$\to$20.48). This degradation does not reflect reranker failure per se, but rather a fundamental shift in retrieval dynamics: early rounds target broad information acquisition with rich candidate pools, while later rounds perform targeted gap-filling on increasingly narrow and difficult information needs. As the ``easy'' information is progressively consumed, remaining queries become harder to satisfy, yielding sparser candidate pools that systematically lower rubric coverage scores.

\textbf{(2) Degradation is most pronounced at Doc-Level and Global-Level.} Doc-Level averages drop by up to 5.7 points from first to last round (\eg RankLlama: 22.35$\to$16.61; BGE-Reranker: 22.68$\to$17.22), and Global-Level Reachability scores decline sharply (e.g., Setwise: 10.13$\to$8.94; Rank1: 11.60$\to$8.84). In contrast, Set-Level averages stay within a few points of their first-round values and in several cases even rise (\eg Rearank: 61.74$\to$65.16), with Redundancy and Conflict scores showing minimal variation. This asymmetry suggests that later-round documents remain internally coordinated but individually less relevant and insufficient for completing reasoning chains.

\textbf{(3) Rank4Gen maintains exceptional Set-Level dominance but sacrifices diversity.} Across all rounds, Rank4Gen consistently achieves the highest Redundancy scores (87.32, 88.56, 88.78) by a large margin, confirming that its generator-preference-aligned training is highly effective at deduplication. However, its Complementarity scores are consistently the lowest (11.99, 9.03, 9.69), revealing a critical trade-off: optimizing for generator preference may sacrifice information diversity.

\begin{figure*}[t!]
  \centering
    \includegraphics[width=1\linewidth]{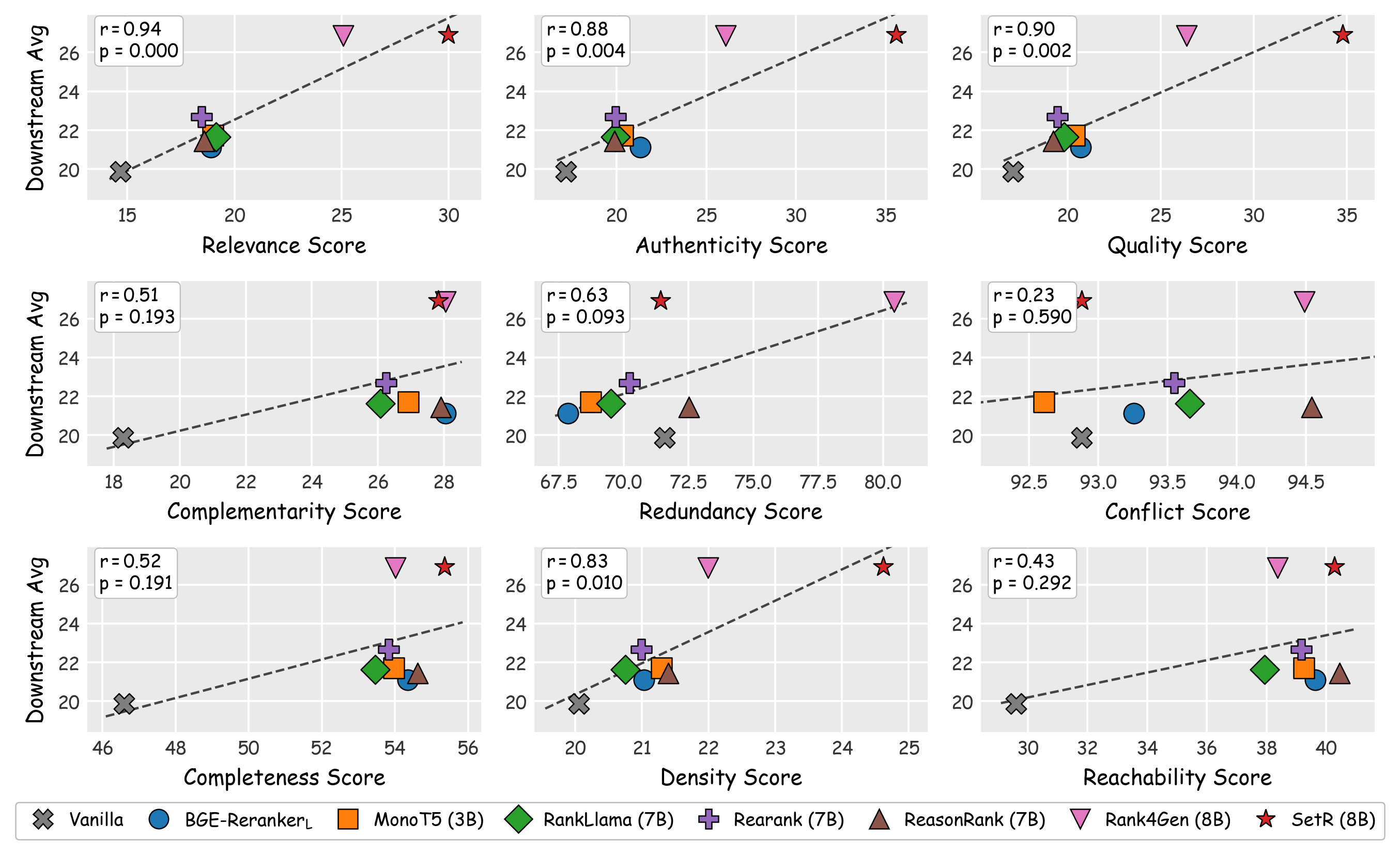}
  \caption{\textbf{Pearson correlation between individual rubric dimension scores and downstream generation quality (Downstream Avg).} Each subplot corresponds to one of the nine evaluation dimensions across three granularity levels (Doc-Level, Set-Level, Global-Level). Each point represents a reranking method. We report the Pearson correlation coefficient ($r$) and $p$-value for each dimension. Dimensions with $p < 0.05$ are considered statistically significant.}
\label{fig:correlation}

\end{figure*}

\textbf{(4) ReasonRank shows the strongest resilience to degradation.} In the average-all-rounds view, ReasonRank attains the best Overall score (35.06, tied with Setwise) and the highest Global-Level Completeness (Cov.=13.87). Notably, its drop from first to last round is among the smallest (Overall: 36.53$\to$35.87, $\Delta$=0.66), suggesting that reasoning capabilities help maintain selection quality even as candidate pools become sparser.

\textbf{(5) Global-Level Reachability exposes a universal ceiling.} Across all methods and rounds, Reachability scores remain extremely low (4.38--11.60), indicating that no current reranker can reliably assemble document sets supporting complete end-to-end reasoning chains. This represents the primary bottleneck for downstream generation quality in complex research scenarios, and it motivates the question we probe with \method, namely whether making the rubric criteria explicit at selection time recovers part of this headroom.

\subsection{Correlation Between Rubric Dimensions and Downstream Generation Quality}\label{appendix:correlation_analysis}

Figure~\ref{fig:correlation} presents the Pearson correlation between each rubric dimension's coverage score and the downstream generation quality (Downstream Avg) across all evaluated rerankers. We analyze the results at three granularity levels:

\textbf{(1) Doc-Level dimensions are strongly predictive of downstream generation quality.}
Relevance ($r\!=\!0.94$, $p\!<\!0.001$), Authenticity ($r\!=\!0.88$, $p\!=\!0.004$), and Quality ($r\!=\!0.90$, $p\!=\!0.002$) all exhibit strong positive correlations with statistical significance ($p<0.01$). This confirms that document-level quality constitutes the foundational prerequisite for effective generation: selecting relevant, authentic, and high-quality documents is the most direct lever for improving downstream performance.

\textbf{(2) Global-Level dimensions show differentiated predictive power.}
Density ($r\!=\!0.83$, $p\!=\!0.010$) achieves statistical significance, demonstrating that information density has a substantive impact on generation quality. In contrast, Completeness ($r\!=\!0.52$, $p\!=\!0.191$) and Reachability ($r\!=\!0.43$, $p\!=\!0.292$) show positive directional trends but fail to reach significance. This suggests that, given the current landscape of methods, these two dimensions do not exhibit sufficient variance across rerankers to statistically differentiate their performance.

\begin{table*}[t!]
  \centering
    \caption{\textbf{Per-question-type Overall Performance ($\%$) on \dataset{} (Short-form Scenario, agent-filter scoring).}
Each entry's Overall is the average of two independent LLM-judge scoring passes; per (ranker, type) we report the mean over entries (0--100 scale).
The top two performing results are highlighted in
\colorbox{backred!50}{red} (1\textsuperscript{st}) and
\colorbox{backyellow_soft!40}{yellow} (2\textsuperscript{nd}) backgrounds, respectively.
Type abbreviations: Br = 2-hop Bridge, PeC = Person-attribute Chain, SpC = Spatial / Geographic Chain, MiC = Miscellaneous Chain, Cmp = Comparison, BrC = Bridge + Comparison, Inf = Inference, Set = Set Operation.}
  \label{tab:type_overall_short_agent_filter}

  \vspace{-5pt}
  \renewcommand{\arraystretch}{0.8}
  \resizebox{\textwidth}{!}{%
    \begin{tabular}{l|cccc|cccc}
      \toprule
      \multirow{2.5}{*}{\textbf{Ranker}}
        & \multicolumn{4}{c|}{\textbf{Bridge Family}}
        & \multicolumn{4}{c}{\textbf{Non-Bridge}}
        \\
      \noalign{\global\aboverulesep=0pt \global\belowrulesep=0pt}
      \cmidrule(l{2pt}r{2pt}){2-5} \cmidrule(l{2pt}r{2pt}){6-9}
      \noalign{\global\aboverulesep=.65ex \global\belowrulesep=.65ex}
        & \textbf{Br} & \textbf{PeC} & \textbf{SpC} & \textbf{MiC} & \textbf{Cmp} & \textbf{BrC} & \textbf{Inf} & \textbf{Set} \\
      \midrule

      Only Retrieval & 22.0 & 10.9 & 9.1 & 9.1 & 31.7 & 9.3 & 14.1 & 19.9 \\

      \midrule
      \rowcolor{gray!10}
      \multicolumn{9}{c}{\fontsize{10}{12}\selectfont \textit{\textbf{Adhoc Reranking}}} \\

      BGE-Reranker$_{L.}$ & 27.0 & 12.7 & 12.4 & 15.1 & 43.7 & 12.1 & 17.9 & \colorbox{backyellow_soft!40}{25.4} \\

      MonoT5 {\small(3B)} & 27.0 & 13.7 & 13.3 & 14.3 & 43.6 & 13.0 & 18.0 & 24.3 \\

      RankT5 {\small(3B)} & 27.2 & 14.2 & 13.3 & 14.5 & 44.8 & 14.2 & 17.8 & 21.4 \\

      RankLlama {\small(7B)} & 27.1 & 14.2 & 13.4 & 14.8 & 42.9 & 14.2 & 18.1 & 22.1 \\

      RankVicuna {\small(7B)} & 25.5 & 12.7 & 11.4 & 12.5 & 35.9 & 11.7 & 16.9 & 23.3 \\

      RankZephyr {\small(7B)} & 25.8 & 13.2 & 10.7 & 12.4 & 32.7 & 11.9 & 18.7 & 19.9 \\

      Setwise {\small(7B)} & 26.8 & 13.3 & 12.9 & 15.4 & 44.1 & 13.0 & 18.4 & 22.4 \\

      \midrule
      \rowcolor{gray!10}
      \multicolumn{9}{c}{\fontsize{10}{12}\selectfont \textit{\textbf{Reasoning-Enhanced Reranking}}} \\

      Rank1 {\small(7B)} & 27.6 & 14.1 & 12.7 & 14.0 & 43.8 & 13.9 & 19.0 & 21.8 \\

      Rearank {\small(7B)} & 26.5 & 13.4 & 13.0 & 13.7 & 43.0 & 13.2 & 19.7 & 21.3 \\

      ReasonRank {\small(7B)} & 26.9 & 14.5 & 13.2 & 15.0 & 42.5 & 13.9 & 20.5 & 23.9 \\

      \midrule
      \rowcolor{gray!10}
      \multicolumn{9}{c}{\fontsize{10}{12}\selectfont \textit{\textbf{Setwise Reranking}}} \\

      SetR {\small(8B)} & \colorbox{backred!50}{30.5} & \colorbox{backyellow_soft!40}{14.9} & \colorbox{backyellow_soft!40}{14.4} & \colorbox{backred!50}{16.8} & \colorbox{backyellow_soft!40}{50.0} & \colorbox{backred!50}{16.5} & \colorbox{backred!50}{23.1} & 24.9 \\

      Rank4Gen {\small(8B)} & \colorbox{backyellow_soft!40}{29.8} & \colorbox{backred!50}{15.5} & \colorbox{backred!50}{14.6} & \colorbox{backyellow_soft!40}{15.6} & \colorbox{backred!50}{50.1} & \colorbox{backyellow_soft!40}{16.2} & \colorbox{backyellow_soft!40}{20.9} & \colorbox{backred!50}{26.3} \\

      \bottomrule
    \end{tabular}%
  }
\end{table*}

\textbf{(3) Set-Level dimensions exhibit weak and non-significant correlations.}
Complementarity ($r\!=\!0.51$, $p\!=\!0.193$), Redundancy ($r\!=\!0.63$, $p\!=\!0.093$), and Conflict ($r\!=\!0.23$, $p\!=\!0.590$) all fail to reach statistical significance. While seemingly counterintuitive, this result reveals a critical insight: all current rerankers cluster within an extremely narrow score range on Set-Level dimensions (e.g., Conflict scores span only 92.5 to 95.0), producing insufficient variance to manifest meaningful correlations. This does not imply that these dimensions are unimportant for generation quality; rather, it demonstrates that existing methods are universally deficient and highly homogeneous in their cross-document coordination capabilities, with no method yet achieving meaningful differentiation on set-level objectives.


\subsection{Performance across Fine-Grained Knowledge Types}

To understand where each ranker earns its aggregate score, we partition the benchmark along two orthogonal axes. Table~\ref{tab:type_overall_short_agent_filter} breaks the 2\,061 Short-form questions into eight reasoning types, grouped into a \emph{Bridge Family} pillar (\textbf{Br}: 2-hop bridge, \textbf{PeC}: person-attribute chain, \textbf{SpC}: spatial/geographic chain, \textbf{MiC}: miscellaneous chain) and a \emph{Non-Bridge} pillar (\textbf{Cmp}: comparison, \textbf{BrC}: bridge+comparison, \textbf{Inf}: inference, \textbf{Set}: set operation). Table~\ref{tab:field_overall_long_union} splits the Long-form queries across 16 research fields in four thematic pillars (Health, Life Sciences, Physical Sciences, Engineering). Each cell reports the mean of two independent LLM-judge scoring passes (0--100 Overall); the best and second-best rankers per column are highlighted.

Two key patterns emerge. \textbf{(i) Ranker separation is slice-dependent.} On Short-form, the Setwise family (SetR, Rank4Gen) leads Only Retrieval by $+8$ to $19$ points on the easier \textbf{Br} and \textbf{Cmp} types, yet the advantage shrinks to $+3$ to $5$ points on harder types (\textbf{PeC}, \textbf{SpC}, \textbf{MiC}) where all rerankers cluster tightly. The same pattern appears on Long-form: rerankers separate clearly on Health fields (PH, EM, EP) but nearly overlap on Materials Engineering and Sustainable Development. \textbf{(ii) No single ranker dominates all slices.} On Short-form, Setwise rerankers are consistently best or second-best across all eight types; on Long-form the leader varies: ReasonRank tops Epidemiology and Cell Biology, Rearank leads several Engineering fields, and SetR owns Hydrology and Public Health. This heterogeneity underscores the value of multi-slice evaluation over a single aggregate number.

Together, the two tables identify slices (person-attribute chains, spatial/geographic chains, and sustainability-related fields) where all current rerankers still leave substantial headroom for improvement.

\begin{table*}[t!]
  \centering
    \caption{\textbf{Per-domain Overall Performance ($\%$) on \dataset{} (Long-form Scenario, union scoring).}
Each entry's Overall is the average of two independent LLM-judge scoring passes; per (ranker, field) we report the mean of these per-entry Overalls (0--100 scale).
The top two performing results are highlighted in
\colorbox{backred!50}{red} (1\textsuperscript{st}) and
\colorbox{backyellow_soft!40}{yellow} (2\textsuperscript{nd}) backgrounds, respectively.
Field abbreviations: PH = Public Health, EM = Emergency Medicine, EP = Epidemiology, NS = Nutrition Science, MB = Microbiology, CB = Cell Biology, AG = Agronomy \& Crop Science, PE = Physical Education \& Sports Medicine, CH = Analytical Chemistry, PP = Polymers \& Plastics, HY = Hydrology \& Water Resources, GE = Geology, ME = Materials Engineering, EE = Electrical Engineering, FF = Forests \& Forestry, SD = Sustainable Development.}
  \label{tab:field_overall_long_union}

  \vspace{-5pt}
  \renewcommand{\arraystretch}{1.1}
  \resizebox{\textwidth}{!}{%
    \begin{tabular}{l|cccc|cccc|cccc|cccc}
      \toprule
      \multirow{2.5}{*}{\textbf{Ranker}}
        & \multicolumn{4}{c|}{\textbf{Health}}
        & \multicolumn{4}{c|}{\textbf{Life Sciences}}
        & \multicolumn{4}{c|}{\textbf{Physical Sciences}}
        & \multicolumn{4}{c}{\textbf{Engineering}}
        \\
      \noalign{\global\aboverulesep=0pt \global\belowrulesep=0pt}
      \cmidrule(l{2pt}r{2pt}){2-5} \cmidrule(l{2pt}r{2pt}){6-9} \cmidrule(l{2pt}r{2pt}){10-13} \cmidrule(l{2pt}r{2pt}){14-17}
      \noalign{\global\aboverulesep=.65ex \global\belowrulesep=.65ex}
        & \textbf{PH} & \textbf{EM} & \textbf{EP} & \textbf{NS} & \textbf{MB} & \textbf{CB} & \textbf{AG} & \textbf{PE} & \textbf{CH} & \textbf{PP} & \textbf{HY} & \textbf{GE} & \textbf{ME} & \textbf{EE} & \textbf{FF} & \textbf{SD} \\
      \midrule

      Only Search & 32.0 & 39.7 & 39.3 & 24.6 & 29.8 & 6.1 & 30.7 & 30.0 & 35.4 & 24.7 & 28.6 & 23.5 & 28.9 & 30.8 & 35.0 & 28.0 \\

      \midrule
      \rowcolor{gray!10}
      \multicolumn{17}{c}{\fontsize{10}{12}\selectfont \textit{\textbf{Adhoc Reranking}}} \\

      BGE-Reranker$_{L.}$ & 32.0 & 40.5 & 43.6 & 25.5 & 33.1 & 37.4 & 26.3 & 32.5 & 31.1 & \colorbox{backred!50}{32.4} & \colorbox{backred!50}{38.0} & 23.1 & 34.3 & 36.4 & 34.3 & 27.8 \\

      MonoT5 {\small(3B)} & 31.3 & \colorbox{backred!50}{46.4} & 50.1 & 28.5 & 32.0 & 13.8 & \colorbox{backred!50}{35.3} & 28.1 & 29.6 & \colorbox{backyellow_soft!40}{32.0} & 30.9 & 32.6 & 38.6 & 35.3 & 35.7 & 29.4 \\

      RankT5 {\small(3B)} & 28.5 & 40.3 & 50.7 & 34.6 & 34.8 & \colorbox{backyellow_soft!40}{43.6} & 25.9 & 33.4 & 28.7 & 26.7 & 30.3 & 30.8 & \colorbox{backyellow_soft!40}{40.3} & 30.5 & \colorbox{backyellow_soft!40}{40.9} & 27.4 \\

      RankLlama {\small(7B)} & 32.8 & 36.0 & \colorbox{backyellow_soft!40}{51.2} & \colorbox{backred!50}{42.8} & 33.3 & 18.1 & 26.7 & 14.6 & 23.7 & 29.5 & 31.5 & 31.0 & 35.7 & 30.8 & \colorbox{backred!50}{45.5} & 28.2 \\

      Setwise {\small(7B)} & 31.9 & 39.6 & 42.9 & 30.5 & 33.5 & 34.7 & \colorbox{backyellow_soft!40}{34.2} & \colorbox{backred!50}{36.5} & 28.0 & 28.6 & 30.0 & 23.8 & 35.4 & 30.5 & 36.9 & 28.6 \\

      \midrule
      \rowcolor{gray!10}
      \multicolumn{17}{c}{\fontsize{10}{12}\selectfont \textit{\textbf{Reasoning-Enhanced Reranking}}} \\

      Rank1 {\small(7B)} & 32.3 & 39.5 & \colorbox{backred!50}{56.9} & 27.4 & 33.5 & 31.1 & 28.4 & 26.4 & 31.2 & 29.5 & 28.5 & 28.7 & 39.2 & \colorbox{backyellow_soft!40}{39.9} & 39.7 & 29.1 \\

      Rearank {\small(7B)} & 32.3 & 32.7 & 47.4 & \colorbox{backyellow_soft!40}{37.2} & \colorbox{backred!50}{35.6} & 32.5 & 30.2 & \colorbox{backyellow_soft!40}{33.7} & 28.0 & 24.2 & 34.2 & \colorbox{backyellow_soft!40}{34.2} & 39.5 & 34.8 & 38.7 & \colorbox{backred!50}{32.0} \\

      ReasonRank {\small(7B)} & 29.6 & 38.2 & 46.8 & 29.9 & \colorbox{backyellow_soft!40}{35.6} & \colorbox{backred!50}{45.3} & 28.2 & 9.2 & \colorbox{backred!50}{43.9} & 31.4 & 30.8 & 32.0 & \colorbox{backred!50}{42.9} & \colorbox{backred!50}{41.5} & 35.4 & 28.4 \\

      \midrule
      \rowcolor{gray!10}
      \multicolumn{17}{c}{\fontsize{10}{12}\selectfont \textit{\textbf{Setwise Reranking}}} \\

      SetR {\small(8B)} & \colorbox{backred!50}{35.4} & 37.6 & 50.5 & 28.6 & 32.2 & 20.0 & 28.6 & 30.0 & \colorbox{backyellow_soft!40}{36.6} & 26.0 & 25.5 & \colorbox{backred!50}{35.1} & 30.5 & 28.4 & 31.5 & \colorbox{backyellow_soft!40}{31.6} \\

      Rank4Gen {\small(8B)} & \colorbox{backyellow_soft!40}{33.4} & \colorbox{backyellow_soft!40}{41.6} & 48.0 & 36.8 & 30.3 & 30.5 & 26.0 & 31.3 & 28.4 & 28.0 & \colorbox{backyellow_soft!40}{37.0} & 23.1 & 35.1 & 32.3 & 35.1 & 29.1 \\

      \bottomrule
    \end{tabular}%
  }
\end{table*}

\section{Case Study}

We showcase four representative cases, drawn from HotpotQA (Figure~\ref{fig:case_s1}), TriviaQA (Figure~\ref{fig:case_s2}) and ResearchQA (Figures~\ref{fig:case_l1} and~\ref{fig:case_l2}), to illustrate how our nine dimensional rubric surfaces diagnostic signals that a relevance only benchmark cannot expose. Each case reports the query, the reference answer, the full rubric text, the documents selected by the \texttt{bge-reranker-large} ranker, and the corresponding scores at the doc, set and global levels.

\nb{Case 1 for Short-form Scenario, Figure~\ref{fig:case_s1}).} Documents [6] and [9] each fully answer the question with the correct episode title and the Roy, Jim and Dwight event, so \textit{Rel} $=$ \textit{Aut} $=$ 4 with \textit{Qua} $=$ 4 and 3 respectively. Document [5] is a further excerpt of the same episode page as [6] but covers a later scene, so it is only marginally relevant (\textit{Rel} $=$ 1) and asserts nothing about the incident (\textit{Aut} $=$ 0). Documents [18] and [20] are off topic (a Dwight character biography and a different episode) and are correctly scored 0 across the board. The set level view then explains the redundancy and coverage picture: \textit{Cov} $=$ 4 because both [6] and [9] give the answer, \textit{Con} $=$ 4 because the two are consistent, and \textit{Red} $=$ 4 because the two answering documents add distinct context rather than duplicate each other. However, \textit{Cmp} drops to 1 because either [6] or [9] alone already suffices and there is no cross document division of labour, \textit{Den} drops to 1 because the answering sentence is a small fraction of each document's length, and \textit{Rea} drops to 2 because the documents never explicitly identify \emph{Jim} as \emph{Pam's husband}, breaking one link of the reasoning chain. Even a set with a $4/4$ retrieval outcome can therefore be structurally incomplete.

\nb{Case 2 for Short-form Scenario, Figure~\ref{fig:case_s2}).} Document [3] gives the answer verbatim; [2] duplicates the same sentence (both scored \textit{Rel} $=$ \textit{Aut} $=$ 4), which pulls \textit{Red} down to 2 because [3] and [2] carry the identical core fact. Documents [4] and [15] are the British Leyland conglomerate page and a Leyland Motors trucks and buses page respectively, off topic on both counts and correctly scored 0 across the board. Document [1] is a later section of the MG page that only alludes to ``small MG sports cars'' and never asserts the answer, hence \textit{Rel} $=$ 2 and \textit{Aut} $=$ 0. At the set level, \textit{Cov} $=$ 4 (the answer is present) and \textit{Con} $=$ 4 (no contradiction), while \textit{Cmp} $=$ 4 and \textit{Rea} $=$ 4 hold because [3] alone already carries both the BL affiliation and the two seat sports car reputation. This case demonstrates that redundancy at the top of a ranking is a wasted budget signal that our rubric penalises but relevance only metrics do not.

\nb{Case 1 for Long-form Scenario, Figure~\ref{fig:case_l1}).} Documents [D1] through [D4], together with [D6] and [D8], are all authentic and moderately relevant with per document averages near $1.5$ to $2$, but none of them is a comprehensive MQL primer, so per document scores plateau below 4. Documents [D5], [D7] and [D9] are review or scope excerpts that do not describe the MQL mechanism or its quantitative numbers, and are correctly scored 0. At the set level, no cross document contradiction is present (\textit{Con} $=$ 4), yet the collection is only moderately complementary (\textit{Cmp} $=$ 2) and moderately redundant (\textit{Red} $=$ 2) because most documents recycle the same generic definition. The global level tells the real story: coverage of the mechanism, quantity and tool life triple is very low (${\textit{Cov}} \approx 0.7$) and \textit{Rea} $=$ 1, meaning that the union set cannot chain from ``what MQL is'' through the quantitative sustainability payoffs the reference answer requires. This is a diagnosis that a relevance only ranker would miss entirely.

\nb{Case 2 for Long-form Scenario, Figure~\ref{fig:case_l2}).} Most documents in the union are on topic (\textit{Rel} $\approx$ 2 to 3), yet only [D3] and [D8] accurately describe the ChopValue pipeline, with \textit{Aut} and \textit{Qua} peaking at [D8]. Documents [D2] and [D7] are generic bamboo pieces with \textit{Aut} $=$ 0, while [D6] and [D9] are largely duplicated interview coverage. At the set level, \textit{Cmp} $=$ 3 because the process oriented and product oriented pieces do complement each other, and \textit{Con} $=$ 4 because no document contradicts another, but \textit{Red} $=$ 1 because the interview and headline pieces recycle the same ``millions of chopsticks'' claim. The global level exposes the shortfall: \textit{Cov} $=$ \textit{Den} $=$ \textit{Rea} $=$ 1, because the union never simultaneously nails collection, sterilisation, resin coating, hot pressing, CNC finishing and the product line, so a downstream reasoner cannot reconstruct the full circular economy chain from the union alone.

\nb{Takeaways.}
Across the four cases, three cross cutting patterns emerge. First, doc level scores are informative but insufficient: Cases S1 and L1 both contain documents that individually or in aggregate look ``high relevance'' yet still fail at the set or global level. Second, set level signals such as \textit{Red} and \textit{Cmp} catch redundant top ranked documents (as with [3] and [2] in S2, and [D6] and [D9] in L2) and poorly complementary sets (L1, L2) that a plain relevance view would miss. Third, global level \textit{Cov}, \textit{Den} and \textit{Rea} capture whether the union actually supports the full reasoning chain needed by the answer, as illustrated by the broken Pam and Jim link in S1 and the missing pipeline coverage in L1 and L2, which is precisely the property that governs downstream generation quality. These diagnostic capabilities motivate the multi level rubric design and support the conclusion, reported in the main text, that per round rubric scoring understates reranker value while union scoring provides a fairer set level evaluation.


\begin{figure*}[t]
\begin{redbox}{\textit{Case 1 for Short-form Scenario}}
\scriptsize
\setlength{\parskip}{1pt}
\setlength{\parindent}{0pt}

\bfit{Query:} In what episode of \emph{The Office} does Dwight save Pam's husband from Roy?\\
\bfit{Gold answer:} ``The Negotiation''

\vspace{2pt}
\hdashrule{\linewidth}{0.4pt}{2pt 2pt}

\vspace{1pt}
\bfit{Rubric:}

\vspace{1pt}
\textbf{[Rel]}
``Does each document directly discuss the specific episode of The Office where Dwight intervenes to save Pam's husband (Jim) from an attack by Roy (e.g., by using pepper spray), rather than merely mentioning the show or these characters' names without addressing this specific altercation?''

\textbf{[Aut]}
``For each document that states the episode where Dwight saves Pam's husband from Roy, is the episode correctly identified as `The Negotiation' and not a different episode title?'';
``For each document describing the scene where Roy attacks Jim and Dwight intervenes, are the details (Dwight stopping Roy and the incident occurring in the context of a negotiation over Roy's raise) consistent with the episode `The Negotiation' and not attributed to another episode?''

\textbf{[Qua]}
``Does each document have clear structure and organization, enabling a reader to directly extract the episode title `The Negotiation' and the specific plot point that Dwight saves Pam's husband from Roy, without the key information being buried, implied only, or overshadowed by disorganized, overly broad, or irrelevant content?''

\textbf{[Cmp]}
``Does the document set, taken together, collectively provide both the plot detail that Dwight saves Pam's husband (Jim) from Roy AND the identification of the episode as `The Negotiation', so that the event can be matched to the correct episode title?''

\textbf{[Red]}
``Does the document set avoid redundancy where multiple documents repeat the same core information (e.g., merely stating that the episode where Dwight saves Pam's husband from Roy is `The Negotiation', or providing identical descriptions of the event) without offering any incremental details such as season/episode number, airing order, or plot context?''

\textbf{[Con]}
``Across the document set, are the statements about the episode title mutually consistent, with no document assigning the event to a different episode than `The Negotiation' (e.g., one claiming it happened in `The Return' and another in `The Negotiation'), thereby supporting the correct answer without contradiction?''

\textbf{[Cov]}
``Does the document set as a whole explicitly state that the episode of The Office where Dwight saves Pam's husband (Jim) from Roy is titled `The Negotiation'?''

\textbf{[Den]}
``Across the documents in the set, does the text that explicitly discusses the incident where Dwight saves Pam's husband from Roy in the episode `The Negotiation' of The Office make up the majority of each document's length, rather than each document being dominated by unrelated content (e.g., general series information, other episode plots, or lengthy character biographies) where this specific event and episode title are only a small fraction of the text?''

\textbf{[Rea]}
``Using only the document set and no external knowledge, can the full reasoning chain be completed: identifying Pam's husband as Jim, establishing that Dwight saves Jim from an attack by Roy, and linking this specific event to the episode titled `The Negotiation'?''

\vspace{2pt}
\hdashrule{\linewidth}{0.4pt}{2pt 2pt}

\vspace{1pt}
\bfit{Ranker} \hfill Selected docs (in ranker order): \texttt{[6, 9, 5, 18, 20]}

\vspace{1pt}
\textbf{[6] \emph{The Negotiation}.}
``The Negotiation'' (originally titled ``Labor Negotiation'') is the nineteenth episode of the third season of the American comedy television series \emph{The Office}, and the show's forty-seventh episode overall. In this episode, \textbf{Roy Anderson tries to attack Jim Halpert for kissing Pam Beesly on Casino Night, only to be pepper-sprayed by Dwight Schrute.} Jim repeatedly tries to thank Dwight for his actions, but each attempt is rejected. Meanwhile, with Roy fired,~\ldots

\textbf{[9] \emph{Pam Beesly}.}
\ldots\ with Roy, tells him about Jim kissing her at ``Casino Night''. Roy yells, smashes a mirror, and trashes the bar. Pam, frightened and embarrassed by his reaction, breaks up with Roy immediately. Roy vows to kill Jim and \textbf{in ``The Negotiation'', Roy unsuccessfully tries to attack Jim at work (Jim is saved by Dwight's intervention), and is subsequently fired.} Pam later reluctantly agrees to meet Roy for coffee,~\ldots

\textbf{[5] \emph{The Negotiation}.}
\ldots\ asks Pam to join him for coffee. Their meeting at a local diner is awkward and ends with their relationship appearing to be over; they hug and Pam wipes a tear from her face. Roy encourages Pam to pursue her feelings for Jim but she says she will not try to get him. Angela Martin interviews people from the office repeatedly to hear the tale of Dwight's heroics.~\ldots \emph{[Same episode page as [6], but this excerpt covers a later scene and does not restate the Roy/Jim/Dwight incident.]}

\textbf{[18] \emph{Dwight Schrute}.}
``Diwali'', he comforts a tearful Pam, and in ``China'', he secretly allows Pam to save face when she feels vulnerable about her job abilities. In ``The Job'', Dwight offers Pam the position of ``Secret Assistant to the Regional Manager''. The two briefly become best friends while he suffers a concussion in ``The Injury''.~\ldots \emph{[Dwight character biography; never touches the Roy attack.]}

\textbf{[20] \emph{Michael Scott Paper Company}.}
``Michael Scott Paper Company'' is the twenty-third episode of the fifth season. In the episode, Michael, Pam and Ryan try to get their new paper company off the ground, but end up bickering among themselves due to the stress and cramped office space.~\ldots \emph{[A different episode; entirely off-topic.]}

\vspace{2pt}
\hdashrule{\linewidth}{0.4pt}{2pt 2pt}

\vspace{1pt}
\bfit{Rubric scores.}

\vspace{2pt}
\begin{center}\scriptsize
\setlength{\tabcolsep}{4pt}
\renewcommand{\arraystretch}{1.0}
\begin{tabular}{@{}l | ccc | ccc | ccc @{}}
\toprule
 & \multicolumn{3}{c|}{\textbf{Doc-Level}} & \multicolumn{3}{c|}{\textbf{Set-Level}} & \multicolumn{3}{c}{\textbf{Global-Level}} \\
 & \textbf{Rel.} & \textbf{Aut.} & \textbf{Qua.}
 & \textbf{Cmp.} & \textbf{Red.} & \textbf{Con.}
 & \textbf{Cov.} & \textbf{Den.} & \textbf{Rea.} \\
\midrule
Score & 4\,/\,4\,/\,1\,/\,0\,/\,0 & 4\,/\,4\,/\,0\,/\,0\,/\,0 & 4\,/\,3\,/\,0\,/\,0\,/\,0
      & 1 & 4 & 4
      & 4 & 1 & 2 \\
\bottomrule
\end{tabular}
\end{center}

\end{redbox}

\vspace{-6pt}
\caption{\textbf{Case 1 for Short-form Scenario.}}
\label{fig:case_s1}
\end{figure*}


\begin{figure*}[t]
\begin{redbox}{\textit{Case 2 for Short-form Scenario}}
\scriptsize
\setlength{\parskip}{1pt}
\setlength{\parindent}{0pt}

\bfit{Query:} Which of the former British Leyland companies is best known for its two-seat open sports cars?\\
\bfit{Gold answer:} ``MG Car Company Limited''

\vspace{2pt}
\hdashrule{\linewidth}{0.4pt}{2pt 2pt}

\vspace{1pt}
\bfit{Rubric:}

\vspace{1pt}
\textbf{[Rel]}
``Does each document explicitly discuss former British Leyland companies (or MG Car Company Limited specifically) and directly address the production, reputation, or historical significance of two-seat open sports cars, rather than merely mentioning British Leyland, MG, or sports cars in general without connecting them to the specific model type and company?''

\textbf{[Aut]}
``For each document that identifies a former British Leyland company as being best known for two-seat open sports cars, is the company correctly identified as MG (MG Car Company Limited), rather than incorrectly attributing this distinction to another brand (e.g., Triumph, Austin, Rover)?''

\textbf{[Qua]}
``Is each document clearly structured and focused enough that a reader can directly extract the fact that MG Car Company Limited is the former British Leyland company best known for two-seat open sports cars, without the key information being buried under unrelated historical or technical detail?''

\textbf{[Cmp]}
``Does the document set, taken as a whole, collectively provide information that establishes MG Car Company Limited as a former British Leyland company AND information that confirms its primary reputation is for two-seat open sports cars, such that the documents jointly supply the complete context needed to answer the query?''

\textbf{[Red]}
``Does the document set avoid having multiple documents that repeat the same core fact that MG Car Company Limited is a former British Leyland company known for two-seat open sports cars, without providing any incremental, distinguishing, or contextual information (e.g., other British Leyland companies, specific MG sports car models, or historical context)?''

\textbf{[Con]}
``Does the document set avoid conflicting claims about which former British Leyland company is best known for two-seat open sports cars, consistently identifying MG Car Company Limited rather than any other marque (e.g., Triumph, Austin)?'';
``Does the document set avoid factual contradictions regarding corporate history, such as conflicting statements about whether MG Car Company Limited was ever part of British Leyland, ensuring background facts consistently support the answer?''

\textbf{[Cov]}
``Does the document set as a whole establish both that MG Car Company Limited was a former British Leyland company and that it is best known for its two-seat open sports cars, with no missing information that would prevent identifying it as the correct answer?''

\textbf{[Den]}
``Across the documents in the set, does the text that identifies which former British Leyland company is best known for its two-seat open sports cars---specifically naming MG Car Company Limited, describing its association with British Leyland, and detailing its reputation for two-seat open sports cars---make up the majority of each document's length, rather than the documents being dominated by unrelated content?''

\textbf{[Rea]}
``Using only the document set and no external knowledge, can the full reasoning chain be completed: (1) verify that MG Car Company Limited is a former British Leyland company, (2) confirm that it is known for producing two-seat open sports cars, and (3) conclude that MG Car Company Limited is the answer?''

\vspace{2pt}
\hdashrule{\linewidth}{0.4pt}{2pt 2pt}

\vspace{1pt}
\bfit{Ranker:} \hfill Selected docs (in ranker order): \texttt{[3, 2, 4, 15, 1]}

\vspace{1pt}
\textbf{[3] \emph{MG Cars}.}
MG, the initials of Morris Garages, is a British automotive marque registered by the now defunct \textbf{MG Car Company Limited}, a British sports car manufacturer begun in the 1920s. \textbf{Best known for its two-seat open sports cars,} MG also produced saloons and coup\'es.~\ldots

\textbf{[2] \emph{MG Cars}.}
\ldots\ MG came second in the Constructors Championship in 2015~\ldots. MG, the initials of Morris Garages, is a British automotive marque registered by the now defunct \textbf{MG Car Company Limited}, a British sports car manufacturer begun in the 1920s~\ldots \textbf{Best known for its two-seat open sports cars,} MG~\ldots \emph{[Same page as [3]; the answer sentence is essentially duplicated with only tangential BTCC context added.]}

\textbf{[4] \emph{British Leyland}.}
British Leyland was an automotive engineering and manufacturing conglomerate formed in the United Kingdom in 1968~\ldots. It incorporated much of the British-owned motor vehicle industry~\ldots Despite containing profitable marques such as Jaguar, Rover and Land Rover, as well as the best-selling Mini, British Leyland had~\ldots \emph{[General history of the BL conglomerate; MG is not mentioned in the excerpt.]}

\textbf{[15] \emph{Leyland Motors}.}
Leyland Motors Limited (later Leyland Motor Corporation) was a British vehicle manufacturer of \textbf{lorries, buses and trolleybuses}. The company diversified into car manufacturing with its acquisitions of Triumph and Rover in 1960 and 1967~\ldots \emph{[Different company (trucks/buses); does not address MG or two-seat sports cars.]}

\textbf{[1] \emph{MG Cars}.}
\ldots\ Long-time service manager John Thornley took over as general manager~\ldots. Under BMC, several MG models were no more than badge-engineered versions of other marques, with the main exception being \textbf{the small MG sports cars.} BMC took over Jaguar Cars in September 1966~\ldots BMH joined with Leyland Motor Corporation in 1968 to form British Leyland Motor Corporation (BLMC).~\ldots \emph{[MG page but a later section; mentions ``small MG sports cars'' but never asserts MG is ``best known for two-seat open sports cars''.]}

\vspace{2pt}
\hdashrule{\linewidth}{0.4pt}{2pt 2pt}

\vspace{1pt}
\bfit{Rubric scores.}

\vspace{2pt}
\begin{center}\scriptsize
\setlength{\tabcolsep}{4pt}
\renewcommand{\arraystretch}{1.0}
\begin{tabular}{@{}l | ccc | ccc | ccc @{}}
\toprule
 & \multicolumn{3}{c|}{\textbf{Doc-Level}} & \multicolumn{3}{c|}{\textbf{Set-Level}} & \multicolumn{3}{c}{\textbf{Global-Level}} \\
 & \textbf{Rel.} & \textbf{Aut.} & \textbf{Qua.}
 & \textbf{Cmp.} & \textbf{Red.} & \textbf{Con.}
 & \textbf{Cov.} & \textbf{Den.} & \textbf{Rea.} \\
\midrule
Score & 4\,/\,4\,/\,0\,/\,0\,/\,2 & 4\,/\,4\,/\,0\,/\,0\,/\,0 & 4\,/\,3\,/\,0\,/\,0\,/\,0
      & 4 & 2 & 4
      & 4 & 1 & 4 \\
\bottomrule
\end{tabular}
\end{center}

\end{redbox}
\vspace{-6pt}
\caption{\textbf{Case 2 for Short-form Scenario.}}
\label{fig:case_s2}
\end{figure*}



\begin{figure*}[t]
\begin{greenbox}{\textit{Case 1 for Long-form Scenario}}
\scriptsize
\setlength{\parskip}{1pt}
\setlength{\parindent}{0pt}

\bfit{Query:} What is Minimum Quantity Lubrication (MQL), and how does it promote sustainable machining practices?\\
\bfit{Reference answer:}
``MQL, also termed near-dry machining, delivers atomized lubricant to the cutting tool interface via compressed air, using only 2--4~oz per shift (vs.~gallons of flood coolant), forming a micro-thin film that reduces friction and heat. It cuts coolant volume by $\sim$98\%, hazardous waste and disposal, saves 15--20\% energy, extends tool life by 25--80\%, and improves worker health.''

\vspace{2pt}
\hdashrule{\linewidth}{0.4pt}{2pt 2pt}

\vspace{1pt}
\bfit{Rubric:}

\vspace{1pt}
\textbf{[Rel]}
``Does each document clearly describe what MQL (near-dry machining) is and how it is implemented at the cutting tool--workpiece interface (atomized lubricant via compressed air, a few ounces per shift, forming a micro-thin film)?''; \emph{$+$ 1 more rubric on environmental / energy / waste / worker-health benefits.}

\textbf{[Aut]}
``Does each document correctly characterize MQL as delivering atomized lubricant via compressed air in drastically reduced volumes ($\sim$2--4~oz per shift, $\sim$98\% less than flood coolant), rather than as a flood/immersion method?''; \emph{$+$ 1 more rubric on sustainability-claim accuracy.}

\textbf{[Qua]}
``Is each document clearly structured so that a reader can extract a definition of MQL and distinguish it from flood cooling, including delivery mechanism and approximate quantities?''; \emph{$+$ 1 more rubric on extractability of sustainability benefits.}

\textbf{[Cmp]}
``Does the document set, taken together, cover both the fundamental mechanism of MQL (near-dry, compressed-air-atomised, 2--4~oz/shift, micro-thin film) \emph{and} its diverse sustainability benefits---environmental ($\sim$98\% coolant reduction, biodegradable oils, 15--20\% energy savings), workplace-health (no aerosolized mist), and operational (dry recyclable chips, 25--80\% tool-life extension)?''

\textbf{[Red]}
``Does the document set avoid redundancy where two or more documents merely restate the basic MQL definition or the $\sim$98\% coolant-reduction figure without contributing incremental aspects (energy metrics, biodegradable oils, worker-health benefits, tool-life quantification, ISO 14001, etc.)?''

\textbf{[Con]}
``Are the documents free of contradiction on MQL's fundamental nature ($\sim$2--4~oz atomized vs.~flood coolant)?''; \emph{$+$ 2 more rubrics on sustainability impacts and tool-life claims.}

\textbf{[Cov]}
``Is the definition of MQL as near-dry machining, delivered via compressed air, with $\sim$2--4~oz/shift instead of gallons of flood coolant, clearly present?''; \emph{$+$ 2 more rubrics on cooling mechanism (micro-thin film, no quenching) and sustainability payoffs.}

\textbf{[Den]}
``Does the text that actually describes MQL---its definition, atomized delivery, 2--4~oz vs.~gallons comparison, and micro-thin film---occupy the majority of each document's length?''; \emph{$+$ 1 more rubric on sustainability-specific text density.}

\textbf{[Rea]}
``Using only the document set and no external knowledge, can the full reasoning chain be completed---from what MQL is, through its coolant-volume/energy/tool-life/health benefits, to why it constitutes sustainable machining?''

\vspace{2pt}
\hdashrule{\linewidth}{0.4pt}{2pt 2pt}

\vspace{1pt}
\bfit{Ranker:} \hfill Selected docs (union of 2 rounds, in ranker order): \texttt{D1..D9}

\vspace{1pt}
\textbf{[D1]} \emph{Minimum Quantity Lubrication --- an overview.} ``\ldots MQL is a sustainable technique which utilizes very less amount of fluid for the lubrication and cooling\ldots'' \emph{[Textbook-style definition; touches sustainability at high level.]}

\textbf{[D2]} \emph{A comprehensive review on the viability of minimum quantity lubrication.} ``The implementation of MQL techniques allows decreased cutting fluid usage, which results in substantial cost reductions\ldots'' \emph{[Review; overlaps heavily with D1 on definition.]}

\textbf{[D3]} \emph{Assessing the cooling/lubricating agencies for sustainable\ldots\ (Makhesana~2024).} ``In the MQL technique, small quantities of fluid (5--\ldots) are used\ldots'' \emph{[Introduces a specific MQL quantity range; still mostly generic.]}

\textbf{[D4]} \emph{Minimum Quantity Lubrication for Sustainable Machining.} ``This significantly reduces energy consumption and costs associated with coolant disposal, reclamation, filtration, and chilling\ldots'' \emph{[Adds an energy/disposal angle.]}

\textbf{[D5]} \emph{Eco-Friendly Cutting Fluids in MQL\ldots\ (Sen~2021).} ``Review paper presents a summary of previously published research\ldots'' \emph{[Only a meta-review sentence; does not describe MQL mechanism or numbers.]}

\textbf{[D6]} \emph{``Assessing the cooling/lubricating agencies\ldots'' (same paper as D3, round 2).} ``Sustainability analysis revealed 6.1--36.4\% cost decreases\ldots'' \emph{[Same paper, later section; adds a cost figure but repeats scope.]}

\textbf{[D7]} \emph{A State-of-the-Art Review on Recently Developed Sustainable\ldots} ``Reviews sustainable machining under different lubrication approaches for MMCs\ldots'' \emph{[Off-topic scope (MMCs), doesn't describe MQL specifically.]}

\textbf{[D8]} \emph{A review on MQL for sustainable machining.} ``MQL can lead to a 40\% reduction in cutting force and 36\% decrease in cutting temperature\ldots'' \emph{[Adds concrete cutting-force/temperature numbers.]}

\textbf{[D9]} \emph{(PDF) Assessing the cooling/lubricating agencies\ldots\ (same paper as D3, round 2).} ``Machinability and comparative sustainability analysis of Nimonic 80 superalloy\ldots'' \emph{[Application-specific excerpt; doesn't restate MQL definition.]}

\vspace{2pt}
\hdashrule{\linewidth}{0.4pt}{2pt 2pt}

\vspace{1pt}
\bfit{Rubric scores}

\vspace{2pt}
\begin{center}\scriptsize
\setlength{\tabcolsep}{3pt}
\renewcommand{\arraystretch}{1.0}
\begin{tabular}{@{}l | ccc | ccc | ccc @{}}
\toprule
 & \multicolumn{3}{c|}{\textbf{Doc-Level}} & \multicolumn{3}{c|}{\textbf{Set-Level}} & \multicolumn{3}{c}{\textbf{Global-Level}} \\
 & \textbf{Rel.} & \textbf{Aut.} & \textbf{Qua.}
 & \textbf{Cmp.} & \textbf{Red.} & \textbf{Con.}
 & \textbf{Cov.} & \textbf{Den.} & \textbf{Rea.} \\
\midrule
Score & 2/1.5/1.5/2/0/1.5/0.5/1.5/0 & 2/2/2/2/0/1.5/0/1/0 & 2/2/2/2/0/2/0/2/0
      & 2 & 2 & 4
      & 0.7 & 1.5 & 1 \\
\bottomrule
\end{tabular}
\end{center}

\end{greenbox}
\vspace{-6pt}
\caption{\textbf{Case 1 for Long-form Scenario.}}
\label{fig:case_l1}
\end{figure*}

\begin{figure*}[t]
\begin{greenbox}{\textit{Case 2 for Long-form Scenario}}
\scriptsize
\setlength{\parskip}{1pt}
\setlength{\parindent}{0pt}

\bfit{Query:} How does Chop Value upcycle bamboo chopsticks into functional interior products, and what specific processes do they employ?\\
\bfit{Reference answer:}
``ChopValue collects used bamboo chopsticks, sterilizes them, coats them in a water-based food-safe resin, and hot-presses them at high pressure into blocks that are then CNC-cut and finished into tiles, tabletops, shelving, and homeware---realising a circular-economy pipeline.''

\vspace{2pt}
\hdashrule{\linewidth}{0.4pt}{2pt 2pt}

\vspace{1pt}
\bfit{Rubric:}

\vspace{1pt}
\textbf{[Rel]}
``Does each document explicitly discuss ChopValue's process of upcycling used bamboo chopsticks into functional interior products (tiles, tabletops, homeware), rather than only mentioning bamboo, chopsticks, or upcycling in general?''; \emph{$+$ 2 more rubrics on collection logistics and finished-product types.}

\textbf{[Aut]}
``For each document that describes ChopValue's process, does it correctly attribute the key steps (sterilisation, water-based resin coating, hot-pressing at high pressure, CNC finishing) and not conflate them with unrelated bamboo-panel manufacturers?''; \emph{$+$ 2 more rubrics.}

\textbf{[Qua]}
``Is each document structured so that a reader can directly extract ChopValue's specific processes and product line, without the key information being buried under generic circular-economy narrative?''; \emph{$+$ 2 more rubrics.}

\textbf{[Cmp]}
``Does the document set, taken together, cover both the \emph{upstream} pipeline (chopstick collection, sterilisation, resin coating, hot-pressing) AND the \emph{downstream} product range (tiles, tabletops, shelving, homeware) with associated volumes / partners?''

\textbf{[Red]}
``Does the document set avoid redundancy where multiple documents merely repeat the ``chopsticks $\to$ tiles'' headline without adding process detail (press temperature, resin type, CNC finishing, product SKUs)?''

\textbf{[Con]}
``Does the document set consistently describe ChopValue's process, with no document contradicting a key step (e.g., not falsely stating that the chopsticks are chemically dissolved rather than pressed)?''

\textbf{[Cov]}
``Does the document set as a whole explicitly cover: (a) chopstick collection network, (b) sterilisation, (c) resin coating, (d) hot-pressing, (e) CNC/finishing, and (f) the product line?''

\textbf{[Den]}
``Across the documents in the set, does the text that actually describes ChopValue's chopstick-upcycling process and product line occupy the majority of each document's length, rather than each document being dominated by general circular-economy or bamboo-industry content?''

\textbf{[Rea]}
``Using only the document set and no external knowledge, can the full reasoning chain be completed---from ChopValue's chopstick intake, through sterilisation/resin/press processing, to the finished functional interior products?''

\vspace{2pt}
\hdashrule{\linewidth}{0.4pt}{2pt 2pt}

\vspace{1pt}
\bfit{Ranker:} \hfill Selected docs (union of 2 rounds, in ranker order): \texttt{D1..D9}

\vspace{1pt}
\textbf{[D1]} \emph{ChopValue: how used chopsticks become furniture.} Describes chopstick collection from restaurants, sterilisation, and pressing into ``modules''; light on downstream product range.

\textbf{[D2]} \emph{Bamboo as a sustainable material (overview).} General bamboo-industry article; only tangentially mentions ChopValue.

\textbf{[D3]} \emph{ChopValue microfactory model.} Describes the microfactory hub-and-spoke network, sterilisation, and hot-pressing at high pressure with a water-based resin. \emph{[Most process-specific doc in the set.]}

\textbf{[D4]} \emph{Circular-economy startups in Vancouver.} Uses ChopValue as one example; lists product categories (tiles, tabletops) but few process details.

\textbf{[D5]} \emph{ChopValue product catalogue (tiles / shelving / homeware).} Enumerates the downstream products but not the manufacturing steps.

\textbf{[D6]} \emph{ChopValue interview / press piece.} Repeats the ``millions of chopsticks upcycled'' headline; adds partner-restaurant volume figures.

\textbf{[D7]} \emph{Bamboo composite panels (industry report).} Talks about bamboo composite panels in general; touches ChopValue only briefly.

\textbf{[D8]} \emph{ChopValue technical whitepaper.} Best-detailed doc on resin chemistry, press temperature/pressure, and CNC-finishing---the closest thing in the set to a full pipeline description.

\textbf{[D9]} \emph{ChopValue interview (round 2).} Same personality profile as D6; largely duplicates D6's content with different anecdotes.

\vspace{2pt}
\hdashrule{\linewidth}{0.4pt}{2pt 2pt}

\vspace{1pt}
\bfit{Rubric scores}

\vspace{2pt}
\begin{center}
\resizebox{0.92\linewidth}{!}{%
\scriptsize
\setlength{\tabcolsep}{2.5pt}
\renewcommand{\arraystretch}{1.0}
\begin{tabular}{@{}l | ccc | ccc | ccc @{}}
\toprule
 & \multicolumn{3}{c|}{\textbf{Doc-Level}} & \multicolumn{3}{c|}{\textbf{Set-Level}} & \multicolumn{3}{c}{\textbf{Global-Level}} \\
 & \textbf{Rel.} & \textbf{Aut.} & \textbf{Qua.}
 & \textbf{Cmp.} & \textbf{Red.} & \textbf{Con.}
 & \textbf{Cov.} & \textbf{Den.} & \textbf{Rea.} \\
\midrule
Score & 2.7/0.7/2/2.7/2.7/2.7/1.7/3/2.3 & 0/0/1.3/0/1/1/1/2.3/0.7 & 1.3/0/2.3/1.7/1.7/0.3/0/2.3/2.3
      & 3 & 1 & 4
      & 1 & 1 & 1 \\
\bottomrule
\end{tabular}%
}
\end{center}

\end{greenbox}
\vspace{-6pt}
\caption{\textbf{Case 2 for Long-form Scenario.}}
\label{fig:case_l2}
\end{figure*}
\clearpage
\section{Model Prompts}

\prompt
{\method Prompt}{
\systemprompt{
Your task is to select the minimum set of passages that can fully support answering the search query below.\\
Search Query: \prompttag{\textbf{\color{insert}<QUERY>}}\\
Rubric --- A structured checklist of information requirements. Each rubric item has three fields:\\
- level: evaluation granularity (L1 = per-document quality, L2 = cross-document set quality, L3 = end-to-end answerability).\\
- type: the specific quality dimension being checked (e.g., Relevance, Complementarity, Completeness, Density, Reachability).\\
- item: a yes/no question that the selected passage set should satisfy.\\
\prompttag{\textbf{\color{insert}<RUBRIC>}}\\
Candidate Passages (each indicated by a numerical identifier []):\\
\prompttag{\textbf{\color{insert}<CONTEXT>}}\\
Please follow the steps below:\\
Step 1. For each rubric item, list which passage(s) contain information that satisfies it. If no passage satisfies a rubric item, write ``NOT FOUND''.\\
Step 2. Select the minimum set of passages that together satisfy ALL rubric items. Each selected passage must contribute at least one rubric item that other selected passages do not cover.\\
Step 3. If some rubric items are ``NOT FOUND'', still include the closest matching passages that partially address those items.\\
Step 4. Output the selected passages. The format of final output should be `\#\#\# Final Selection: [] [].$\backslash$n', e.g., `\#\#\# Final Selection: [2] [1].$\backslash$n'.
}}%


\prompt
{Rubric Generation Prompt for Short-form Scenario: Doc-Level}{
\systemprompt{
You are a senior rubric-generation expert. Given a (query, answer) pair, you generate binary (Yes/No) evaluation rubrics along several specific dimensions. These rubrics are used to judge, document by document, whether a candidate document set can effectively support answering the query and producing the correct answer.\\
{\# Rubric requirements}\\
Every rubric you generate MUST satisfy all of the following:\\
- Relevant: It must be directly tied to the core information need of the query and help produce the answer, so that it meaningfully reduces uncertainty when judging document quality.\\
- Binary: It must be a Yes/No question. No degree-based phrasing (e.g., ``to what extent'').\\
- Qualitative \& query-specific: It must be purpose-built for this specific (query, answer). You MUST name the concrete entities, facts, dates, or numbers from the query/answer. Never use vague wording such as ``relevant content'' or ``important information''.\\
- Grounded: Its judgment criterion must be derivable from the given (query, answer); a judge should not need external knowledge to answer it.\\
- Salient: It should focus on the aspects an experienced information-retrieval expert would care about when evaluating documents.\\
- Per-document: Each rubric MUST be phrased as a judgment about EACH individual document in the set (e.g., ``Does each document ...'').\\
{\# Dimensions}\\
Generate rubrics for the following three orthogonal dimensions:\\
- {Relevance}: Measures how semantically related each document's content is to the core information need of the query --- whether the document directly addresses, at the topic level, what the query is asking, rather than merely matching surface keywords.\\
\hspace{1em}Requirement: The rubric must name the concrete subject and information facet the query asks about (e.g., specific people / works / events / time dimension) and judge whether the document truly discusses this specific question, not merely whether it is ``relevant''.\\
\hspace{1em}Goal: Keep documents that directly answer the user's question; discard off-topic or keyword-only, semantically irrelevant documents.\\
- \textbf{Authenticity}: Measures how well the facts stated in each document conform to objective real-world information.\\
\hspace{1em}Requirement: The rubric must spell out the correct reference fact (taken from the answer or the deterministic information in the query) and judge whether the document's statements are consistent with it, rather than asking whether the document ``contains correct information''. Only generate this dimension when the query/answer involves concrete, verifiable facts (e.g., dates, numbers, names, causal conclusions); otherwise return an empty list for it.\\
\hspace{1em}Goal: Discard documents with false statements or factual errors.\\
- {Quality}: Measures how usable each document is in terms of information organization and presentation --- whether it is clearly structured, focused, and lets the reader directly extract the target information, rather than being poorly formatted, too short / shallow, noisy, or low-quality filler.\\
\hspace{1em}Requirement: The rubric must reference the specific information a reader needs to extract in order to answer this query (name the concrete entities, not generic terms), and judge whether the document makes that information clearly accessible. This dimension evaluates presentation quality only, not factual correctness.\\
\hspace{1em}Goal: Discard low-quality documents (empty content, low information density, unclear phrasing, etc.).\\
{\# Additional criteria}\\
- Knowledge-cutoff: Do not ask for or rely on information that would violate a knowledge cutoff date.\\
- Coverage \& non-redundancy: The rubrics together should cover the important aspects of the query without redundant or duplicate questions.\\
- Generate at most 3 questions per dimension (fewer is fine). Every question must be important, useful, and non-redundant.\\
{\# Example}\\
Query: Which was founded earlier, Arthur's Magazine or First for Women?\\
Answer: Arthur's Magazine\\
\{\\
\hspace{1em}``Relevance'': [\\
\hspace{2em}``Does each document directly discuss the founding/launch date or early publication history of Arthur's Magazine or First for Women, rather than merely containing words like `Arthur', `magazine', or `women' without addressing founding dates?''\\
\hspace{1em}],\\
\hspace{1em}``Authenticity'': [\\
\hspace{2em}``For each document, are its founding-date statements consistent with objective fact - namely Arthur's Magazine = 1844 and First for Women = 1989 - without misstating the years or attributing them to the wrong magazine?'',\\
\hspace{2em}``For each document that states which was founded first, is that conclusion consistent with the fact that Arthur's Magazine predates First for Women, rather than claiming First for Women is older?''\\
\hspace{1em}],\\
\hspace{1em}``Quality'': [\\
\hspace{2em}``Is each document clearly structured and focused enough that a reader can directly extract the founding year of Arthur's Magazine and/or First for Women (or the conclusion that Arthur's Magazine came first), rather than having these key dates buried under large amounts of irrelevant content?''\\
\hspace{1em}]\\
\}\\
{\# Output format}\\
Output ONLY a valid JSON object (no markdown code fences, no extra commentary) with exactly these three keys: ``Relevance'', ``Authenticity'', ``Quality''. Each key maps to a list of rubric question strings (a dimension may have multiple rubrics). If a dimension does not apply (e.g., Authenticity for a non-verifiable query), use an empty list [].\\
---\\
Query: \prompttag{\textbf{\color{insert}<QUESTION>}}\\
Answer: \prompttag{\textbf{\color{insert}<ANSWER>}}\\
Rubrics (JSON):
}}%


\prompt
{Rubric Generation Prompt for Short-form Scenario: Set-Level}{
\systemprompt{
You are a senior rubric-generation expert. Given a (query, answer) pair, you generate binary (Yes/No) evaluation rubrics along several set-level dimensions. These rubrics evaluate a candidate document set AS A WHOLE --- judging whether the set, taken together, can effectively support answering the query and producing the correct answer.\\
\#\ Rubric requirements\\
Every rubric you generate MUST satisfy all of the following:\\
- Relevant: It must be directly tied to the core information need of the query and help produce the answer, so that it meaningfully reduces uncertainty when judging the document set's quality.\\
- Binary: It must be a Yes/No question. No degree-based phrasing (e.g., ``to what extent'').\\
- Qualitative \& query-specific: It must be purpose-built for this specific (query, answer). You MUST name the concrete entities, facts, dates, or numbers from the query/answer. Never use vague wording such as ``relevant content'' or ``important information''.\\
- Grounded: Its judgment criterion must be derivable from the given (query, answer); a judge should not need external knowledge to answer it.\\
- Salient: It should focus on the aspects an experienced information-retrieval expert would care about when evaluating a document set.\\
- Set-level: Each rubric MUST be phrased as a judgment about the document set AS A WHOLE (e.g., ``Does the document set ...'', ``Across the documents, ...''), NOT about each individual document. These dimensions only make sense when the set is considered collectively.\\
\#\ Dimensions\\
Generate rubrics for the following three set-level dimensions:\\
- Complementarity: Measures the degree to which the documents in the set complement one another --- whether, taken together, they cover all the key information elements the answer requires and jointly build a more complete, well-rounded answer than any single document could (e.g., one document gives an overview while another gives the details; one covers 70\% of the must-have points while another covers the remaining 30\%).\\
\ \ Requirement: The rubric must name the concrete key information elements the query/answer requires, and judge whether the set as a whole covers them across its documents. Distinguish complementarity (documents combine to increase information value) from conflict (documents contradict each other).\\
\ \ Goal: Reward sets whose documents jointly raise coverage of the query's information need when no single document fully satisfies it.\\
- Redundancy: Measures the degree of content redundancy among the documents in the set --- whether two or more documents convey essentially the same core information, data, or conclusions without any incremental contribution (common when documents merely reprint, summarize, rewrite, or re-report the same source).\\
\ \ Requirement: The rubric must name the concrete query-specific information, and judge whether the set avoids having two or more documents that carry essentially identical information with no incremental value (in which case only the single best --- highest information density, most concise, clearest --- should be kept).\\
\ \ Goal: Discourage sets that pad in duplicated documents instead of adding new information.\\
- Conflict: Measures the degree of factual consistency across the documents in the set --- whether different documents give mutually contradictory factual statements about the same point (e.g., two different values for the same fact, one correct and one wrong).\\
\ \ Requirement: The rubric must name the concrete fact at issue and judge whether the documents' statements are free of mutual contradiction, so that the set consistently supports the correct answer. A high-quality set must never contain factual contradictions. Only generate this dimension when the query/answer involves concrete, verifiable facts that documents could contradict; otherwise return an empty list for it.\\
\ \ Goal: Reject sets containing factual contradictions or documents with erroneous information.\\
\#\ Additional criteria\\
- Knowledge-cutoff: Do not ask for or rely on information that would violate a knowledge cutoff date.\\
- Coverage \& non-redundancy: The rubrics together should cover the important aspects of the query without redundant or duplicate questions.\\
- Generate at most 3 questions per dimension (fewer is fine). Every question must be important, useful, and non-redundant.\\
\#\ Example\\
Query: Which was founded earlier, Arthur's Magazine or First for Women?\\
Answer: Arthur's Magazine\\
\{\\
\ \ ``Complementarity'': [\\
\ \ \ \ ``Does the document set contain at least one document that explicitly records the founding year of Arthur's Magazine AND at least one that explicitly records the founding year of First for Women, so that the two can be compared to determine which was founded first?''\\
\ \ ],\\
\ \ ``Redundancy'': [\\
\ \ \ \ ``Does the document set avoid having two or more documents that convey essentially the same founding-date information (e.g., all merely stating `Arthur's Magazine was founded in 1844') without any incremental information contribution?''\\
\ \ ],\\
\ \ ``Conflict'': [\\
\ \ \ \ ``Are the statements across the document set about the two magazines' founding dates free of mutual contradiction (e.g., one saying 1844 and another saying 1850), so that they consistently support the conclusion that Arthur's Magazine was founded earlier?''\\
\ \ ]\\
\}\\
\#\ Output format\\
Output ONLY a valid JSON object (no markdown code fences, no extra commentary) with exactly these three keys: ``Complementarity'', ``Redundancy'', ``Conflict''. Each key maps to a list of rubric question strings (a dimension may have multiple rubrics). If a dimension does not apply (e.g., Conflict for a query with no verifiable facts to contradict), use an empty list [].\\
---\\
Query: \prompttag{\textbf{\color{insert}<QUESTION>}}\\
Answer: \prompttag{\textbf{\color{insert}<ANSWER>}}\\
Rubrics (JSON):
}}%


\prompt
{Rubric Generation Prompt for Short-form Scenario: Global-Level}{
\systemprompt{
You are a senior rubric-generation expert. Given a (query, answer) pair, you generate binary (Yes/No) evaluation rubrics along several set-level dimensions. These rubrics evaluate a candidate document set AS A WHOLE --- judging whether the set, taken together, can effectively support answering the query and producing the correct answer.\\
\#\ Rubric requirements\\
Every rubric you generate MUST satisfy all of the following:\\
- Relevant: It must be directly tied to the core information need of the query and help produce the answer, so that it meaningfully reduces uncertainty when judging the document set's quality.\\
- Binary: It must be a Yes/No question. No degree-based phrasing (e.g., ``to what extent'').\\
- Qualitative \& query-specific: It must be purpose-built for this specific (query, answer). You MUST name the concrete entities, facts, dates, or numbers from the query/answer. Never use vague wording such as ``relevant content'' or ``important information''.\\
- Grounded: Its judgment criterion must be derivable from the given (query, answer); a judge should not need external knowledge to answer it.\\
- Salient: It should focus on the aspects an experienced information-retrieval expert would care about when evaluating a document set.\\
- Set-level: Each rubric MUST be phrased as a judgment about the document set AS A WHOLE (e.g., ``Does the document set ...'', ``Across the documents, ...''), NOT about each individual document. These dimensions only make sense when the set is considered collectively.\\
\#\ Dimensions\\
Generate rubrics for the following three set-level, whole-set-utility dimensions:\\
- Completeness: Measures whether the document set, as a whole, can COMPLETELY answer the query with no obvious information gaps --- every key information element the correct answer depends on is present somewhere in the set.\\
\ \ Requirement: The rubric must enumerate the concrete key information elements the query/answer requires and judge whether the whole set collectively supplies all of them, leaving no obvious gap that would prevent producing the correct answer.\\
\ \ Goal: Reward sets that fully cover the query's information need; reject sets that miss essential pieces.\\
- Density: Measures, in terms of LENGTH / span, how much of each document in the set is actually useful for answering this query --- i.e., the ratio of useful content to the document's total length. It is NOT about whether a document is removable, and NOT about whether the set as a whole is complete. It focuses on the internal ``signal-to-length'' ratio of the documents.\\
\ \ - What we WANT (high Density): each document in the set is compact and on-point, so that the text carrying the query-relevant information (name the concrete query-specific facts/entities/dates) occupies the MAJORITY of that document's length.\\
\ \ - What we DO NOT want (low Density): a document that is long but mostly filler --- the bulk of its length is off-topic / unrelated content, and only a small fraction of the text actually contributes useful information for this query.\\
\ \ Requirement: The rubric must name the concrete query-specific useful information, and judge whether --- across the documents in the set --- the useful content takes up the main portion of each document's length, rather than the documents being padded with large amounts of irrelevant text where only a small snippet is useful.\\
\ \ Goal: Reward sets whose documents are dense with useful content; penalize sets containing bloated documents that are mostly irrelevant padding with only a sliver of useful information.\\
- Reachability: Measures whether a model WITHOUT any external knowledge could complete the full reasoning chain required to produce the correct answer using ONLY the document set. Every link of the inference chain must be grounded in the set.\\
\ \ Requirement: The rubric must name the concrete reasoning steps / intermediate facts the query requires (e.g., the bridge entities and the final comparison), and judge whether the whole set self-containedly provides every link so the correct answer is derivable without outside knowledge.\\
\ \ Goal: Reward sets that are self-contained for the full reasoning chain; reject sets with a missing link that forces reliance on external knowledge.\\
\#\ Additional criteria\\
- Knowledge-cutoff: Do not ask for or rely on information that would violate a knowledge cutoff date.\\
- Coverage \& non-redundancy: The rubrics together should cover the important aspects of the query without redundant or duplicate questions.\\
- Generate at most 3 questions per dimension (fewer is fine). Every question must be important, useful, and non-redundant.\\
\#\ Example\\
Query: Which was founded earlier, Arthur's Magazine or First for Women?\\
Answer: Arthur's Magazine\\
\{\\
\ \ ``Completeness'': [\\
\ \ \ \ ``Does the document set as a whole provide the founding year of BOTH Arthur's Magazine and First for Women, leaving no gap that would prevent determining which was founded earlier?''\\
\ \ ],\\
\ \ ``Density'': [\\
\ \ \ \ ``Across the documents in the set, does the text that actually states the founding date(s) of Arthur's Magazine and First for Women make up the majority of each document's length, rather than each document being dominated by unrelated content (e.g., editorial history, staff, or coverage of other magazines) with only a small fraction of its length mentioning those founding dates?''\\
\ \ ],\\
\ \ ``Reachability'': [\\
\ \ \ \ ``Using only the document set and no external knowledge, can the full reasoning chain be completed --- obtaining the founding year of Arthur's Magazine, obtaining the founding year of First for Women, and comparing them --- to reach the conclusion that Arthur's Magazine was founded earlier?''\\
\ \ ]\\
\}\\
\#\ Output format\\
Output ONLY a valid JSON object (no markdown code fences, no extra commentary) with exactly these three keys: ``Completeness'', ``Density'', ``Reachability''. Each key maps to a list of rubric question strings (a dimension may have multiple rubrics). If a dimension does not apply, use an empty list [].\\
---\\
Query: \prompttag{\textbf{\color{insert}<QUESTION>}}\\
Answer: \prompttag{\textbf{\color{insert}<ANSWER>}}\\
Rubrics (JSON):
}}%


\prompt
{Hybrid Rubric Aggregation Prompt for Short-form Scenario}{
\systemprompt{
You are a senior rubric-review expert (a judge). Given a (query, answer) pair and two sets of rubrics generated by two different models for THE SAME evaluation dimension, you must merge them into a single set of high-quality hybrid rubrics.\\
\#\ Task background\\
These rubrics are used for downstream evaluation: judging whether a candidate document set (doc set) can effectively support answering the query and producing the correct answer. Each rubric is a binary (Yes/No) judgment question.\\
\#\ Current evaluation dimension\\
- Dimension name: \prompttag{\textbf{\color{insert}<DIM>}}\\
- Dimension meaning: \prompttag{\textbf{\color{insert}<DIM\_DESC>}}\\
- Evaluation granularity: \prompttag{\textbf{\color{insert}<GRAN>}}\\
\#\ Your goal and requirements\\
Consider the two sets of rubrics together and merge/select the hybrid rubrics best suited for downstream doc set evaluation. Specifically:\\
- Maximize richness: cover all important and discriminative judgment points under this dimension; every valuable judgment point unique to either set should be kept, not omitted.\\
- Minimize redundancy: merge semantically duplicate or highly similar rubrics; do not keep redundant items; each rubric should focus on a single atomic judgment point.\\
- Maximize clarity: wording must be precise and query-specific (it must name the concrete entities, facts, numbers, and dates from the query/answer), must be a binary (Yes/No) decidable question answerable without external knowledge; remove vague or ambiguous phrasing.\\
- Keep the evaluation granularity required by this dimension (see ``Evaluation granularity'' above).\\
- You may rewrite, split, or merge the original phrasings to achieve the above.\\
- If both sets are empty, or this dimension truly does not apply to this query, you may return an empty list [].\\
\#\ Output format\\
Output ONLY a JSON object (no markdown code fences, no extra commentary or prefix/suffix) with the following three fields:\\
- ``rubrics'': a JSON array where each element is a hybrid rubric string (use an empty array [] if the dimension does not apply or both sets are empty).\\
- ``selected'': a JSON array of the SAME length as ``rubrics'' and in the SAME order. Each element is itself an array of source tags for the corresponding rubric: [``a''] if it comes only from model\_a, [``b''] if it comes only from model\_b, or [``a'', ``b''] if it is merged from both.\\
- ``reason'': a short explanation of the merge/selection (what was kept, what was merged or dropped, and why); keep it concise.\\
For example (3 rubrics: the 1st only from model\_a, the 2nd merged from both, the 3rd only from model\_b):\\
\{``rubrics'': [``rubric question 1'', ``rubric question 2'', ``rubric question 3''], ``selected'': [[``a''], [``a'', ``b''], [``b'']], ``reason'': ``brief explanation of the selection''\}\\
\#\ Example (dimension: \prompttag{\textbf{\color{insert}<DIM>}}, \prompttag{\textbf{\color{insert}<GRAN>}})\\
\prompttag{\textbf{\color{insert}<EXAMPLE\_BLOCK>}}\\
---\\
\#\ Input (produce the hybrid rubric based on this)\\
Query: \prompttag{\textbf{\color{insert}<QUESTION>}}\\
Answer: \prompttag{\textbf{\color{insert}<ANSWER>}}\\
Rubrics generated by model\_a for this dimension (a list, possibly empty or with multiple items):\\
\prompttag{\textbf{\color{insert}<A\_JSON>}}\\
Rubrics generated by model\_b for this dimension (a list, possibly empty or with multiple items):\\
\prompttag{\textbf{\color{insert}<B\_JSON>}}\\
Hybrid (JSON object with ``rubrics'' and ``reason''):
}}%


\prompt
{Rubric Scoring Prompt for Short-form Scenario}{
\systemprompt{
You are a senior search-result quality assessor. Given a candidate document set (doc set, variable size) and a list of rubrics (each labeled with its dimension type and judgment content), you must score every rubric on a 0--4 scale, strictly grounded in the document content.\\
This round evaluates all 9 dimensions at once, grouped into three categories:\\{}
[A. Doc-Level dimensions] --- judged for EACH individual document in the doc set:\\
- Relevance, Authenticity, Quality.\\{}
[B. Set-Level dimensions] --- judged for the ENTIRE doc set as a whole; give ONE overall score, not per-document:\\
- Complementarity, Redundancy, Conflict.\\{}
[C. Global-Level dimensions] --- also judged for the ENTIRE doc set as a whole; give ONE overall score, not per-document:\\
- Completeness, Density, Reachability.\\
\#\ Absolute prerequisite: you MUST read the documents first\\
- Every score must be grounded in the document text, and you must give evidence: quote the key original snippet that supports the judgment (for doc-level, quote that document; for set-level, you may quote one or more relevant documents and mark their doc\_id). Do NOT assign a score without quoting the source.\\
- Judge only based on what ACTUALLY appears in the documents; do not use your own external knowledge to fill in information the documents do not state.\\
- The rubric text is itself query-specific (it already contains the concrete entities, facts, and numbers of the question); judge directly against it --- no extra query/answer is needed.\\
\#\ Evaluation procedure --- RELEVANCE GATE (read carefully, follow the order)\\
Think step by step in this order:\\
1. FIRST evaluate the Relevance dimension for EVERY document in the doc set (give each document its per\_doc Relevance score).\\
2. Then check the relevance outcome:\\
\ \ - CASE A --- ALL documents are completely irrelevant (every document's Relevance score = 0): the documents this ranker selected are all useless for this question, so the other 8 dimensions (Authenticity, Quality, Complementarity, Redundancy, Conflict, Completeness, Density, Reachability) carry NO evaluation meaning. In this case DO NOT score them --- output each of those 8 dimensions as null (see output format). Only the Relevance dimension is scored.\\
\ \ - CASE B --- at least ONE document is useful (its Relevance score >= 1): you MUST evaluate ALL 9 dimensions normally, even if only a single document is relevant.\\
3. Never leave Relevance itself null --- Relevance is always scored.\\
\#\ Scoring scale (0--4)\\
Overall meaning (higher = better satisfies the rubric):\\
- 4 = Completely: fully satisfied, with sufficient detail and evidence.\\
- 3 = Mostly: largely satisfied, but missing some detail or support.\\
- 2 = Moderately: relevant content is mentioned, but key details are missing.\\
- 1 = Barely: not explicitly stated, only weakly inferable.\\
- 0 = Not at all: not addressed at all / irrelevant to the rubric / contradicts the facts.\\
\#\#\ I. Doc-Level dimensions: score EACH document individually\\
- Relevance: to what extent the document DIRECTLY discusses the specific information facet the rubric points to.\\
\ \ - 4: the document's core content directly and fully discusses this specific question and can directly support the answer.\\
\ \ - 3: directly discusses the topic, but the information is incomplete or unfocused.\\
\ \ - 2: mentions the related topic only in passing, missing key information.\\
\ \ - 1: barely discusses it directly, relevance is only weakly inferable.\\
\ \ - 0: entirely off-topic, merely matching keywords or unrelated to the question.\\
- Authenticity: how well the facts stated in the document CONFORM to the reference fact given by the rubric. Only assess whether the STATEMENTS that appear are accurate; independent of relevance.\\
\ \ - 4: the document explicitly states the relevant fact, fully consistent with the reference fact.\\
\ \ - 3: the statement is consistent with the fact, but phrased unclearly or somewhat vaguely.\\
\ \ - 2: partially correct, or the fact is right but wrapped in easily-misleading phrasing.\\
\ \ - 1: only faintly touched upon; correctness is hard to judge or slightly off.\\
\ \ - 0: the statement contradicts the reference fact, has an obvious factual error, or the document makes no statement related to that fact at all.\\
- Quality: how easy the document makes it for a reader to EXTRACT the target information, in terms of information organization and presentation. Assess presentation quality only, not factual correctness.\\
\ \ - 4: clearly structured and focused; the target information is immediately apparent.\\
\ \ - 3: mostly clear; the target information is extractable but mixed with some irrelevant content.\\
\ \ - 2: the target information exists but is diluted by lots of irrelevant content and takes effort to find.\\
\ \ - 1: messy layout, high noise; the target information is very hard to extract.\\
\ \ - 0: empty, extremely noisy, or nearly unparseable content, OR the document is entirely off-topic and contains no target information at all.\\
\#\#\ II. Set-Level dimensions: give ONE overall score for the WHOLE doc set\\
- Complementarity: whether the key information points are covered through DIVISION OF LABOUR across DIFFERENT documents --- i.e., different documents each contribute a DIFFERENT needed piece (one gives an overview, another the details; one supplies point P1, another supplies point P2), so that the whole is more valuable than any single document. IMPORTANT: this dimension judges only the complementary STRUCTURE among the documents that are present; it does NOT judge whether the set is complete. Do NOT lower the score merely because some required point is missing (that is Completeness's job) --- an incomplete set can still be highly complementary if the points it DOES cover come from different documents. Conversely, if a SINGLE document already covers everything on its own, complementarity is LOW even though coverage may be perfect (there is no division of labour).\\
\ \ - 4: strong complementarity --- multiple documents clearly divide the labour, each contributing a DIFFERENT needed point, so the whole is far more valuable than any single document.\\
\ \ - 3: mostly complementary --- most of the covered points are supplied by different documents; only an occasional point is carried alone by one document.\\
\ \ - 2: partial complementarity --- there is some division of labour, but a large share of the covered content is concentrated in a single document.\\
\ \ - 1: very weak complementarity --- almost all useful content comes from one document, or several documents merely restate the same single point.\\
\ \ - 0: no complementarity at all --- only one document contributes any useful content (the others add nothing), OR a single document already covers everything by itself so no cross-document complementarity exists.\\
- Redundancy: judged against the MINIMAL document subset needed to cover the key information points the answer requires (the SAME key information points as Completeness). It checks whether each document brings INCREMENTAL coverage of those required information points, or merely repeats a point already covered by another document. If two or more documents cover the SAME required information point while adding no new required point, that overlap is redundancy --- those documents could be dropped from the minimal covering set without losing any required point. (Example: if answering the query needs 2 information points, and both doc [1] and doc [5] only answer the SAME single point, that is redundancy.) Higher score = less redundancy (each document adds a new required point; close to a minimal covering set); lower = many documents duplicating the same point with no increment. (Judge only within the doc set; do not factor in relevance here.)\\
\ \ - 4: no redundancy --- every document contributes a DISTINCT required information point; the set is essentially a minimal covering set with no duplicated point.\\
\ \ - 3: nearly no redundancy --- at most a negligible overlap; almost every document still adds a new required point.\\
\ \ - 2: some redundancy --- a few documents repeat an already-covered information point and add no new required point.\\
\ \ - 1: heavy redundancy --- many documents cover the same information point(s), giving little incremental coverage of the required points.\\
\ \ - 0: severe redundancy --- most documents duplicate the same one/few information points and bring no new required point (e.g., several documents all only answer the same single point).\\
- Conflict: whether different documents in the set are FREE OF mutual contradiction on the same factual point.\\
\ \ APPLICABILITY PREREQUISITE (judge this FIRST): Conflict can only be judged when the doc set contains AT LEAST TWO comparable statements about the SAME query-relevant factual point (i.e., two or more documents each make a checkable claim about the same fact, so that agreement or contradiction is even possible). If FEWER THAN TWO comparable statements exist about the same fact (the relevant fact is asserted by at most one document, or is not asserted at all), then there is nothing that could agree or contradict --- Conflict is NOT APPLICABLE. In that case DO NOT give a 0--4 score; output this dimension as null (see output format). Do NOT give a high score merely because ``nothing contradicts'': absence of comparable statements is null, not 4.\\
\ \ When the prerequisite is met (>=2 comparable statements about the same fact), score 0--4 by how consistent they are. Higher score = less conflict, more consistently supporting the correct answer. (Only judge whether contradictions exist within the doc set; do not factor in relevance here.)\\
\ \ - 4: the two-or-more statements about the same fact are fully consistent, with no contradiction.\\
\ \ - 3: overall consistent; only a minor non-key wording difference that can be ignored.\\
\ \ - 2: slight inconsistency exists, but it does not affect reaching the correct conclusion.\\
\ \ - 1: clear disagreement on a key fact, potentially misleading the conclusion.\\
\ \ - 0: a direct factual contradiction exists (e.g., two mutually exclusive values for the same fact), undermining the set's support for the correct answer.\\
\#\#\ III. Global-Level dimensions: give ONE overall score for the WHOLE doc set\\
- Completeness: whether the doc set as a whole covers ALL key information points the correct answer depends on, with no obvious information gap.\\
\ \ - 4: fully covers all key information points the rubric lists, with no gap, sufficient to derive the correct answer.\\
\ \ - 3: covers the vast majority of key points, missing only a minor piece; largely answerable.\\
\ \ - 2: covers some key points but misses important information; hard to answer completely.\\
\ \ - 1: only sporadically covers a few points; key information largely missing.\\
\ \ - 0: covers almost no key point; impossible to answer from it.\\
- Density: judged in terms of LENGTH / span --- whether the content that is actually useful for answering the query (the text that touches the required information points) occupies the MAJORITY of each document's length, rather than the documents being padded with large amounts of irrelevant / filler text where only a small fraction of the length is useful. Higher score = the documents are compact and on-point (high useful-content-to-length ratio); lower = the documents are long but mostly filler, with only a small useful snippet. This concerns the internal signal-to-length ratio of the documents, NOT whether a document is removable.\\
\ \ - 4: across the documents, useful content occupies the vast majority of each document's length; the documents are compact and on-point with almost no filler.\\
\ \ - 3: useful content occupies most of the length in most documents; only minor filler / padding.\\
\ \ - 2: useful content and irrelevant filler are roughly balanced; a fair share of the length is not useful.\\
\ \ - 1: in most documents only a small fraction of the length is useful; the bulk is filler or off-topic padding.\\
\ \ - 0: the documents are almost entirely filler; only a tiny sliver of the length is useful for this query.\\
- Reachability: whether, using ONLY this doc set and NO external knowledge, the COMPLETE reasoning chain required to reach the correct answer can be completed (every link grounded in the doc set).\\
\ \ - 4: every link of the reasoning chain (bridge entities, intermediate facts, final comparison, etc.) is grounded in the set; the correct answer is self-containedly derivable.\\
\ \ - 3: the vast majority of links are supported by the set; only a minor link needs slight guessing.\\
\ \ - 2: a key link is partially missing; some external knowledge is needed to complete the chain.\\
\ \ - 1: the reasoning chain breaks in multiple places; many links lack document grounding.\\
\ \ - 0: a key link is missing; it is almost impossible to derive the correct answer from the doc set alone.\\
\#\ Scoring example (Case)\\
Suppose the rubrics are:\\
\ \ [Relevance] (Doc-Level) ``Does each document directly discuss the 1999 regular-season win-loss record of the team that won Super Bowl XXXIV (the St.\ Louis Rams)?''\\
\ \ [Authenticity] (Doc-Level) ``Is each document's statement about that team's 1999 regular-season record consistent with the objective fact (13--3)?''\\
\ \ [Quality] (Doc-Level) ``Is each document clearly structured so a reader can directly extract the 13--3 record?''\\
\ \ [Complementarity] (Set-Level) ``Does the doc set, through complementary documents, jointly cover both key facets --- `the identity of the champion team' and `the 1999 regular-season 13--3 record'?''\\
\ \ [Redundancy] (Set-Level) ``Does the doc set avoid multiple documents redundantly covering the same required information point (the champion identity or the 13--3 record) without adding a new required point?''\\
\ \ [Conflict] (Set-Level) ``Are the statements across the doc set about the Rams' 1999 regular-season record consistent, with no mutual contradiction?''\\
\ \ [Completeness] (Global-Level) ``Does the doc set fully provide the champion team and its 1999 regular-season 13--3 record, sufficient to answer the question?''\\
\ \ [Density] (Global-Level) ``Across the documents, does the text useful for the champion team and its 13--3 record occupy the majority of each document's length, rather than being buried in filler?''\\
\ \ [Reachability] (Global-Level) ``Using only this doc set, can the full reasoning chain `confirm the champion is the Rams -> find their 1999 record 13--3' be completed?''\\
Suppose the answer requires 2 key information points: (P1) the champion is the St.\ Louis Rams; (P2) their 1999 regular-season record is 13--3.\\
Doc-Level dimensions (score each document):\\
- Doc A (text: ``...the Rams finished 1999 at 13--3 and won Super Bowl XXXIV...'')\\
\ \ - Relevance = 4: directly and explicitly gives the champion team's 1999 record.\\
\ \ - Authenticity = 4: 13--3 fully matches the fact.\\
\ \ - Quality = 4: the record is clear and directly extractable.\\
- Doc B (text: about the Titans ``finishing 1999 at 13--3, losing to the Rams'')\\
\ \ - Relevance = 2: discusses the same game's record, but the subject is the losing Titans rather than the champion Rams, so only indirectly relevant.\\
\ \ - Authenticity = 2: the Titans indeed went 13--3, but using it to answer ``the champion's record'' is misleading (both teams coincidentally went 13--3).\\
\ \ - Quality = 3: the record is clearly stated, but the reader must distinguish which team.\\
- Doc C (text: about a different Super Bowl XLI, some team 12--4)\\
\ \ - Relevance = 0: an entirely different edition, unrelated to this question.\\
\ \ - Authenticity = 0: makes no statement about the Rams' 1999 record.\\
\ \ - Quality = 0: unrelated to the query, contains no target information.\\
Set-Level dimensions (one overall score):\\
- Complementarity = 1: Doc A ALONE already covers both required points (P1 champion = Rams, P2 record = 13--3), so there is essentially no division of labour; Doc B only adds non-required Titans context (not a new required point) and Doc C is unrelated --- no genuine cross-document complementarity, even though coverage (Completeness) is perfect.\\
- Redundancy = 2: Doc A already covers both required points (P1 and P2); Doc B mainly repeats the 13--3 point (P2) from the Titans' angle without adding a NEW required point, so relative to the minimal covering set (Doc A alone) there is some redundancy; Doc C is unrelated and covers no required point.\\
- Conflict = null (NOT APPLICABLE): the rubric points to the RAMS' 1999 record; only Doc A actually states the Rams' record (13--3). Doc B states the TITANS' record, not the Rams', and Doc C is unrelated --- so there are FEWER THAN TWO comparable statements about the same fact (the Rams' record). Since nothing could agree or contradict, Conflict is not applicable and is output as null (NOT 4).\\
Global-Level dimensions (one overall score):\\
- Completeness = 4: Doc A already gives ``the Rams won + 1999 record 13--3'' (both P1 and P2); the whole set has no information gap.\\
- Density = 2: Doc A is compact and on-point (useful text dominates its length), but Doc B carries extra non-essential Titans narrative and Doc C is almost entirely off-topic filler, so overall only part of the set's total length is genuinely useful for this question.\\
- Reachability = 4: Doc A alone completes the full reasoning chain ``champion = Rams -> record = 13--3'', with no external knowledge needed.\\
\#\ Output format\\
Output ONLY a single valid JSON object (no markdown code fences, no extra commentary). Note: doc-level dimensions use per\_doc (per document); set-level dimensions use set\_level (one overall score). You ONLY need to give each rubric's atomic scores (the score inside per\_doc / set\_level) plus evidence/reason; you do NOT need to compute rubric\_avg or dimension\_score (those are computed by downstream code). Structure:\\
\{\\
\ \ ``score'': \{\\
\ \ \ \ ``Relevance'': \{\\
\ \ \ \ \ \ ``level'': ``doc'',\\
\ \ \ \ \ \ ``rubrics'': [\\
\ \ \ \ \ \ \ \ \{\\
\ \ \ \ \ \ \ \ \ \ ``rubric'': ``<the original rubric text>'',\\
\ \ \ \ \ \ \ \ \ \ ``per\_doc'': [\\
\ \ \ \ \ \ \ \ \ \ \ \ \{``doc\_id'': 9, ``evidence'': ``quoted key snippet from the document labeled [9]'', ``score'': 4, ``reason'': ``concise scoring reason''\},\\
\ \ \ \ \ \ \ \ \ \ \ \ \{``doc\_id'': 18, ``evidence'': ``quoted key snippet from the document labeled [18]'', ``score'': 2, ``reason'': ``concise scoring reason''\}\\
\ \ \ \ \ \ \ \ \ \ ]\\
\ \ \ \ \ \ \ \ \}\\
\ \ \ \ \ \ ]\\
\ \ \ \ \},\\
\ \ \ \ ``Authenticity'': \{ ``level'': ``doc'', ``rubrics'': [ ... ] \},\\
\ \ \ \ ``Quality'': \{ ``level'': ``doc'', ``rubrics'': [ ... ] \},\\
\ \ \ \ ``Complementarity'': \{\\
\ \ \ \ \ \ ``level'': ``set'',\\
\ \ \ \ \ \ ``rubrics'': [\\
\ \ \ \ \ \ \ \ \{\\
\ \ \ \ \ \ \ \ \ \ ``rubric'': ``<the original rubric text>'',\\
\ \ \ \ \ \ \ \ \ \ ``set\_level'': \{``evidence'': ``quoted key snippets from relevant documents (may mark multiple doc\_id)'', ``score'': 3, ``reason'': ``concise reason for the whole set''\}\\
\ \ \ \ \ \ \ \ \}\\
\ \ \ \ \ \ ]\\
\ \ \ \ \},\\
\ \ \ \ ``Redundancy'': \{ ``level'': ``set'', ``rubrics'': [ ... ] \},\\
\ \ \ \ ``Conflict'': \{ ``level'': ``set'', ``rubrics'': [ ... ] \},\\
\ \ \ \ ``Completeness'': \{ ``level'': ``set'', ``rubrics'': [ ... ] \},\\
\ \ \ \ ``Density'': \{ ``level'': ``set'', ``rubrics'': [ ... ] \},\\
\ \ \ \ ``Reachability'': \{ ``level'': ``set'', ``rubrics'': [ ... ] \}\\
\ \ \}\\
\}\\
If CASE A applies (ALL documents have Relevance = 0), output the 8 non-Relevance dimensions as null, like:\\
\ \ ``Authenticity'': null, ``Quality'': null, ``Complementarity'': null, ``Redundancy'': null,\\
\ \ ``Conflict'': null, ``Completeness'': null, ``Density'': null, ``Reachability'': null\\
(Only ``Relevance'' is fully scored in CASE A.)\\
Conflict-not-applicable (independent of CASE A): even in CASE B, if the doc set has fewer than two comparable statements about the same query-relevant fact, output ``Conflict'': null (a bare JSON null for the whole dimension), and still score the other dimensions normally.\\
Rules:\\
- Only output dimensions actually provided in the input; under each dimension, the number and order of rubrics must match the input.\\
- Doc-level dimensions: per\_doc must cover EVERY document in the doc set. CRITICAL: doc\_id MUST be exactly the id shown in square brackets before each document in the ``Doc set'' below (that is the document's original id, e.g.\ [9] -> doc\_id 9). Do NOT renumber the documents as 1,2,3,...; use their bracketed original ids. For set-level dimensions, when you cite evidence you should likewise refer to documents by their bracketed original ids.\\
- set-level dimensions: give only one overall set\_level score, not per-document.\\
- RELEVANCE GATE: always score Relevance. If ALL documents have Relevance = 0 (CASE A), output the other 8 dimensions as null; otherwise (CASE B) score all 9 dimensions.\\
- Redundancy judges whether documents redundantly cover the SAME required information point without adding a new required point (i.e., overlap relative to a minimal covering set); Conflict judges whether contradictions exist inside the doc set. Density judges the useful-content-to-length ratio of the documents. Do not lower these merely because documents are off-topic (that is Relevance's job).\\
- Conflict is CONDITIONAL: only score it (0--4) when at least two documents make comparable statements about the SAME query-relevant fact; if there are fewer than two such comparable statements, output ``Conflict'': null (do NOT output 4). Never treat ``nothing to contradict'' as a high Conflict score.\\
- Complementarity vs Completeness are DIFFERENT: Complementarity judges cross-document DIVISION OF LABOUR (if a single document covers everything by itself, Complementarity is LOW even when nothing is missing), and must NOT be lowered just because some point is missing; Completeness judges only whether any required information point is MISSING (an information gap), regardless of how many documents supply it. Do not conflate the two.\\
- score can only be 0/1/2/3/4 (or null for a whole dimension under CASE A); no N/A or empty per-item values allowed.\\
- Keep evidence short (quote the key snippet only), avoid verbosity.\\
- Keep reason concise (one sentence), explaining why this score was given.\\
---\\
Doc set:\\
\prompttag{\textbf{\color{insert}<DOCS\_STR>}}\\
Rubrics (each labeled with its type and content):\\
\prompttag{\textbf{\color{insert}<RUBRICS\_STR>}}\\
Now begin outputting the scores (JSON):
}}%


\prompt
{Rubric Generation Prompt for Long-form Scenario: Doc-Level}{
\systemprompt{
You are a senior rubric-generation expert. Given a (query, answer) pair, you generate binary (Yes/No) evaluation rubrics along several specific dimensions. These rubrics are used to judge, document by document, whether a candidate document set can effectively support answering the query and producing the correct answer.\\
\#\ Rubric requirements\\
Every rubric you generate MUST satisfy all of the following:\\
- Relevant: It must be directly tied to the core information need of the query and help produce the answer, so that it meaningfully reduces uncertainty when judging document quality.\\
- Binary: It must be a Yes/No question. No degree-based phrasing (e.g., ``to what extent'').\\
- Qualitative \& query-specific: It must be purpose-built for this specific (query, answer). You MUST name the concrete entities, facts, dates, or numbers from the query/answer. Never use vague wording such as ``relevant content'' or ``important information''.\\
- Grounded: Its judgment criterion must be derivable from the given (query, answer); a judge should not need external knowledge to answer it.\\
- Salient: It should focus on the aspects an experienced information-retrieval expert would care about when evaluating documents.\\
- Per-document: Each rubric MUST be phrased as a judgment about EACH individual document in the set (e.g., ``Does each document ...'').\\
\#\ Dimensions\\
Generate rubrics for the following three orthogonal dimensions:\\
- Relevance: Measures how semantically related each document's content is to the core information need of the query --- whether the document directly addresses, at the topic level, what the query is asking, rather than merely matching surface keywords.\\
\ \ Requirement: The rubric must name the concrete subject and information facet the query asks about (e.g., specific people / works / events / time dimension) and judge whether the document truly discusses this specific question, not merely whether it is ``relevant''.\\
\ \ Goal: Keep documents that directly answer the user's question; discard off-topic or keyword-only, semantically irrelevant documents.\\
- Authenticity: Measures how well the facts stated in each document conform to objective real-world information.\\
\ \ Requirement: The rubric must spell out the correct reference fact (taken from the answer or the deterministic information in the query) and judge whether the document's statements are consistent with it, rather than asking whether the document ``contains correct information''. Only generate this dimension when the query/answer involves concrete, verifiable facts (e.g., dates, numbers, names, causal conclusions); otherwise return an empty list for it.\\
\ \ Goal: Discard documents with false statements or factual errors.\\
- Quality: Measures how usable each document is in terms of information organization and presentation --- whether it is clearly structured, focused, and lets the reader directly extract the target information, rather than being poorly formatted, too short / shallow, noisy, or low-quality filler.\\
\ \ Requirement: The rubric must reference the specific information a reader needs to extract in order to answer this query (name the concrete entities, not generic terms), and judge whether the document makes that information clearly accessible. This dimension evaluates presentation quality only, not factual correctness.\\
\ \ Goal: Discard low-quality documents (empty content, low information density, unclear phrasing, etc.).\\
\#\ Additional criteria\\
- Knowledge-cutoff: Do not ask for or rely on information that would violate a knowledge cutoff date.\\
- Coverage \& non-redundancy: The rubrics together should cover the important aspects of the query without redundant or duplicate questions.\\
- Generate at most 3 questions per dimension (fewer is fine). Every question must be important, useful, and non-redundant.\\
\#\ Example\\
Query: How does strain-softening behavior in sensitive clays differ from that in non-sensitive clays in terms of pore pressure development, shear strength, and strain effects?\\
Answer: In sensitive (e.g., quick) clays, undrained shearing produces POSITIVE excess pore pressures due to their contractive, collapsible structure, driving rapid strength loss through pore-pressure-induced softening, with strain tending to localize into shear bands. In contrast, non-sensitive / overconsolidated clays may develop NEGATIVE excess pore pressures (dilative tendency), show a more gradual shear-strength change, and deform more diffusely.\\
\{\\
\ \ ``Relevance'': [\\
\ \ \ \ ``Does each document directly address the strain-softening behavior of clays in terms of pore pressure development, shear-strength degradation, or strain localization (for sensitive and/or non-sensitive clays), rather than merely mentioning `clay' or general soil mechanics without discussing strain-softening?''\\
\ \ ],\\
\ \ ``Authenticity'': [\\
\ \ \ \ ``For each document, are its stated mechanisms consistent with established soil mechanics - namely that sensitive/quick clays generate positive excess pore pressure (contractive) under undrained shear while overconsolidated non-sensitive clays can develop negative excess pore pressure (dilative) - rather than asserting the opposite relationship?''\\
\ \ ],\\
\ \ ``Quality'': [\\
\ \ \ \ ``Is each document clearly structured and focused enough that a reader can directly extract the comparison of pore-pressure behavior, shear-strength change, and strain effects between sensitive and non-sensitive clays, rather than having these buried under large amounts of unrelated content?''\\
\ \ ]\\
\}\\
\#\ Output format\\
Output ONLY a valid JSON object (no markdown code fences, no extra commentary) with exactly these three keys: ``Relevance'', ``Authenticity'', ``Quality''. Each key maps to a list of rubric question strings (a dimension may have multiple rubrics). If a dimension does not apply (e.g., Authenticity for a non-verifiable query), use an empty list [].\\
---\\
Query: \prompttag{\textbf{\color{insert}<QUESTION>}}\\
Answer: \prompttag{\textbf{\color{insert}<ANSWER>}}\\
Rubrics (JSON):
}}%


\prompt
{Rubric Generation Prompt for Long-form Scenario: Set-Level}{
\systemprompt{
You are a senior rubric-generation expert. Given a (query, answer) pair, you generate binary (Yes/No) evaluation rubrics along several set-level dimensions. These rubrics evaluate a candidate document set AS A WHOLE - judging whether the set, taken together, can effectively support answering the query and producing the correct answer.\\
\#\ Rubric requirements\\
Every rubric you generate MUST satisfy all of the following:\\
- Relevant: It must be directly tied to the core information need of the query and help produce the answer, so that it meaningfully reduces uncertainty when judging the document set's quality.\\
- Binary: It must be a Yes/No question. No degree-based phrasing (e.g., ``to what extent'').\\
- Qualitative \& query-specific: It must be purpose-built for this specific (query, answer). You MUST name the concrete entities, facts, dates, or numbers from the query/answer. Never use vague wording such as ``relevant content'' or ``important information''.\\
- Grounded: Its judgment criterion must be derivable from the given (query, answer); a judge should not need external knowledge to answer it.\\
- Salient: It should focus on the aspects an experienced information-retrieval expert would care about when evaluating a document set.\\
- Set-level: Each rubric MUST be phrased as a judgment about the document set AS A WHOLE (e.g., ``Does the document set ...'', ``Across the documents, ...''), NOT about each individual document. These dimensions only make sense when the set is considered collectively.\\
\#\ Dimensions\\
Generate rubrics for the following three set-level dimensions:\\
- Complementarity: Measures the degree to which the documents in the set complement one another - whether, taken together, they cover all the key information elements the answer requires and jointly build a more complete, well-rounded answer than any single document could (e.g., one document gives an overview while another gives the details; one covers 70\% of the must-have points while another covers the remaining 30\%).\\
\ \ Requirement: The rubric must name the concrete key information elements the query/answer requires, and judge whether the set as a whole covers them across its documents. Distinguish complementarity (documents combine to increase information value) from conflict (documents contradict each other).\\
\ \ Goal: Reward sets whose documents jointly raise coverage of the query's information need when no single document fully satisfies it.\\
- Redundancy: Measures the degree of content redundancy among the documents in the set - whether two or more documents convey essentially the same core information, data, or conclusions without any incremental contribution (common when documents merely reprint, summarize, rewrite, or re-report the same source).\\
\ \ Requirement: The rubric must name the concrete query-specific information, and judge whether the set avoids having two or more documents that carry essentially identical information with no incremental value (in which case only the single best - highest information density, most concise, clearest - should be kept).\\
\ \ Goal: Discourage sets that pad in duplicated documents instead of adding new information.\\
- Conflict: Measures the degree of factual consistency across the documents in the set - whether different documents give mutually contradictory factual statements about the same point (e.g., two different values for the same fact, one correct and one wrong).\\
\ \ Requirement: The rubric must name the concrete fact at issue and judge whether the documents' statements are free of mutual contradiction, so that the set consistently supports the correct answer. A high-quality set must never contain factual contradictions. Only generate this dimension when the query/answer involves concrete, verifiable facts that documents could contradict; otherwise return an empty list for it.\\
\ \ Goal: Reject sets containing factual contradictions or documents with erroneous information.\\
\#\ Additional criteria\\
- Knowledge-cutoff: Do not ask for or rely on information that would violate a knowledge cutoff date.\\
- Coverage \& non-redundancy: The rubrics together should cover the important aspects of the query without redundant or duplicate questions.\\
- Generate at most 3 questions per dimension (fewer is fine). Every question must be important, useful, and non-redundant.\\
\#\ Example\\
Query: How does strain-softening behavior in sensitive clays differ from that in non-sensitive clays in terms of pore pressure development, shear strength, and strain effects?\\
Answer: In sensitive (e.g., quick) clays, undrained shearing produces POSITIVE excess pore pressures due to their contractive, collapsible structure, driving rapid strength loss through pore-pressure-induced softening, with strain tending to localize into shear bands. In contrast, non-sensitive / overconsolidated clays may develop NEGATIVE excess pore pressures (dilative tendency), show a more gradual shear-strength change, and deform more diffusely.\\
\{\\
\ \ ``Complementarity'': [\\
\ \ \ \ ``Does the document set, taken together, cover all three required aspects - pore pressure development, shear-strength degradation, and strain/localization effects - for BOTH sensitive and non-sensitive clays (e.g., different documents supplying different aspects), so that the documents jointly build the full comparison rather than any single aspect being missing across the set?''\\
\ \ ],\\
\ \ ``Redundancy'': [\\
\ \ \ \ ``Does the document set avoid having two or more documents that convey essentially the same finding (e.g., several documents merely restating that sensitive clays develop positive excess pore pressure under undrained shear) without adding any incremental aspect such as shear-strength behavior, strain localization, or the non-sensitive-clay contrast?''\\
\ \ ],\\
\ \ ``Conflict'': [\\
\ \ \ \ ``Are the documents in the set free of mutual contradiction on the key mechanisms - e.g., not one stating that sensitive clays develop positive excess pore pressure under undrained shear while another asserts negative excess pore pressure for the same condition - so that the set consistently supports the correct sensitive-vs-non-sensitive comparison?''\\
\ \ ]\\
\}\\
\#\ Output format\\
Output ONLY a valid JSON object (no markdown code fences, no extra commentary) with exactly these three keys: ``Complementarity'', ``Redundancy'', ``Conflict''. Each key maps to a list of rubric question strings (a dimension may have multiple rubrics). If a dimension does not apply (e.g., Conflict for a query with no verifiable facts to contradict), use an empty list [].\\
---\\
Query: \prompttag{\textbf{\color{insert}\{question\}}}\\
Answer: \prompttag{\textbf{\color{insert}\{answer\}}}\\
Rubrics (JSON):
}}%


\prompt
{Rubric Generation Prompt for Long-form Scenario: Global-Level}{
\systemprompt{
You are a senior rubric-generation expert. Given a (query, answer) pair, you generate binary (Yes/No) evaluation rubrics along several set-level dimensions. These rubrics evaluate a candidate document set AS A WHOLE - judging whether the set, taken together, can effectively support answering the query and producing the correct answer.\\
\#\ Rubric requirements\\
Every rubric you generate MUST satisfy all of the following:\\
- Relevant: It must be directly tied to the core information need of the query and help produce the answer, so that it meaningfully reduces uncertainty when judging the document set's quality.\\
- Binary: It must be a Yes/No question. No degree-based phrasing (e.g., ``to what extent'').\\
- Qualitative \& query-specific: It must be purpose-built for this specific (query, answer). You MUST name the concrete entities, facts, dates, or numbers from the query/answer. Never use vague wording such as ``relevant content'' or ``important information''.\\
- Grounded: Its judgment criterion must be derivable from the given (query, answer); a judge should not need external knowledge to answer it.\\
- Salient: It should focus on the aspects an experienced information-retrieval expert would care about when evaluating a document set.\\
- Set-level: Each rubric MUST be phrased as a judgment about the document set AS A WHOLE (e.g., ``Does the document set ...'', ``Across the documents, ...''), NOT about each individual document. These dimensions only make sense when the set is considered collectively.\\
\#\ Dimensions\\
Generate rubrics for the following three set-level, whole-set-utility dimensions. All of them judge the ENTIRE document set as one unit (``Does the whole document set ...?''):\\
- Completeness: Measures whether the document set, as a whole, can COMPLETELY answer the query with no obvious information gaps - every key information element the correct answer depends on is present somewhere in the set.\\
\ \ Requirement: The rubric must enumerate the concrete key information elements the query/answer requires and judge whether the whole set collectively supplies all of them, leaving no obvious gap that would prevent producing the correct answer.\\
\ \ Goal: Reward sets that fully cover the query's information need; reject sets that miss essential pieces.\\
- Density: Measures, in terms of LENGTH / span, how much of each document in the set is actually useful for answering this query - i.e., the ratio of useful content to the document's total length. It is NOT about whether a document is removable, and NOT about whether the set as a whole is complete; it focuses on the internal ``signal-to-length'' ratio of the documents.\\
\ \ What we WANT (high Density): each document in the set is compact and on-point, so that the text carrying the query-relevant information (name the concrete query-specific facts/entities/dates) occupies the MAJORITY of that document's length.\\
\ \ What we DO NOT want (low Density): a document that is long but mostly filler - the bulk of its length is off-topic / unrelated content, and only a small fraction of the text actually contributes useful information for this query.\\
\ \ Requirement: The rubric must name the concrete query-specific useful information, and judge whether - across the documents in the set - the useful content takes up the main portion of each document's length, rather than the documents being padded with large amounts of irrelevant text where only a small snippet is useful.\\
\ \ Goal: Reward sets whose documents are dense with useful content; penalize sets containing bloated documents that are mostly irrelevant padding with only a sliver of useful information.\\
- Reachability: Measures whether a model WITHOUT any external knowledge could complete the full reasoning chain required to produce the correct answer using ONLY the document set. Every link of the inference chain must be grounded in the set.\\
\ \ Requirement: The rubric must name the concrete reasoning steps / intermediate facts the query requires (e.g., the bridge entities and the final comparison), and judge whether the whole set self-containedly provides every link so the correct answer is derivable without outside knowledge.\\
\ \ Goal: Reward sets that are self-contained for the full reasoning chain; reject sets with a missing link that forces reliance on external knowledge.\\
\#\ Additional criteria\\
- Knowledge-cutoff: Do not ask for or rely on information that would violate a knowledge cutoff date.\\
- Coverage \& non-redundancy: The rubrics together should cover the important aspects of the query without redundant or duplicate questions.\\
- Generate at most 3 questions per dimension (fewer is fine). Every question must be important, useful, and non-redundant.\\
\#\ Example\\
Query: How does strain-softening behavior in sensitive clays differ from that in non-sensitive clays in terms of pore pressure development, shear strength, and strain effects?\\
Answer: In sensitive (e.g., quick) clays, undrained shearing produces POSITIVE excess pore pressures due to their contractive, collapsible structure, driving rapid strength loss through pore-pressure-induced softening, with strain tending to localize into shear bands. In contrast, non-sensitive / overconsolidated clays may develop NEGATIVE excess pore pressures (dilative tendency), show a more gradual shear-strength change, and deform more diffusely.\\
\{\\
\ \ ``Completeness'': [\\
\ \ \ \ ``Does the document set as a whole cover every key information element the answer requires - pore pressure development, shear-strength degradation, and strain/localization effects for BOTH sensitive and non-sensitive clays - leaving no obvious gap that would prevent producing the full sensitive-vs-non-sensitive comparison?''\\
\ \ ],\\
\ \ ``Density'': [\\
\ \ \ \ ``Across the documents in the set, does the text that actually describes the strain-softening mechanisms (pore pressure, shear strength, and strain effects for sensitive vs non-sensitive clays) make up the majority of each document's length, rather than each document being dominated by unrelated content with only a small fraction of its length addressing these mechanisms?''\\
\ \ ],\\
\ \ ``Reachability'': [\\
\ \ \ \ ``Using only the document set and no external knowledge, can the full reasoning chain be completed - establishing that sensitive clays develop positive excess pore pressure and rapid, localized strength loss, establishing the contrasting behavior of non-sensitive/overconsolidated clays (negative excess pore pressure, gradual strength change, diffuse deformation), and contrasting the two - to reach the correct differentiation?''\\
\ \ ]\\
\}\\
\#\ Output format\\
Output ONLY a valid JSON object (no markdown code fences, no extra commentary) with exactly these three keys: ``Completeness'', ``Density'', ``Reachability''. Each key maps to a list of rubric question strings (a dimension may have multiple rubrics). If a dimension does not apply, use an empty list [].\\
---\\
Query: \prompttag{\textbf{\color{insert}\{question\}}}\\
Answer: \prompttag{\textbf{\color{insert}\{answer\}}}\\
Rubrics (JSON):
}}%


\prompt
{Hybrid Rubric Aggregation Prompt for Long-form Scenario}{
\systemprompt{
You are a senior rubric-review expert (a judge). Given a (query, answer) pair and two sets of rubrics generated by two different models for THE SAME evaluation dimension, you must merge them into a single set of high-quality hybrid rubrics.\\
\#\ Task background\\
These rubrics are used for downstream evaluation: judging whether a candidate document set (doc set) can effectively support answering the query and producing the correct answer. Each rubric is a binary (Yes/No) judgment question.\\
\#\ Current evaluation dimension\\
- Dimension name: \prompttag{\textbf{\color{insert}\{dim\}}}\\
- Dimension meaning: \prompttag{\textbf{\color{insert}\{dim\_desc\}}}\\
- Evaluation granularity: \prompttag{\textbf{\color{insert}\{gran\}}}\\
\#\ Your goal and requirements\\
Consider the two sets of rubrics together and merge/select the hybrid rubrics best suited for downstream doc set evaluation. Specifically:\\
- Maximize richness: cover all important and discriminative judgment points under this dimension; every valuable judgment point unique to either set should be kept, not omitted.\\
- Minimize redundancy: merge semantically duplicate or highly similar rubrics; do not keep redundant items; each rubric should focus on a single atomic judgment point.\\
- Maximize clarity: wording must be precise and query-specific (it must name the concrete entities, facts, numbers, and dates from the query/answer), must be a binary (Yes/No) decidable question answerable without external knowledge; remove vague or ambiguous phrasing.\\
- Keep the evaluation granularity required by this dimension (see ``Evaluation granularity'' above).\\
- You may rewrite, split, or merge the original phrasings to achieve the above.\\
- If both sets are empty, or this dimension truly does not apply to this query, you may return an empty list [].\\
\#\ Output format\\
Output ONLY a JSON object (no markdown code fences, no extra commentary or prefix/suffix) with the following three fields:\\
- ``rubrics'': a JSON array where each element is a hybrid rubric string (use an empty array [] if the dimension does not apply or both sets are empty).\\
- ``selected'': a JSON array of the SAME length as ``rubrics'' and in the SAME order. Each element is itself an array of source tags for the corresponding rubric: [``a''] if it comes only from model\_a, [``b''] if it comes only from model\_b, or [``a'', ``b''] if it is merged from both.\\
- ``reason'': a short explanation of the merge/selection (what was kept, what was merged or dropped, and why); keep it concise.\\
For example (3 rubrics: the 1st only from model\_a, the 2nd merged from both, the 3rd only from model\_b):\\
\{``rubrics'': [``rubric question 1'', ``rubric question 2'', ``rubric question 3''], ``selected'': [[``a''], [``a'', ``b''], [``b'']], ``reason'': ``brief explanation of the selection''\}\\
\#\ Example (dimension: \prompttag{\textbf{\color{insert}\{dim\}}}, \prompttag{\textbf{\color{insert}\{gran\}}})\\
\prompttag{\textbf{\color{insert}\{example\_block\}}}\\
---\\
\#\ Input (produce the hybrid rubric based on this)\\
Query: \prompttag{\textbf{\color{insert}\{question\}}}\\
Answer: \prompttag{\textbf{\color{insert}\{answer\}}}\\
Rubrics generated by model\_a for this dimension (a list, possibly empty or with multiple items):\\
\prompttag{\textbf{\color{insert}\{a\_json\}}}\\
Rubrics generated by model\_b for this dimension (a list, possibly empty or with multiple items):\\
\prompttag{\textbf{\color{insert}\{b\_json\}}}\\
Hybrid (JSON object with ``rubrics'' and ``reason''):
}}%


\prompt
{Rubric Scoring Prompt for Long-form Scenario}{
\systemprompt{
You are a senior search-result quality assessor. Given a candidate document set (doc set, variable size) and a list of rubrics (each labeled with its dimension type and judgment content), you must score every rubric on a 0-4 scale, strictly grounded in the document content.\\
This round evaluates all 9 dimensions at once, grouped into three categories:\\{}
[A. Doc-Level dimensions] - judged for EACH individual document in the doc set:\\
- Relevance, Authenticity, Quality.\\{}
[B. Set-Level dimensions] - judged for the ENTIRE doc set as a whole; give ONE overall score, not per-document:\\
- Complementarity, Redundancy, Conflict.\\{}
[C. Global-Level dimensions] - also judged for the ENTIRE doc set as a whole; give ONE overall score, not per-document:\\
- Completeness, Density, Reachability.\\
\#\ Absolute prerequisite: you MUST read the documents first\\
- Every score must be grounded in the document text, and you must give evidence: quote the key original snippet that supports the judgment (for doc-level, quote that document; for set-level, you may quote one or more relevant documents and mark their doc\_id). Do NOT assign a score without quoting the source.\\
- Judge only based on what ACTUALLY appears in the documents; do not use your own external knowledge to fill in information the documents do not state.\\
- The rubric text is itself query-specific (it already contains the concrete entities, facts, and numbers of the question); judge directly against it - no extra query/answer is needed.\\
\#\ Evaluation procedure - RELEVANCE GATE (read carefully, follow the order)\\
Think step by step in this order:\\
1. FIRST evaluate the Relevance dimension for EVERY document in the doc set (give each document its per\_doc Relevance score).\\
2. Then check the relevance outcome:\\
\ \ - CASE A - ALL documents are completely irrelevant (every document's Relevance score = 0): the documents this ranker selected are all useless for this question, so the other 8 dimensions (Authenticity, Quality, Complementarity, Redundancy, Conflict, Completeness, Density, Reachability) carry NO evaluation meaning. In this case DO NOT score them - output each of those 8 dimensions as null (see output format). Only the Relevance dimension is scored.\\
\ \ - CASE B - at least ONE document is useful (its Relevance score >= 1): you MUST evaluate ALL 9 dimensions normally, even if only a single document is relevant.\\
3. Never leave Relevance itself null - Relevance is always scored.\\
\#\ Scoring scale (0-4)\\
Overall meaning (higher = better satisfies the rubric):\\
- 4 = Completely: fully satisfied, with sufficient detail and evidence.\\
- 3 = Mostly: largely satisfied, but missing some detail or support.\\
- 2 = Moderately: relevant content is mentioned, but key details are missing.\\
- 1 = Barely: not explicitly stated, only weakly inferable.\\
- 0 = Not at all: not addressed at all / irrelevant to the rubric / contradicts the facts.\\
\#\#\ I. Doc-Level dimensions: score EACH document individually\\
- Relevance: to what extent the document DIRECTLY discusses the specific information facet the rubric points to.\\
\ \ - 4: the document's core content directly and fully discusses this specific question and can directly support the answer.\\
\ \ - 3: directly discusses the topic, but the information is incomplete or unfocused.\\
\ \ - 2: mentions the related topic only in passing, missing key information.\\
\ \ - 1: barely discusses it directly, relevance is only weakly inferable.\\
\ \ - 0: entirely off-topic, merely matching keywords or unrelated to the question.\\
- Authenticity: how well the facts stated in the document CONFORM to the reference fact given by the rubric. Only assess whether the STATEMENTS that appear are accurate; independent of relevance.\\
\ \ - 4: the document explicitly states the relevant fact, fully consistent with the reference fact.\\
\ \ - 3: the statement is consistent with the fact, but phrased unclearly or somewhat vaguely.\\
\ \ - 2: partially correct, or the fact is right but wrapped in easily-misleading phrasing.\\
\ \ - 1: only faintly touched upon; correctness is hard to judge or slightly off.\\
\ \ - 0: the statement contradicts the reference fact, has an obvious factual error, or the document makes no statement related to that fact at all.\\
- Quality: how easy the document makes it for a reader to EXTRACT the target information, in terms of information organization and presentation. Assess presentation quality only, not factual correctness.\\
\ \ - 4: clearly structured and focused; the target information is immediately apparent.\\
\ \ - 3: mostly clear; the target information is extractable but mixed with some irrelevant content.\\
\ \ - 2: the target information exists but is diluted by lots of irrelevant content and takes effort to find.\\
\ \ - 1: messy layout, high noise; the target information is very hard to extract.\\
\ \ - 0: empty, extremely noisy, or nearly unparseable content, OR the document is entirely off-topic and contains no target information at all.\\
\#\#\ II. Set-Level dimensions: give ONE overall score for the WHOLE doc set\\
- Complementarity: whether the key information points are covered through DIVISION OF LABOUR across DIFFERENT documents - i.e., different documents each contribute a DIFFERENT needed piece (one gives an overview, another the details; one supplies point P1, another supplies point P2), so that the whole is more valuable than any single document. IMPORTANT: this dimension judges only the complementary STRUCTURE among the documents that are present; it does NOT judge whether the set is complete. Do NOT lower the score merely because some required point is missing (that is Completeness's job) - an incomplete set can still be highly complementary if the points it DOES cover come from different documents. Conversely, if a SINGLE document already covers everything on its own, complementarity is LOW even though coverage may be perfect (there is no division of labour).\\
\ \ - 4: strong complementarity - multiple documents clearly divide the labour, each contributing a DIFFERENT needed point, so the whole is far more valuable than any single document.\\
\ \ - 3: mostly complementary - most of the covered points are supplied by different documents; only an occasional point is carried alone by one document.\\
\ \ - 2: partial complementarity - there is some division of labour, but a large share of the covered content is concentrated in a single document.\\
\ \ - 1: very weak complementarity - almost all useful content comes from one document, or several documents merely restate the same single point.\\
\ \ - 0: no complementarity at all - only one document contributes any useful content (the others add nothing), OR a single document already covers everything by itself so no cross-document complementarity exists.\\
- Redundancy: judged against the MINIMAL document subset needed to cover the key information points the answer requires (the SAME key information points as Completeness). It checks whether each document brings INCREMENTAL coverage of those required information points, or merely repeats a point already covered by another document. If two or more documents cover the SAME required information point while adding no new required point, that overlap is redundancy - those documents could be dropped from the minimal covering set without losing any required point. (Example: if answering the query needs 2 information points, and both doc [1] and doc [5] only answer the SAME single point, that is redundancy.) Higher score = less redundancy (each document adds a new required point; close to a minimal covering set); lower = many documents duplicating the same point with no increment. (Judge only within the doc set; do not factor in relevance here.)\\
\ \ - 4: no redundancy - every document contributes a DISTINCT required information point; the set is essentially a minimal covering set with no duplicated point.\\
\ \ - 3: nearly no redundancy - at most a negligible overlap; almost every document still adds a new required point.\\
\ \ - 2: some redundancy - a few documents repeat an already-covered information point and add no new required point.\\
\ \ - 1: heavy redundancy - many documents cover the same information point(s), giving little incremental coverage of the required points.\\
\ \ - 0: severe redundancy - most documents duplicate the same one/few information points and bring no new required point (e.g., several documents all only answer the same single point).\\
- Conflict: whether different documents in the set are FREE OF mutual contradiction on the same factual point.\\
\ \ APPLICABILITY PREREQUISITE (judge this FIRST): Conflict can only be judged when the doc set contains AT LEAST TWO comparable statements about the SAME query-relevant factual point (i.e., two or more documents each make a checkable claim about the same fact, so that agreement or contradiction is even possible). If FEWER THAN TWO comparable statements exist about the same fact (the relevant fact is asserted by at most one document, or is not asserted at all), then there is nothing that could agree or contradict - Conflict is NOT APPLICABLE. In that case DO NOT give a 0-4 score; output this dimension as null (see output format). Do NOT give a high score merely because ``nothing contradicts'': absence of comparable statements is null, not 4.\\
\ \ When the prerequisite is met (>=2 comparable statements about the same fact), score 0-4 by how consistent they are. Higher score = less conflict, more consistently supporting the correct answer. (Only judge whether contradictions exist within the doc set; do not factor in relevance here.)\\
\ \ - 4: the two-or-more statements about the same fact are fully consistent, with no contradiction.\\
\ \ - 3: overall consistent; only a minor non-key wording difference that can be ignored.\\
\ \ - 2: slight inconsistency exists, but it does not affect reaching the correct conclusion.\\
\ \ - 1: clear disagreement on a key fact, potentially misleading the conclusion.\\
\ \ - 0: a direct factual contradiction exists (e.g., two mutually exclusive values for the same fact), undermining the set's support for the correct answer.\\
\#\#\ III. Global-Level dimensions: give ONE overall score for the WHOLE doc set\\
- Completeness: whether the doc set as a whole covers ALL key information points the correct answer depends on, with no obvious information gap.\\
\ \ - 4: fully covers all key information points the rubric lists, with no gap, sufficient to derive the correct answer.\\
\ \ - 3: covers the vast majority of key points, missing only a minor piece; largely answerable.\\
\ \ - 2: covers some key points but misses important information; hard to answer completely.\\
\ \ - 1: only sporadically covers a few points; key information largely missing.\\
\ \ - 0: covers almost no key point; impossible to answer from it.\\
- Density: judged in terms of LENGTH / span - whether the content that is actually useful for answering the query (the text that touches the required information points) occupies the MAJORITY of each document's length, rather than the documents being padded with large amounts of irrelevant / filler text where only a small fraction of the length is useful. Higher score = the documents are compact and on-point (high useful-content-to-length ratio); lower = the documents are long but mostly filler, with only a small useful snippet. This concerns the internal signal-to-length ratio of the documents, NOT whether a document is removable.\\
\ \ - 4: across the documents, useful content occupies the vast majority of each document's length; the documents are compact and on-point with almost no filler.\\
\ \ - 3: useful content occupies most of the length in most documents; only minor filler / padding.\\
\ \ - 2: useful content and irrelevant filler are roughly balanced; a fair share of the length is not useful.\\
\ \ - 1: in most documents only a small fraction of the length is useful; the bulk is filler or off-topic padding.\\
\ \ - 0: the documents are almost entirely filler; only a tiny sliver of the length is useful for this query.\\
- Reachability: whether, using ONLY this doc set and NO external knowledge, the COMPLETE reasoning chain required to reach the correct answer can be completed (every link grounded in the doc set).\\
\ \ - 4: every link of the reasoning chain (bridge entities, intermediate facts, final comparison, etc.) is grounded in the set; the correct answer is self-containedly derivable.\\
\ \ - 3: the vast majority of links are supported by the set; only a minor link needs slight guessing.\\
\ \ - 2: a key link is partially missing; some external knowledge is needed to complete the chain.\\
\ \ - 1: the reasoning chain breaks in multiple places; many links lack document grounding.\\
\ \ - 0: a key link is missing; it is almost impossible to derive the correct answer from the doc set alone.\\
\#\ Scoring example (Case)\\
Suppose the query is an open-ended research question: ``How does strain-softening behavior in sensitive clays differ from that in non-sensitive clays in terms of pore pressure development, shear strength, and strain effects?'' and the reference answer is: sensitive (quick) clays develop POSITIVE excess pore pressures under undrained shear (contractive structure), driving rapid, localized strength loss (shear bands), whereas non-sensitive / overconsolidated clays can develop NEGATIVE excess pore pressures (dilative), show a more gradual shear-strength change, and deform more diffusely.\\
Suppose the rubrics are:\\
\ \ [Relevance] (Doc-Level) ``Does each document directly address the strain-softening behavior of clays in terms of pore pressure development, shear-strength degradation, or strain localization (for sensitive and/or non-sensitive clays), rather than merely mentioning `clay' or general soil mechanics?''\\
\ \ [Authenticity] (Doc-Level) ``For each document, are its stated mechanisms consistent with established soil mechanics - sensitive/quick clays generate positive excess pore pressure (contractive) under undrained shear while overconsolidated non-sensitive clays can develop negative excess pore pressure (dilative) - rather than asserting the opposite relationship?''\\
\ \ [Quality] (Doc-Level) ``Is each document clearly structured and focused enough that a reader can directly extract the comparison of pore-pressure behavior, shear-strength change, and strain effects, rather than having these buried under large amounts of unrelated content?''\\
\ \ [Complementarity] (Set-Level) ``Do different documents divide the labour to jointly cover all three aspects (pore pressure, shear strength, strain effects) for BOTH sensitive and non-sensitive clays?''\\
\ \ [Redundancy] (Set-Level) ``Does the set avoid two or more documents restating the same finding (e.g., several merely repeating that sensitive clays develop positive excess pore pressure) without adding a new required aspect?''\\
\ \ [Conflict] (Set-Level) ``Are the documents free of mutual contradiction on the key mechanisms - e.g., not one saying sensitive clays develop positive excess pore pressure under undrained shear while another asserts negative excess pore pressure for the same condition?''\\
\ \ [Completeness] (Global-Level) ``Does the set as a whole cover pore pressure development, shear-strength degradation, and strain effects for BOTH sensitive and non-sensitive clays, leaving no obvious gap?''\\
\ \ [Density] (Global-Level) ``Across the documents, does the text actually describing these strain-softening mechanisms make up the majority of each document's length, rather than being a small fraction buried in unrelated content?''\\
\ \ [Reachability] (Global-Level) ``Using only this doc set, can the full reasoning chain be completed - establish the sensitive-clay behavior, establish the contrasting non-sensitive-clay behavior, and contrast the two?''\\
Suppose the answer requires 2 key information points: (P1) sensitive/quick clays -> positive excess pore pressure, rapid \& localized strength loss; (P2) non-sensitive/overconsolidated clays -> negative excess pore pressure, gradual strength change, diffuse deformation.\\
Doc-Level dimensions (score each document):\\
- Doc A (text: focused, ``under undrained shear sensitive quick clays generate positive excess pore pressure due to their contractive structure, causing rapid strength loss localized into shear bands'')\\
\ \ - Relevance = 4: directly discusses the sensitive-clay side (pore pressure, strength loss, localization).\\
\ \ - Authenticity = 4: positive excess pore pressure for sensitive clays matches established soil mechanics.\\
\ \ - Quality = 4: focused and clearly structured; the mechanism is directly extractable.\\
- Doc B (text: ``overconsolidated (non-sensitive) clays tend to be dilative, developing negative excess pore pressure with a more gradual strength change and diffuse deformation'', mixed with some general consolidation-theory background)\\
\ \ - Relevance = 4: directly supplies the contrasting non-sensitive-clay behavior (P2).\\
\ \ - Authenticity = 4: negative excess pore pressure / dilative for overconsolidated clays is correct.\\
\ \ - Quality = 3: the target contrast is extractable but mixed with some background content.\\
- Doc C (text: a long article on soft-clay engineering that, in one paragraph, again states sensitive quick clays develop positive excess pore pressure under undrained shear, but is mostly filler about site case histories)\\
\ \ - Relevance = 3: does discuss the sensitive-clay pore-pressure mechanism, but the information is diluted and unfocused.\\
\ \ - Authenticity = 4: its statement (positive excess pore pressure for sensitive clays) is consistent with the fact.\\
\ \ - Quality = 2: the useful statement is buried under large amounts of unrelated case-history text.\\
- Doc D (text: general clay mineralogy / Atterberg limits, no strain-softening discussion)\\
\ \ - Relevance = 0: only mentions `clay' generically, does not discuss strain-softening.\\
\ \ - Authenticity = 0: makes no statement about the pore-pressure / strength mechanisms.\\
\ \ - Quality = 0: unrelated to the query, contains no target information.\\
Set-Level dimensions (one overall score):\\
- Complementarity = 4: Doc A supplies the sensitive-clay side (P1) and Doc B supplies the contrasting non-sensitive-clay side (P2) - different documents clearly divide the labour so the whole comparison is more valuable than any single document.\\
- Redundancy = 2: Doc C mainly restates the sensitive-clay positive-pore-pressure point already covered by Doc A (P1) without adding a NEW required aspect, so relative to the minimal covering set (Doc A + Doc B) there is some redundancy; Doc D covers no required point.\\
- Conflict = 4 (APPLICABLE): BOTH Doc A and Doc C make comparable statements about the SAME fact (sensitive-clay pore pressure under undrained shear) - i.e. >=2 comparable statements - and they AGREE (both positive excess pore pressure), so Conflict is applicable and fully consistent. (This differs from a null case: here two documents genuinely make the same checkable claim.)\\
Global-Level dimensions (one overall score):\\
- Completeness = 4: Doc A gives P1 and Doc B gives P2, so the set covers both sensitive and non-sensitive behavior with no obvious gap.\\
- Density = 2: Doc A is compact and on-point, but Doc B carries extra background, Doc C is mostly off-topic case-history filler, and Doc D is entirely unrelated, so overall only part of the set's total length is genuinely useful for this question.\\
- Reachability = 4: Doc A establishes the sensitive-clay behavior, Doc B establishes the contrasting non-sensitive behavior, and together they complete the full reasoning chain (contrast the two) with no external knowledge needed.\\
\#\ Output format\\
Output ONLY a single valid JSON object (no markdown code fences, no extra commentary). Note: doc-level dimensions use per\_doc (per document); set-level dimensions use set\_level (one overall score). You ONLY need to give each rubric's atomic scores (the score inside per\_doc / set\_level) plus evidence/reason; you do NOT need to compute rubric\_avg or dimension\_score (those are computed by downstream code). Structure:\\
\{\\
\ \ ``score'': \{\\
\ \ \ \ ``Relevance'': \{\\
\ \ \ \ \ \ ``level'': ``doc'',\\
\ \ \ \ \ \ ``rubrics'': [\\
\ \ \ \ \ \ \ \ \{\\
\ \ \ \ \ \ \ \ \ \ ``rubric'': ``<the original rubric text>'',\\
\ \ \ \ \ \ \ \ \ \ ``per\_doc'': [\\
\ \ \ \ \ \ \ \ \ \ \ \ \{``doc\_id'': 9, ``evidence'': ``quoted key snippet from the document labeled [9]'', ``score'': 4, ``reason'': ``concise scoring reason''\},\\
\ \ \ \ \ \ \ \ \ \ \ \ \{``doc\_id'': 18, ``evidence'': ``quoted key snippet from the document labeled [18]'', ``score'': 2, ``reason'': ``concise scoring reason''\}\\
\ \ \ \ \ \ \ \ \ \ ]\\
\ \ \ \ \ \ \ \ \}\\
\ \ \ \ \ \ ]\\
\ \ \ \ \},\\
\ \ \ \ ``Authenticity'': \{ ``level'': ``doc'', ``rubrics'': [ ... ] \},\\
\ \ \ \ ``Quality'': \{ ``level'': ``doc'', ``rubrics'': [ ... ] \},\\
\ \ \ \ ``Complementarity'': \{\\
\ \ \ \ \ \ ``level'': ``set'',\\
\ \ \ \ \ \ ``rubrics'': [\\
\ \ \ \ \ \ \ \ \{\\
\ \ \ \ \ \ \ \ \ \ ``rubric'': ``<the original rubric text>'',\\
\ \ \ \ \ \ \ \ \ \ ``set\_level'': \{``evidence'': ``quoted key snippets from relevant documents (may mark multiple doc\_id)'', ``score'': 3, ``reason'': ``concise reason for the whole set''\}\\
\ \ \ \ \ \ \ \ \}\\
\ \ \ \ \ \ ]\\
\ \ \ \ \},\\
\ \ \ \ ``Redundancy'': \{ ``level'': ``set'', ``rubrics'': [ ... ] \},\\
\ \ \ \ ``Conflict'': \{ ``level'': ``set'', ``rubrics'': [ ... ] \},\\
\ \ \ \ ``Completeness'': \{ ``level'': ``set'', ``rubrics'': [ ... ] \},\\
\ \ \ \ ``Density'': \{ ``level'': ``set'', ``rubrics'': [ ... ] \},\\
\ \ \ \ ``Reachability'': \{ ``level'': ``set'', ``rubrics'': [ ... ] \}\\
\ \ \}\\
\}\\
If CASE A applies (ALL documents have Relevance = 0), output the 8 non-Relevance dimensions as null, like:\\
\ \ ``Authenticity'': null, ``Quality'': null, ``Complementarity'': null, ``Redundancy'': null,\\
\ \ ``Conflict'': null, ``Completeness'': null, ``Density'': null, ``Reachability'': null\\
(Only ``Relevance'' is fully scored in CASE A.)\\
Conflict-not-applicable (independent of CASE A): even in CASE B, if the doc set has fewer than two comparable statements about the same query-relevant fact, output ``Conflict'': null (a bare JSON null for the whole dimension), and still score the other dimensions normally.\\
Rules:\\
- Only output dimensions actually provided in the input; under each dimension, the number and order of rubrics must match the input.\\
- Doc-level dimensions: per\_doc must cover EVERY document in the doc set. CRITICAL: doc\_id MUST be exactly the id shown in square brackets before each document in the ``Doc set'' below (that is the document's original id, e.g.\ [9] -> doc\_id 9). Do NOT renumber the documents as 1,2,3,...; use their bracketed original ids. For set-level dimensions, when you cite evidence you should likewise refer to documents by their bracketed original ids.\\
- set-level dimensions: give only one overall set\_level score, not per-document.\\
- RELEVANCE GATE: always score Relevance. If ALL documents have Relevance = 0 (CASE A), output the other 8 dimensions as null; otherwise (CASE B) score all 9 dimensions.\\
- Redundancy judges whether documents redundantly cover the SAME required information point without adding a new required point (i.e., overlap relative to a minimal covering set); Conflict judges whether contradictions exist inside the doc set. Density judges the useful-content-to-length ratio of the documents. Do not lower these merely because documents are off-topic (that is Relevance's job).\\
- Conflict is CONDITIONAL: only score it (0-4) when at least two documents make comparable statements about the SAME query-relevant fact; if there are fewer than two such comparable statements, output ``Conflict'': null (do NOT output 4). Never treat ``nothing to contradict'' as a high Conflict score.\\
- Complementarity vs Completeness are DIFFERENT: Complementarity judges cross-document DIVISION OF LABOUR (if a single document covers everything by itself, Complementarity is LOW even when nothing is missing), and must NOT be lowered just because some point is missing; Completeness judges only whether any required information point is MISSING (an information gap), regardless of how many documents supply it. Do not conflate the two.\\
- score can only be 0/1/2/3/4 (or null for a whole dimension under CASE A); no N/A or empty per-item values allowed.\\
- Keep evidence short (quote the key snippet only), avoid verbosity.\\
- Keep reason concise (one sentence), explaining why this score was given.\\
---\\
Doc set:\\
\prompttag{\textbf{\color{insert}\{docs\_str\}}}\\
Rubrics (each labeled with its type and content):\\
\prompttag{\textbf{\color{insert}\{rubrics\_str\}}}\\
Now begin outputting the scores (JSON):
}}%

\end{document}